\documentclass[10pt,twocolumn,letterpaper]{article}
\usepackage[pagenumbers]{cvpr}
\usepackage{amsmath,amssymb,bm,booktabs,multirow,array,graphicx}
\usepackage{tikz}
\usepackage{cuted}
\usetikzlibrary{arrows.meta,positioning,fit,backgrounds}
\newcommand{\R}{\mathbb{R}}
\newcommand{\sg}{\operatorname{sg}}

\newcommand{\diag}{\operatorname{diag}}
\newcommand{\argmin}{\operatorname*{arg\,min}}

\usepackage{dblfloatfix}
\definecolor{cvprblue}{rgb}{0.21,0.49,0.74}
\usepackage[breaklinks,colorlinks,allcolors=cvprblue,pdfusetitle]{hyperref}
\hypersetup{pdfauthor={Mohamad Jouni, Brayan Monroy, Jorge Bacca, and Mauro Dalla Mura}}

\title{SHARPnP: Self-Supervised and Plug-and-Play Hyperspectral Super-Resolution from Real Noisy Unaligned Camera Measurements}
\author{
    Mohamad Jouni\textsuperscript{1} \quad
    Brayan Monroy\textsuperscript{1,2} \quad
    Jorge Bacca\textsuperscript{1,2} \quad
    Mauro Dalla Mura\textsuperscript{1,3} \\
    {\small \textsuperscript{1}Univ. Grenoble Alpes, CNRS, Grenoble INP, GIPSA-Lab, 38000 Grenoble, France} \\
    {\small \textsuperscript{2}Universidad Industrial de Santander, Bucaramanga, Colombia} \\
    {\small \textsuperscript{3}Institut Universitaire de France (IUF), Paris, France}
}

\begin{document}
\maketitle

\begin{strip}
    \centering
    \includegraphics[width=\linewidth]{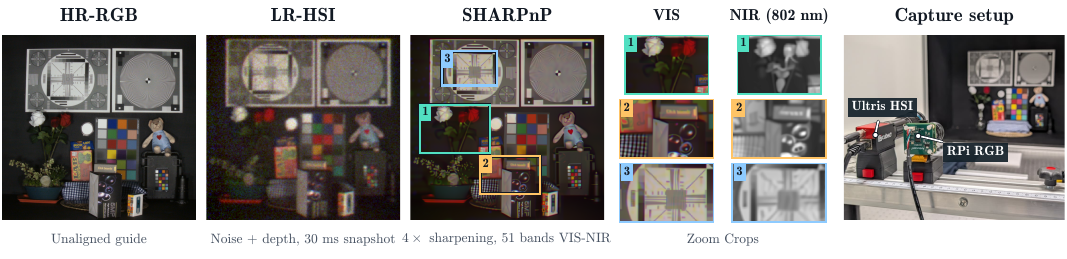}
    \captionsetup{hypcap=false}
    \captionof{figure}{%
        \textbf{Hyperspectral super-resolution under unaligned cameras.}
        SHARPnP fuses a noisy 30\,ms Ultris capture with an unaligned high-resolution RGB view to reconstruct 51 bands spanning 450--850\,nm. The reconstruction achieves over $4\times$ effective detail upscaling. Paired visible (VIS) and near-infrared (NIR, 802\,nm) crops show the same three regions:
        (1) flowers,
        (2) a foreground depth boundary, and (3) a resolution chart.
        The right panel shows the Ultris--RPi capture rig.
    }
    \label{fig:teaser}
    \captionsetup{hypcap=true}
\end{strip}

\begin{abstract}
Affordable snapshot hyperspectral cameras capture spectral images in a single exposure but suffer from limited spatial resolution and sensor noise. Low-cost RGB cameras provide fine visible details to guide NIR reconstruction, but introduce spatially varying cross-camera misalignment.
Supervised fusion methods require relatively large datasets of clean, high-resolution hyperspectral images, are sensitive to sensor noise, and generalize poorly across new sensors. We introduce SHARPnP, an interpretable, flexible framework coupling registration and reconstruction for RGB-guided hyperspectral super-resolution from noisy, unaligned measurements without clean references.
We adapt a pretrained registration model via a cross-view loss backpropagated through a fixed plug-and-play fusion module. Temporal and recorrupted pairs enable reference-free self-supervision on noisy inputs. At test time, we calibrate cross-camera radiometry and refine alignment. Plug-and-play fusion employs sequential three-band denoising.
On the REAL benchmark, SHARPnP outperforms supervised fusion baselines at $4\times$, $8\times$, and $16\times$ on clean inputs and at $4\times$ under Gaussian noise ($\sigma=5/255, 10/255$). On native RGB-HSI snapshot measurements, it reconstructs all 51 VIS-NIR bands, beyond the conventional 31-band VIS setting, without high-resolution references.
\end{abstract}

\section{Introduction}
\label{sec:intro}

Compact snapshot hyperspectral imaging (HSI) captures dense spectral measurements in a single exposure, enabling material characterization in dynamic scenes~\cite{manolakis2016hyperspectral}. Materials that look alike in RGB can have distinct spectral signatures, which are useful in applications such as robotic inspection, precision agriculture, waste sorting, nondestructive testing, heritage imaging, gas detection, and biomedicine~\cite{jouni2025multiple,9050560,10447567,contreras2025multimodal,ng2023hyper,schmitt2023rgb}. Low-cost snapshot sensors trade spatial resolution and signal-to-noise ratio for rich spectral information and acquisition speed, obscuring fine structures.

RGB-guided hyperspectral super-resolution (HSR) uses a separate relatively high-resolution camera to recover spatial detail~\cite{lai2025hsifn,jouni2026bmvc}. On ground-level, depth-dependent parallax and occlusions complicate cross-camera alignment, while differences in sensor response and HSI noise make it harder to match pixel correspondences.
Misaligned guidance can introduce spurious edges, while suppressing guidance indiscriminately leaves the reconstruction blurred~\cite{jouni2026bmvc}.

Supervised methods such as SSC~\cite{zhang2025sschs} and UAFL~\cite{zhang2026uafl} improve unaligned HSR by learning spatial-spectral features from synthesized inputs and clean high-resolution references, which are realistically hardly available. Their training conditions specify fixed degradation, spectral sampling, and scale factor, making transfer to native RGB-HSI camera pairs with noise and different sensor responses challenging.
This motivates estimating cross-camera alignment and radiometric differences directly from the acquired images. CMF-GSA-wg~\cite{jouni2026bmvc} addresses this challenge by transferring RGB details based on registration confidence. However, its single-pass fusion relies on low-resolution data in unreliable regions and has limited ability to suppress noise.

Building on confidence-aware detail transfer, we introduce \textbf{SHARPnP},
an interpretable, flexible model that couples registration and reconstruction for RGB-guided HSR from noisy, unaligned camera measurements with no clean reference. We adapt a pretrained registration model, CrossRAFT~\cite{zhou2022crossraft}, by backpropagating a cross-view loss through a fixed fusion module. This lets the quality of the fused estimate inform the flow update. Temporal and recorrupted pairs provide self-supervision on noisy inputs with no clean high-resolution references, in order to limit flow variability caused by noise.
At test time, SHARPnP calibrates cross-camera radiometry and refines residual alignment.
Plug-and-play (PnP) fusion weights RGB guidance through a pixel-wise alignment confidence and reuses a frozen pretrained denoiser. Where the warp is uncertain, the estimate relies more on the HSI observation and the denoising prior.
An efficient frozen pretrained RGB denoiser~\cite{picone2026icip} applies that prior to spectral groups, allowing it to operate across different spectral grids. Our contributions are:

\begin{figure*}[!t]
    \centering
    \includegraphics[width=1\linewidth]{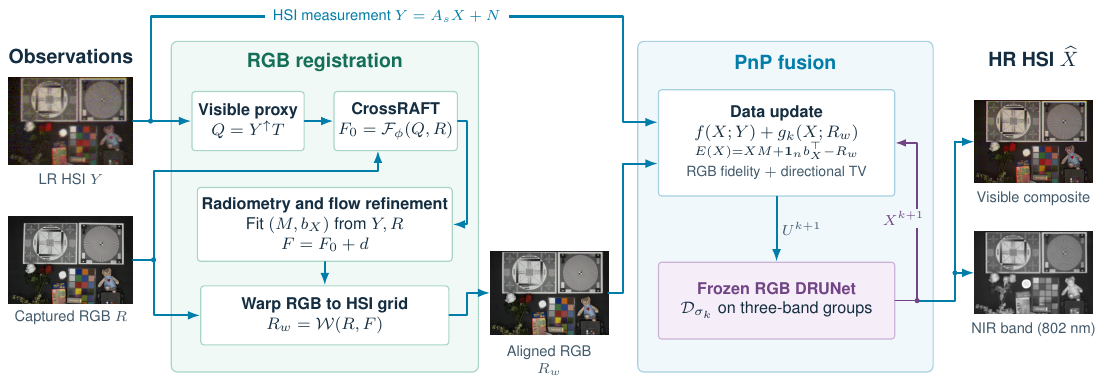}
    \caption{\textbf{Forward pass of SHARPnP.} CrossRAFT estimates an initial flow from the captured RGB image $R$ and a visible proxy of the low-resolution HSI $Y$. We fit the radiometry and refine the flow to align $R$ to the HSI grid, producing $R_w$. Plug-and-play fusion alternates a data update that combines HSI fidelity, RGB fidelity, and directional TV with a frozen RGB DRUNet applied to three bands at a time. The visible composite and the 802\,nm band are two views of the same reconstructed cube $\widehat X$ from a native dual-camera capture.  \label{fig:sharpnp-forward}}   
\end{figure*}

\begin{itemize}
    \item \textbf{Self-supervised fusion-aware registration.} We adapt registration for hyperspectral reconstruction using noisy observations with no clean high-resolution references.
    \item \textbf{Sensor-aware PnP fusion.} We combine an explicit spatial degradation model and fitted cross-camera radiometry with confidence-weighted RGB guidance and a reusable denoising prior.
    \item \textbf{Controlled and native sensor evaluation.} SHARPnP surpasses the supervised baselines on the controlled REAL benchmark on clean $4\times$ to $16\times$ and on noisy $4\times$, including sequential denoise-then-fuse controls using the same denoiser. On our collected native RGB-HSI snapshot setup, it also reconstructs 51 bands of a native VIS-NIR acquisition over 450 to 850\,nm.
\end{itemize}

\section{Related Work}
\label{sec:related}

\paragraph{Unaligned hyperspectral super-resolution.}
Single-image methods such as ESSAformer~\cite{zhang2023essa} recover structure from the HSI alone. RGB guidance supplies additional spatial detail and introduces the need to establish correspondence between views. HSIFN~\cite{lai2025hsifn} introduces REAL for evaluating guided HSR from unaligned ground-level views. In this setting, SSC~\cite{zhang2025sschs} improves flow estimation and spatial-spectral aggregation, while UAFL~\cite{zhang2026uafl} reconstructs abundance maps using deformable aggregation and cross-attention. SA-RFM~\cite{zhang2026sarfm} couples registration and fusion through Riemannian flow matching, using multispectral gradients to preserve structural boundaries.

\paragraph{Measurement-driven and model-based fusion.}
Unsupervised fusion methods can infer cross-sensor relationships from the observations. For example, $u^2$-MDN~\cite{qu2022u2mdn} relates modalities through shared representations and mutual information, while BUSIFusion~\cite{li2023busifusion} estimates degradation parameters and reconstructs the HSI from a single image pair. A complementary question is how much high-resolution spectral information is needed. SparseHSR~\cite{zhao2026sparsehsr} studies how the spectral coverage of sparse high-resolution VIS-NIR guidance affects reconstruction from spectrally complete low-resolution HSI.

\paragraph{Pretrained priors and noisy-view self-supervision.}
When the measurements leave the reconstruction underdetermined, denoising priors help regularize the estimate. Plug-and-play restoration incorporates denoisers into iterative inverse-problem solvers, as in DPIR~\cite{zhang2022dpir}, and Picone et al.~\cite{picone2026icip} explore spectral grouping to reuse RGB denoisers on hyperspectral data. Noisy observations can also supply supervision when clean references are scarce. Noise2Noise~\cite{lehtinen2018noise2noise} learns from repeated noisy observations, and GR2R~\cite{monroy2025gr2r} constructs training views from a single measurement under a specified noise model.

\section{Image Formation Model}
\label{sec:method}

\subsection{Acquistion Model}
\label{sec:preparation}
We seek a high-resolution hyperspectral image $X\in\R^{n\times B}$ from a low-resolution hyperspectral observation $Y\in\R^{n_\ell\times B}$ and a high-resolution RGB image $R\in\R^{n_R\times3}$ of the same scene from a different viewpoint. Each column is a vectorized band or color channel. We adopt the blur-and-downsample model used for REAL~\cite{zhang2026uafl},
\begin{equation}
    Y=A_sX+N,\qquad A_s=S_sB_h,
    \label{eq:observations}
\end{equation}
where $B_h\in\R^{n\times n}$ applies spatial blur and $S_s\in\R^{n_\ell\times n}$ downsamples by factor $s$. The additive noise $N\in\R^{n_\ell\times B}$ is unobserved.

\subsection{Registration Model}
\label{sec:registration-model}
The registration function $\mathcal{W}$ uses a dense correspondence $F$ to sample the RGB $R$ on the HSI output grid. On the valid overlap, we model the aligned RGB as an affine spectral projection of the latent image~\cite{simoes2015hysure,jouni2026bmvc},
\begin{equation}
    R_w=\mathcal{W}(R,F)\simeq XM+\mathbf 1_n b_X^\top,
    \label{eq:matrix-warp}
\end{equation}
where $M\in\R^{B\times3}$ is the effective spectral response, $b_X\in\R^3$ is a scene-specific RGB offset, and $\mathbf 1_m$ is a column of $m$ ones. For calibration pair $j$, $M_j=P_R\diag(g_j)$, where $P_R\in\R^{B\times3}$ has unit column sums and $g_j\in\R^3_{>0}$ contains the positive RGB gains for that pair. 

\section{Method}
\label{sec:reconstruction}
SHARPnP has two main blocks (Fig.~\ref{fig:sharpnp-forward}). Fusion-aware RGB registration aligns the captured image $R$ to the HSI grid to obtain the guide $R_w$. Sensor-aware plug-and-play (PnP) fusion combines $R_w$ with the HSI observation $Y$ using its sensing operator $A_s$ to produce the final hyperspectral estimate $\widehat X$.

\subsection{Fusion-Aware RGB Registration}

\label{sec:loss}

Registration proceeds in three stages. CrossRAFT adaptation uses reconstruction feedback from pairs of observed views. At test time, radiometric calibration fits the affine cross-camera response using the initial flow $F_0$, and residual alignment refines that flow with a bounded displacement $d_\Theta$. The refined flow $F$ samples the captured RGB image $R$ on the HSI output grid to produce the guide $R_w$ for fusion,
\[
    F=F_0+d_\Theta,\qquad R_w=\mathcal W(R,F_0+d_\Theta).
\]

\paragraph{CrossRAFT adaptation.}
\label{sec:adaptation}
To compare the RGB and HSI in a common three-channel representation, we form the visible proxy $Q=Y^\uparrow T$ on the output grid~\cite{jouni2026bmvc}. Here $Y^\uparrow\in\R^{n\times B}$ is the bicubic upsampling of $Y$, and $T\in\R^{B\times3}$ selects three visible bands. We fine-tune pretrained CrossRAFT~\cite{zhou2022crossraft} using feedback from hyperspectral reconstruction. The adapted model predicts the initial flow for residual alignment,
\[
    F_0=\mathcal F_\phi(Q,R),
\]
where $F_0\in\R^{n\times2}$ maps locations on the output grid to the captured RGB image $R$, and $\phi$ denotes the trainable parameters.

During fine-tuning, each of two views of the same scene yields a flow $F_i$ from its visible proxy $Q_i=Y_i^\uparrow T$ and RGB image $R_i$. A fixed reconstruction module follows the PnP scheme in Sec.~\ref{sec:pnp-fusion} to produce a provisional high-resolution estimate $Z_i$ from the bicubic HSI lift $Y_i^\uparrow$ and the warped RGB guide $R_{w,i}=\mathcal W(R_i,F_i)$. The objective combines alignment within each view with supervision across views,
\begin{equation}
\begin{split}
 \mathcal L_{\rm ft}(\phi)={}&\mathcal L_{\rm proxy}+\mathcal L_{\rm frame}\\
 &+w_f\mathcal L_{\rm fusion}+w_s\mathcal L_{\rm smooth}.
\end{split}
 \label{eq:training}
\end{equation}
We set $w_f=0.25$ for cross-view fusion and $w_s=10^{-3}$ for flow smoothness. Table~\ref{tab:crossraft-losses} defines the four terms. Proxy agreement aligns RGB with the HSI within each view, and frame consistency encourages the two reconstructions to agree. Cross-view fusion uses both modalities of the other view to guide spectral content and spatial detail. Flow smoothness limits abrupt changes in the estimated flow.

\begin{table}[t]
    \centering
    \caption{\textbf{Registration Losses.} $Z_i$ denotes the initial HSI estimate. $R_{w,i}$ and $\widetilde R_i$ denote the warped RGB guide and proxy, respectively. $\bar V_i$ is a unit-mean mask shared across color channels. MAE is mean absolute error, $\sg$ indicates stop-gradient, and $d\in\{x,y\}$ denotes the spatial direction.}
    \label{tab:crossraft-losses}
    \small
    \renewcommand{\arraystretch}{1.25}
    \begin{tabular}{@{}p{.23\columnwidth}@{}p{.77\columnwidth}@{}}
        \toprule
        \textbf{Loss Name} & \textbf{Equation}\\
        \midrule
        $\mathcal L_{\rm proxy}$
        & $\sum_i\operatorname{MAE}\bigl(\bar V_i\odot(\widetilde R_i-Q_i)\bigr)$\\
        \midrule
        $\mathcal L_{\rm frame}$
        & $\tfrac12\sum_{i\ne j}\operatorname{MAE}\bigl(Z_i-\sg(Z_j)\bigr)$\\
        \midrule
        $\mathcal L_{\rm fusion}$
        & $\begin{aligned}[t]
            &\tfrac12\sum_{i\ne j}\Bigl[\operatorname{MAE}\bigl(Z_i-\sg(Y_j^\uparrow)\bigr)\\
            &\quad+\operatorname{MAE}\bigl(\sg(\bar V_j)\odot(Z_iT-\sg(R_{w,j}))\bigr)\Bigr]
        \end{aligned}$\\
        \midrule
        $\mathcal L_{\rm smooth}$
        & $\tfrac12\sum_{i,d}\operatorname{MAE}(\partial_dF_i)$\\
        \bottomrule
    \end{tabular}
\end{table}

\paragraph{Noise adaptation.}
To improve robustness to noisy HSI inputs, we fine-tune on GR2R~\cite{monroy2025gr2r} views $Y_1=Y+\tau N'$ and $Y_2=Y-\tau^{-1}N'$ formed from one measurement $Y$. For Gaussian measurement noise with standard deviation $\sigma_n$, we draw $N'\sim\mathcal N(0,\sigma_n^2I)$ independently of the measurement noise and choose $\tau\in(0,1)$. Under this model, the views share the same signal and have independent noise, allowing each to supervise the other. For native temporal data, we use two nearby frames of the same stationary scene as a Noise2Noise~\cite{lehtinen2018noise2noise} pair.

\paragraph{Radiometric calibration.}
With the shared response shape $P_R$ fixed, we fit the gains $g_s$ of $M$ and the RGB bias $b_X$ from Sec.~\ref{sec:registration-model} on the LR grid using the initial warp and a weighted variant of sensor-response fitting~\cite{simoes2015hysure}. The RGB prediction $R_\ell$ and its residual $E_\ell$ are
\[
 R_\ell=YM+\mathbf 1_{n_\ell}b_X^\top,\qquad
 E_\ell=R_\ell-A_s\mathcal W(R,F_0)\in\R^{n_\ell\times3}.
\]
Spatial degradation commutes with spectral mixing and preserves the bias on the fitting support. The calibration loss is
\begin{equation}
 \mathcal L_{\rm cal}(g_s,b_X)
 =\sum_p w_p\|E_\ell(p,:)\|_2^2+\|g_s\|_2^2.
 \label{eq:calibration}
\end{equation}
Here $w_p\ge0$ is the fixed fitting weight of LR pixel $p$.

\paragraph{Residual alignment.}
For each test pair, we parameterize the bounded residual displacement as
\begin{equation}
 d_\Theta=v\odot\Delta\tanh\!\left(\mathcal I(\Theta)\right),
 \label{eq:residual}
\end{equation}
Here $\mathcal I$ bilinearly interpolates the coarse displacement lattice $\Theta$, $\Delta$ bounds each component in output pixels, and $v$ restricts the correction to valid locations.

Let $c(\Theta)$ be the local normalized cross-correlation (LNCC) between $R_\ell$ and $A_s\mathcal W(R,F)$, averaged across RGB channels and over pixels with fixed weights $V^{\rm align}$.
\begin{equation}
 \mathcal L_{\rm align}(\Theta)=1-c(\Theta)
 +\frac{\beta_d}{n_e}\|\nabla_s d_\Theta\|_F^2
 +\frac{\beta_m}{n}\|d_\Theta\|_F^2.
 \label{eq:residual-loss}
\end{equation}
The regularizers penalize changes across $n_e$ spatial edges and displacement over $n$ pixels. Where configured, we refit the gains and bias after alignment.


\subsection{Sensor-Aware Plug-and-Play Fusion}
\label{sec:pnp-fusion}

\paragraph{Reconstruction objective.}
With alignment and radiometry fixed, we estimate $X$ from $Y$ and $R_w$ in their original numerical units. The data-update objective is $f(X;Y)+g_k(X;R_w)$, with HSI fidelity $f$ and RGB guidance $g_k$. Using the affine prediction in Eq.~\eqref{eq:matrix-warp}, we write
\begin{equation}
 \begin{aligned}
 f(X;Y)&=\frac{\|Y-A_sX\|_F^2}{2Bn_\ell},\\
 g_k(X;R_w)&=
 \underbrace{\ell_{\mathrm{RGB},k}(E(X))}_{\text{RGB fidelity}}
 +\underbrace{\mu\mathcal V(X;R_w,C_P)}_{\text{edge preservation}},\\
 E(X)&=XM+\mathbf 1_n b_X^\top-R_w.
 \end{aligned}
\end{equation}
The RGB fidelity $\ell_{\mathrm{RGB},k}$ combines confidence-weighted squared residuals in pixel values and spatial gradients. The map $C_P$ weights the influence of RGB edges in directional total variation~\cite{bungert2018directional}, with strength $\mu$. For native data, RGB guidance uses both value and gradient fidelity, with directional-TV strength $\mu=10^{-5}$.

\paragraph{Iterative updates.}
Each iteration alternates a data update and a denoising update. With half-quadratic splitting, the data update computes the approximate minimizer
\begin{equation}
 U^{k+1}\approx\argmin_{U\in\R^{n\times B}}
 \left\{ f(U;Y)+g_k(U;R_w)+\frac{\rho_k}{2}\|U-X^k\|_F^2 \right\}.
\end{equation}
The quadratic term couples the data update to $X^k$ with weight $\rho_k$. We use a quadratic majorizer for directional TV and solve the resulting subproblem by conjugate gradients. The denoiser $\mathcal D_{\sigma_k}$ supplies the implicit PnP prior~\cite{zhang2022dpir}, with conditioning level $\sigma_k$ and correction strength $a_k$,
\begin{equation}
 X^{k+1}=U^{k+1}+a_k\bigl[\mathcal D_{\sigma_k}(U^{k+1})-U^{k+1}\bigr].
 \label{eq:prior}
\end{equation}

\paragraph{Spectral denoising.}
We use pretrained RGB DRUNet~\cite{zhang2022dpir}. It processes the original spectral bands in wavelength order, three at a time~\cite{picone2026icip}. 

\begin{table*}[t]
\centering
\caption{\textbf{REAL dataset, clean reconstruction.} All reproduced results use ten test scenes. Literature rows ($\dagger$) and ($\ddagger$) come from~\cite{zhang2026uafl} and~\cite{zhang2025sschs} respectively, following the training/testing protocol of~\cite{zhang2026uafl}. Bold/underline indicate the best/second-best displayed means.}
\label{tab:clean}
\setlength{\tabcolsep}{3.2pt}
\renewcommand{\arraystretch}{1.07}
\resizebox{\textwidth}{!}{%
\begin{tabular}{llc*{12}{c}}
\toprule
 & & & \multicolumn{4}{c}{$4\times$} & \multicolumn{4}{c}{$8\times$} & \multicolumn{4}{c}{$16\times$} \\
\cmidrule(lr){4-7}\cmidrule(lr){8-11}\cmidrule(lr){12-15}
Type & Methods & Venue
 & PSNR$\uparrow$ & SSIM$\uparrow$ & SAM$\downarrow$ & RMSE$\downarrow$
 & PSNR$\uparrow$ & SSIM$\uparrow$ & SAM$\downarrow$ & RMSE$\downarrow$
 & PSNR$\uparrow$ & SSIM$\uparrow$ & SAM$\downarrow$ & RMSE$\downarrow$ \\
 & & & (dB) & & ($^\circ$) & ($10^{-2}$)
     & (dB) & & ($^\circ$) & ($10^{-2}$)
     & (dB) & & ($^\circ$) & ($10^{-2}$) \\
\midrule
\multirow{5}{*}{SISR} & Bicubic & -- & 35.40 & 0.951 & 2.137 & 2.255 & 31.90 & 0.908 & 2.735 & 3.389 & 27.49 & 0.850 & 4.015 & 5.599 \\
 & SSPSR~\cite{jiang2020sspsr}$^{\dagger}$ & TCI'20 & 39.04 & 0.976 & 2.292 & -- & 31.13 & 0.906 & 3.953 & -- & 27.20 & 0.851 & 4.412 & -- \\
 & MCNet~\cite{li2020mcnet}$^{\dagger}$ & RS'20 & 39.07 & 0.974 & 2.177 & -- & 31.31 & 0.904 & 3.610 & -- & 27.04 & 0.845 & 5.042 & -- \\
 & BiQRNN~\cite{fu2021biqrnn}$^{\dagger}$ & JSTAR'21 & 37.80 & 0.969 & 2.521 & -- & 31.66 & 0.908 & 3.438 & -- & 26.53 & 0.832 & 6.360 & -- \\
 & ESSA~\cite{zhang2023essa}$^{\dagger}$ & ICCV'23 & 39.94 & 0.982 & 2.292 & -- & 31.85 & 0.913 & 3.667 & -- & 26.70 & 0.918 & 6.360 & -- \\
\cmidrule(lr){1-15}
\multirow{10}{*}{Fusion} & Optimized~\cite{fu2019optimized}$^{\dagger}$ & CVPR'19 & 27.26 & 0.916 & 10.829 & -- & 26.99 & 0.909 & 11.173 & -- & 26.93 & 0.912 & 12.605 & -- \\
 & $u^2$-MDN~\cite{qu2022u2mdn}$^{\ddagger}$ & TGRS'22 & 30.58 & 0.936 & 4.985 & -- & 30.18 & 0.936 & 5.500 & -- & 29.84 & 0.932 & 6.245 & -- \\
 & NonReg~\cite{zheng2022nonreg}$^{\dagger}$ & TGRS'22 & 31.60 & 0.951 & 4.641 & -- & 30.00 & 0.923 & 9.568 & -- & 29.09 & 0.904 & 33.461 & -- \\
 & MoE-PNP~\cite{qu2024moepnp}$^{\dagger}$ & TNNLS'25 & 32.57 & 0.951 & 5.787 & -- & 31.12 & 0.946 & 6.818 & -- & 29.91 & 0.934 & 6.933 & -- \\
 & CMF-GSA-wg~\cite{jouni2026bmvc} & BMVC'26 & 37.66 & 0.976 & 2.117 & 1.612 & 33.90 & 0.947 & 2.687 & 2.605 & 28.34 & 0.872 & 3.836 & 5.080 \\
 & HSIFN~\cite{lai2025hsifn} & TNNLS'25 & 40.28 & 0.983 & 2.146 & 1.129 & 34.76 & 0.956 & 3.016 & 2.219 & 27.98 & 0.864 & 4.149 & 5.193 \\
 & LRTN~\cite{liu2025lrtn}$^{\dagger}$ & IJCV'25 & 38.62 & 0.978 & 3.094 & -- & 34.79 & 0.957 & 3.782 & -- & 31.41 & 0.937 & 4.641 & -- \\
 & SRLF~\cite{liu2025srlf}$^{\dagger}$ & CVPR'25 & 39.06 & 0.978 & 2.464 & -- & 34.63 & 0.953 & 3.667 & -- & 31.12 & 0.929 & 5.214 & -- \\
 & SSC-S~\cite{zhang2025sschs} & IJCV'25 & 40.70 & 0.985 & 2.151 & 1.070 & 35.72 & 0.962 & 2.963 & 1.986 & 31.51 & 0.939 & 4.512 & 3.220 \\
 & UAFL~\cite{zhang2026uafl} & CVPR'26 & \underline{41.91} & \underline{0.988} & \underline{1.908} & \underline{0.910} & \underline{37.14} & \underline{0.971} & \underline{2.669} & \underline{1.645} & \underline{31.89} & \underline{0.940} & \underline{3.784} & \underline{3.096} \\
\cmidrule(lr){1-15}
SSL/PnP & SHARPnP (ours) & -- & \textbf{43.37} & \textbf{0.991} & \textbf{1.595} & \textbf{0.770} & \textbf{37.38} & \textbf{0.973} & \textbf{2.174} & \textbf{1.640} & \textbf{32.36} & \textbf{0.946} & \textbf{3.125} & \textbf{2.970} \\
\bottomrule
\end{tabular}%
}
\end{table*}
\section{Experiments}
\label{sec:results}

\subsection{Controlled reconstruction on REAL}
\label{sec:real}
\paragraph{Data and metrics.}
REAL is composed of genuinely unaligned HSI views, from which the benchmark constructs RGB guidance by spectral projection and LR-HSI by blur/downsampling \cite{lai2025hsifn,zhang2026uafl}. Its 31 bands span 400 to 700\,nm. We use the official ten test scenes and the retained fitting split of 47 scenes. The benchmark blur is the $8\times8$ Gaussian with width 3 and an upper-left sampling phase. We report mean bandwise PSNR and SSIM, mean spectral angle (SAM, in degrees), and global RMSE. Reproduced metrics follow the training/testing protocol of~\cite{zhang2026uafl}.
For the noisy cases, independent Gaussian noise of standard deviation $5/255$ or $10/255$ is added to LR-HSI with a fixed dataset-indexed realization. Geometry and radiometric fitting use noisy observations.

\paragraph{Baselines.}
We reevaluate Bicubic, CMF-GSA, HSIFN, SSC and UAFL under the training/testing protocol of~\cite{zhang2026uafl}, together with published SISR and fusion evaluations ported from the same source.
On noisy cases, there was no direct SOTA to compare with, as the aforementioned ones propagate the noise with the signal, so we create a sequential denoise-then-fuse benchmark using the same DRUNet mechanism followed by either SSC or UAFL, dubbed as DRU-SSC and DRU-UAFL, respectively.
These controls test the practical alternative of denoising before fusion. \vspace{-1em}

\paragraph{Clean reconstruction.}
\Cref{tab:clean} shows that the retained SHARPnP configurations exceed the supervised baselines on all four metrics at each scale. At $4\times$, CrossRAFT-based reconstruction reaches 43.37\,dB versus 41.91\,dB for UAFL. At $8\times$, reconstruction reaches 37.38\,dB versus UAFL's 37.14\,dB and improves all four means over UAFL. At $16\times$, CrossRAFT-based reconstruction reaches 32.36\,dB and 3.125$^\circ$ SAM versus 31.89\,dB and 3.784$^\circ$ for UAFL. The retained configurations thus extend reconstruction gains across three spatial sampling regimes. \vspace{-1em}

\paragraph{Noise-corrupted reconstruction.}
The direct supervised baselines deteriorate substantially under added HSI noise (\cref{tab:noise}). Denoising their inputs is a much stronger comparator than direct noisy inference: DRU-UAFL reaches 39.21/37.17\,dB at the two noise levels. SHARPnP reaches 39.33/37.25\,dB and improves all four displayed metrics over both DRU baselines. Its RMSE is 0.01185/0.01512 and SAM is 2.736$^\circ$/3.375$^\circ$. These results show that the benefit extends beyond inserting a denoiser before an existing fusion network.
\begin{figure*}[!t]
    \centering
    \captionsetup[subfigure]{skip=1pt}
    \begin{minipage}{\textwidth}
        \centering
        \includegraphics[width=\linewidth]{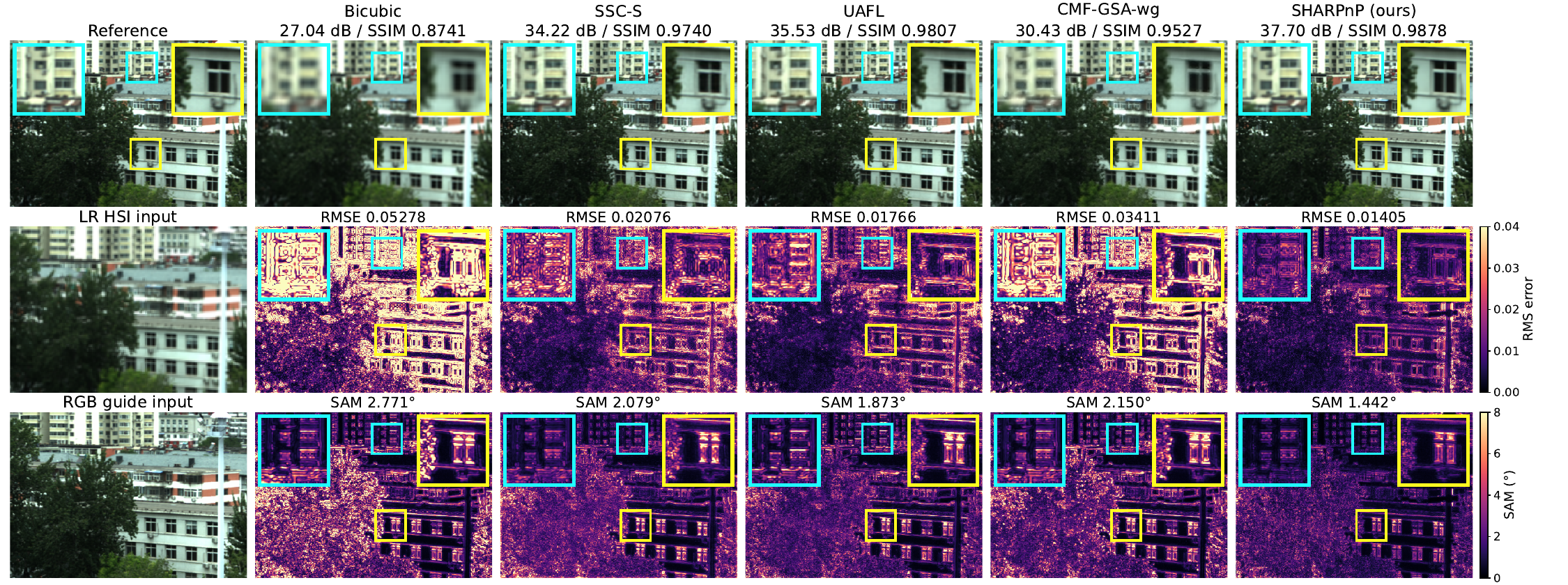}
        \subcaption{REAL scene 0; clean; $4\times$}
    \end{minipage}\\[1pt]
    \begin{minipage}{\textwidth}
        \centering
        \includegraphics[width=\linewidth]{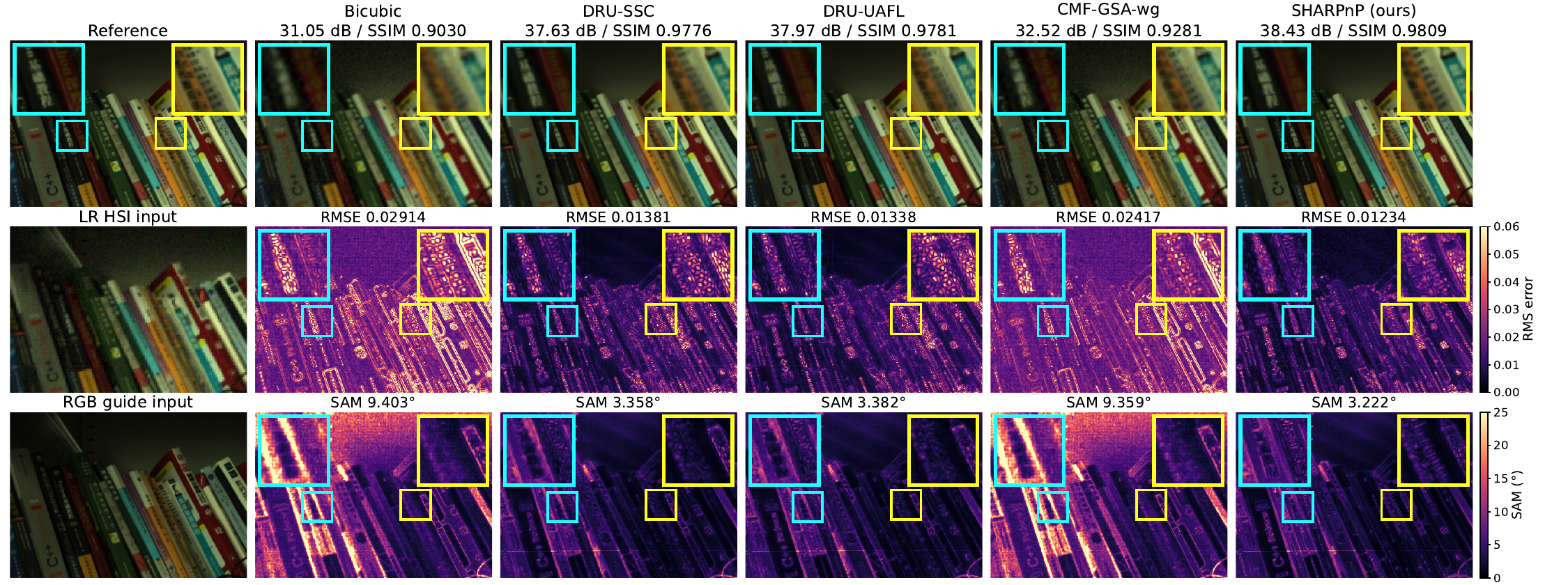}
        \subcaption{REAL scene 7; $5/255$ LR-HSI noise; $4\times$; denoise-then-fuse comparison}
    \end{minipage}
    \caption{REAL comparisons with Bicubic, SSC-S~\cite{zhang2025sschs}, UAFL~\cite{zhang2026uafl}, CMF-GSA-wg~\cite{jouni2026bmvc}, and SHARPnP (ours). The noisy panel uses DRU-SSC and DRU-UAFL. Rows show reconstructed RGB previews, RMSE maps, and SAM maps; reference and input views are at left. Annotations are full-frame metrics.}
    \label{fig:real-bmvc-dru}
\end{figure*}


\begin{table}[t]
\centering
\caption{\textbf{Noisy REAL at $4\times$.} Gaussian noise is added to LR-HSI only. DRU denotes frozen DRUNet applied before the frozen supervised fusion model. Evaluation uses ten test scenes.}
\label{tab:noise}
\setlength{\tabcolsep}{3.2pt}
\renewcommand{\arraystretch}{1.07}
\resizebox{\linewidth}{!}{%
\begin{tabular}{lcccccccc}
\toprule
 & \multicolumn{4}{c}{$\sigma_n=5/255$} & \multicolumn{4}{c}{$\sigma_n=10/255$} \\
\cmidrule(lr){2-5}\cmidrule(lr){6-9}
Methods & PSNR$\uparrow$ & SSIM$\uparrow$ & SAM$\downarrow$ & RMSE$\downarrow$ & PSNR$\uparrow$ & SSIM$\uparrow$ & SAM$\downarrow$ & RMSE$\downarrow$ \\
 & (dB) &  & ($^\circ$) & ($10^{-2}$) & (dB) &  & ($^\circ$) & ($10^{-2}$) \\
\midrule
Bicubic & 31.60 & 0.901 & 8.170 & 2.870 & 27.90 & 0.785 & 14.458 & 4.135 \\
\cmidrule(lr){1-9}
CMF-GSA~\cite{jouni2026bmvc} & 32.85 & 0.927 & 8.052 & 2.382 & 28.66 & 0.822 & 13.630 & 3.745 \\
HSIFN~\cite{lai2025hsifn} & 32.34 & 0.905 & 9.349 & 2.481 & 27.18 & 0.751 & 16.079 & 4.463 \\
SSC-S~\cite{zhang2025sschs} & 31.93 & 0.891 & 9.642 & 2.611 & 26.78 & 0.725 & 16.625 & 4.693 \\
UAFL~\cite{zhang2026uafl} & 32.21 & 0.896 & 9.813 & 2.496 & 26.53 & 0.706 & 17.677 & 4.790 \\
\cmidrule(lr){1-9}
DRU-SSC & 39.01 & 0.978 & \underline{2.837} & 1.298 & 37.08 & \underline{0.969} & \underline{3.546} & 1.604 \\
DRU-UAFL & \underline{39.21} & \underline{0.979} & 2.868 & \underline{1.231} & \underline{37.17} & \underline{0.969} & 3.557 & \underline{1.567} \\
\cmidrule(lr){1-9}
SHARPnP & \textbf{39.33} & \textbf{0.981} & \textbf{2.736} & \textbf{1.185} & \textbf{37.25} & \textbf{0.971} & \textbf{3.375} & \textbf{1.512} \\
\bottomrule
\end{tabular}}
\end{table}

  Figure~\ref{fig:real-bmvc-dru}(a) illustrates a clean $4\times$ reconstruction, complementing the quantitative comparison in Table~\ref{tab:clean}. SHARPnP recovers finer structures than the evaluated fusion baselines. Figure~\ref{fig:real-bmvc-dru}(b) presents a noisy $4\times$ example with $\sigma_n=5/255$, corresponding to Table~\ref{tab:noise}. Here, the denoise-then-fuse baselines first apply frozen DRUNet to the LR-HSI and then perform supervised fusion. SHARPnP instead jointly denoises and reconstructs through its PnP optimization, recovering finer details with fewer noise artifacts. This visual comparison supports the quantitative improvements over the sequential pipelines.

\subsection{Ablations studies}

\paragraph{Choice of the prior.}
In the matched clean $4\times$ control, replacing DRUNet by identity reduces PSNR from 43.37 to 41.83\,dB (\cref{tab:denoiser}). Other pretrained denoisers and Haar shrinkage remain usable but give lower accuracy at the tested settings. This supports reusing a capable pretrained prior rather than attributing the entire result to calibration or RGB fidelity.

\paragraph{Geometry and guidance interact.}
\Cref{tab:geometry-guidance} isolates residual registration and RGB-guidance strength under the final clean $4\times$ SHARPnP configuration. Residual correction improves PSNR by 0.54\,dB at $0.1\times$ guidance and by 1.7\,dB at the retained strength. Without residual correction, stronger guidance reduces PSNR by 0.94\,dB and worsens all four mean metrics. With corrected geometry, it improves PSNR by 0.245\,dB and improves SSIM and RMSE, but increases SAM. Adding directional VTV then improves all four metrics in every scene, reaching 43.37\,dB and SAM $1.595^\circ$. These results demonstrate the complementary effects of correspondence correction, detail transfer and spatial regularization. For more details in the implementation see SM.


\begin{table}[t]
\centering\small
\caption{\textbf{Denoiser prior ablation.} Ten-scene clean $4\times$ ablation. Haar entries are wavelet-shrinkage controls. Their thresholds were selected on four benchmark scenes. Identity applies no denoising.}
\label{tab:denoiser}
\setlength{\tabcolsep}{3pt}
\resizebox{0.8\columnwidth}{!}{%
\begin{tabular}{lrrrr}
\toprule
Denoiser & PSNR$\uparrow$ & SSIM$\uparrow$ & SAM$\downarrow$ & RMSE$\downarrow$ \\
      & (dB)           &                & ($^\circ$)      & ($\times 10^{-2}$) \\
\midrule
DRUNet~\cite{zhang2022dpir}     & \textbf{43.37} & \textbf{0.991} & \textbf{1.595} & \textbf{0.770} \\
FFDNet~\cite{zhang2018ffdnet}   & 42.88 & 0.990 & 1.624 & 0.823 \\
SCUNet~\cite{zhang2023scunet}   & 42.79 & 0.989 & 1.632 & 0.838 \\
GRUNet~\cite{lai2022dphsir}     & 42.39 & 0.989 & 1.686 & 0.860 \\
3D Haar~\cite{donoho1994wavelet} & 41.88 & 0.988 & 1.716 & 0.905 \\
2D Haar~\cite{donoho1994wavelet} & 41.88 & 0.988 & 1.718 & 0.906 \\
Identity                       & 41.83 & 0.988 & 1.722 & 0.910 \\
\bottomrule
\end{tabular}%
}
\end{table}

\begin{table}[t]
\centering\small
\caption{\textbf{Geometry, RGB guidance, and directional VTV.} Clean $4\times$, averaged over ten scenes at iteration 50. RGB strength scales both value and gradient coefficients relative to their retained values ($10^{-4}$). Calibration, confidence, frozen DRUNet and solver settings are fixed. Residual correction is bounded by $\pm2$ HR pixels; dVTV uses weight $10^{-5}$.}
\label{tab:geometry-guidance}
\setlength{\tabcolsep}{3pt}
\renewcommand{\arraystretch}{1.07}
\resizebox{\linewidth}{!}{%
\begin{tabular}{ccccccc}
\toprule
Residual & RGB strength & dVTV & PSNR$\uparrow$ & SAM$\downarrow$ & SSIM$\uparrow$ & RMSE$\downarrow$ \\
 & & & (dB) & ($^\circ$) & & ($10^{-2}$) \\
\cmidrule(lr){1-7}
Off & $\times0.1$ & Off & 41.61 & 1.694 & 0.985 & 1.062 \\
On  & $\times0.1$ & Off & 42.15 & 1.658 & 0.987 & 0.953 \\
Off & $\times1$   & Off & 40.67 & 2.040 & 0.981 & 1.259 \\
On  & $\times1$   & Off & 42.40 & 1.833 & 0.989 & 0.909 \\
\cmidrule(lr){1-7}
On  & $\times1$   & On  & \textbf{43.37} & \textbf{1.596} & \textbf{0.991} & \textbf{0.771} \\
\bottomrule
\end{tabular}%
}
\end{table}

\subsection{Native real RGB-HSI snapshot data}
\label{sec:native}

\paragraph{Acquisition and adaptation.}
We acquire sequences of a stationary scene with objects at different depths using an Ultris SR5 compact snapshot HSI camera and a Raspberry Pi HQ RGB camera. The HSI contains 51 VIS-NIR bands from 450 to 850\,nm at 8\,nm intervals; its first 31 bands span 450 to 690\,nm. We use 480\,ms exposures for higher-SNR observations and 30\,ms exposures for a 10\,fps hyperspectral video with stronger sensor noise. Measurements are dark/flat corrected, without synthetic degradation or added noise. Temporal Noise2Noise pairs adapt CrossRAFT, while radiometry and residual alignment are fitted from the measurements. Each exposure uses 70 adaptation, 10 validation and 10 diagnostic frames, without clean HR references.

\begin{figure*}[p]
    \centering
    \captionsetup[subfigure]{skip=0pt}

    \makebox[0pt][r]{%
        \rotatebox[origin=c]{90}{\footnotesize 480\,ms exposure (1.8 fps)}%
        \hspace{.026\textwidth}%
    }%
    \begin{minipage}{.90\textwidth}
        \centering
        \begin{minipage}[t]{.145\linewidth}
            \centering
            \includegraphics[width=\linewidth]{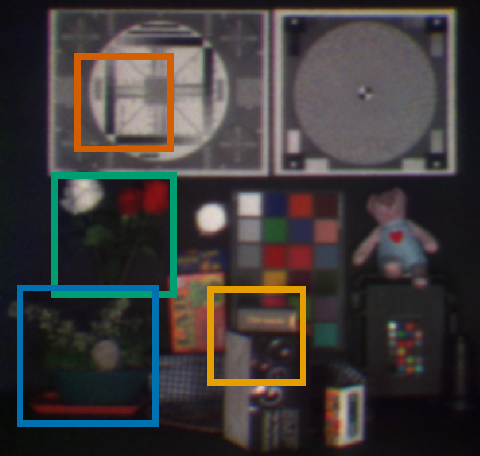}\\[1pt]
            \includegraphics[width=.4875\linewidth]{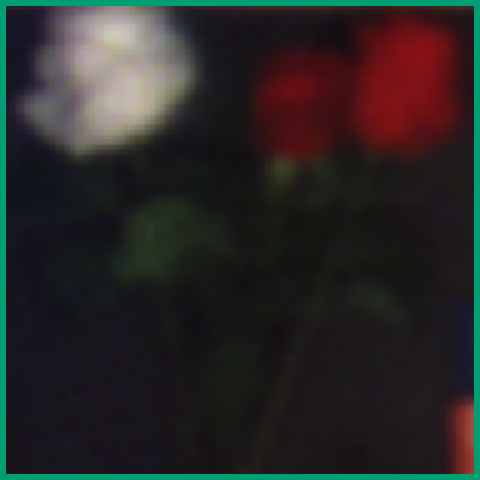}\hfill
            \includegraphics[width=.4875\linewidth]{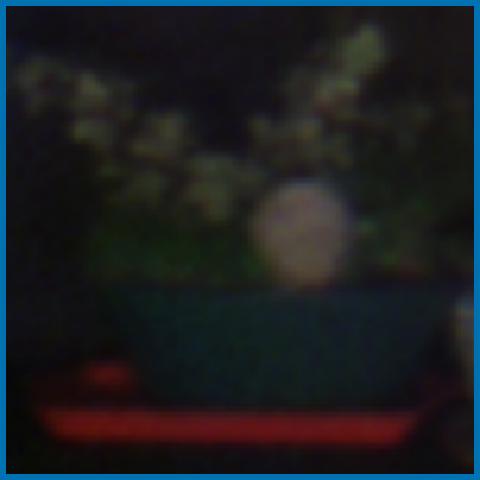}\\[1pt]
            \includegraphics[width=.4875\linewidth]{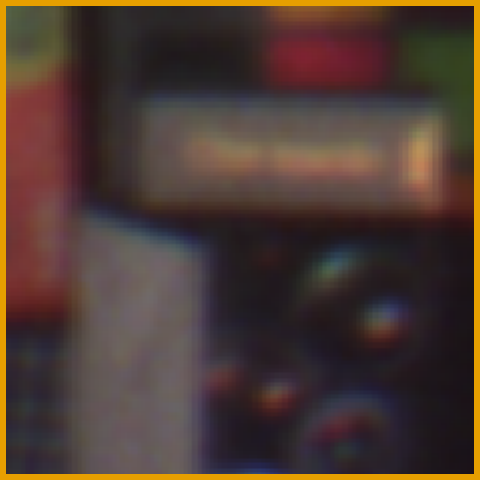}\hfill
            \includegraphics[width=.4875\linewidth]{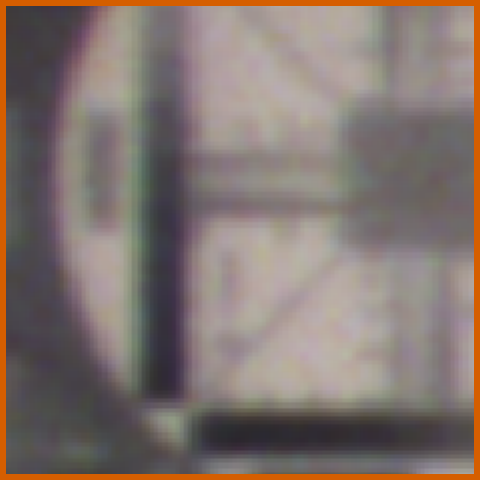}\\[1pt]
            {\scriptsize\mbox{Bicubic}\\[-1pt]\mbox{51 bands}\par}
        \end{minipage}%
        \hfill
        \begin{minipage}[t]{.145\linewidth}
            \centering
            \includegraphics[width=\linewidth]{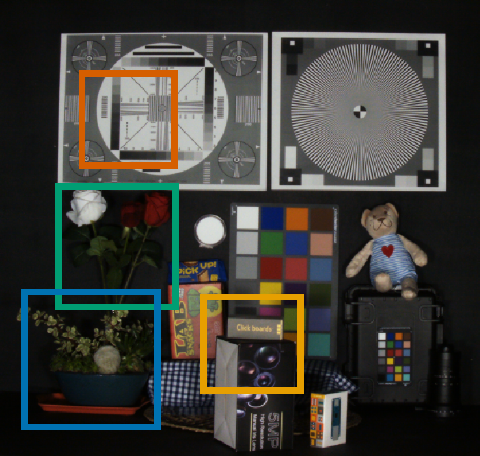}\\[1pt]
            \includegraphics[width=.4875\linewidth]{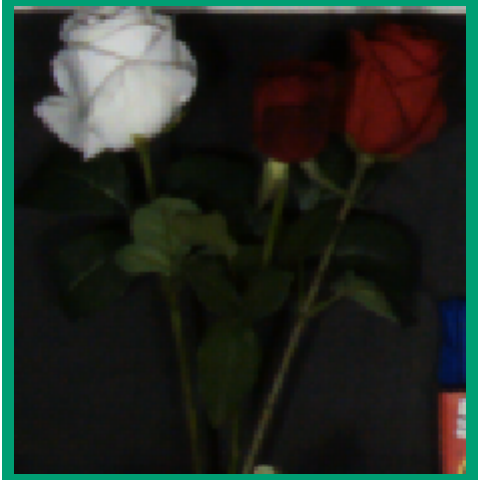}\hfill
            \includegraphics[width=.4875\linewidth]{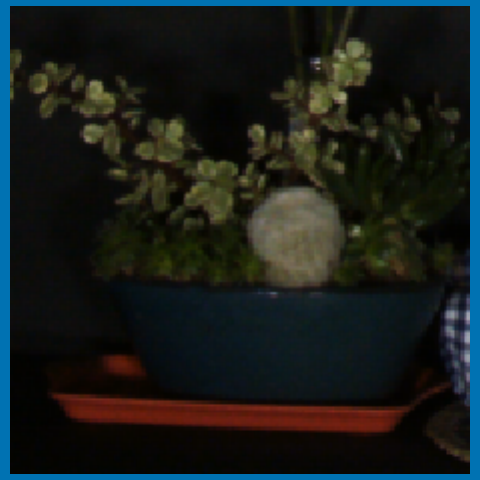}\\[1pt]
            \includegraphics[width=.4875\linewidth]{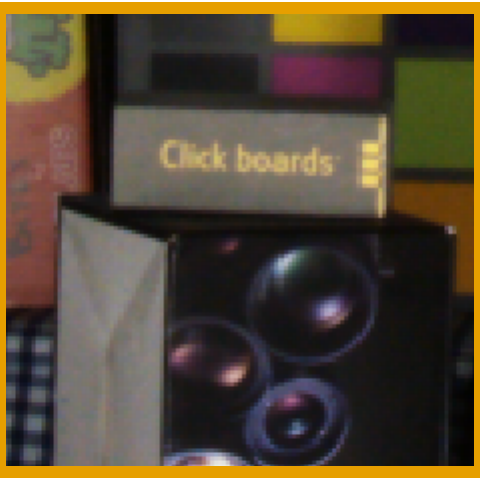}\hfill
            \includegraphics[width=.4875\linewidth]{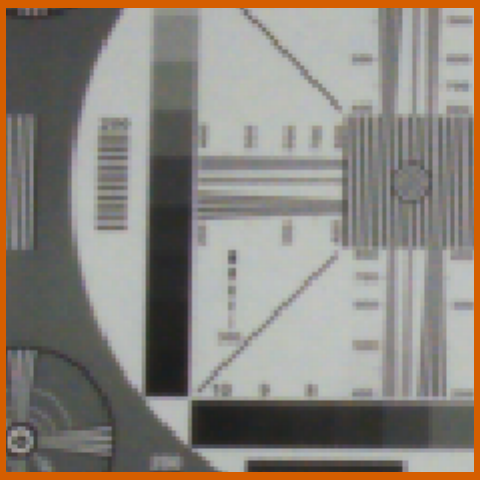}\\[1pt]
            {\scriptsize\mbox{RGB guide}\\[-1pt]\mbox{3 channels}\par}
        \end{minipage}%
        \hfill
        \begin{minipage}[t]{.145\linewidth}
            \centering
            \includegraphics[width=\linewidth]{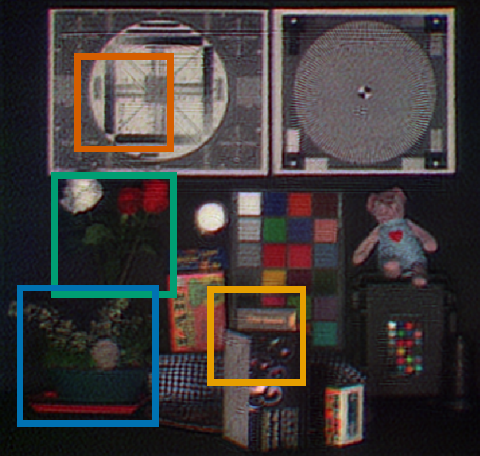}\\[1pt]
            \includegraphics[width=.4875\linewidth]{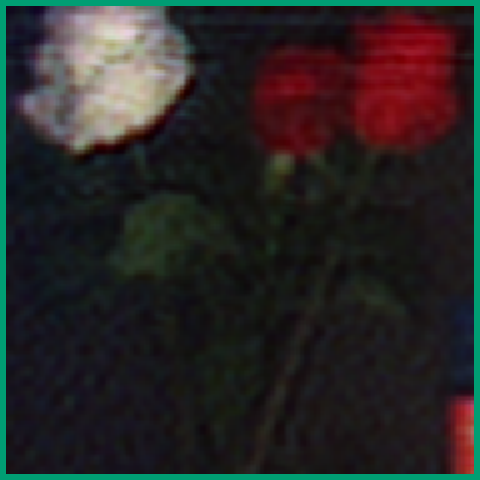}\hfill
            \includegraphics[width=.4875\linewidth]{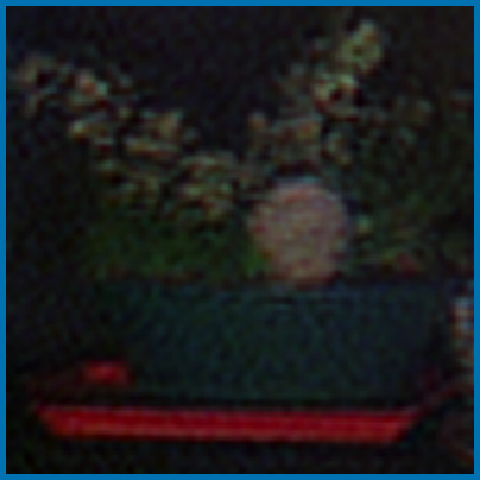}\\[1pt]
            \includegraphics[width=.4875\linewidth]{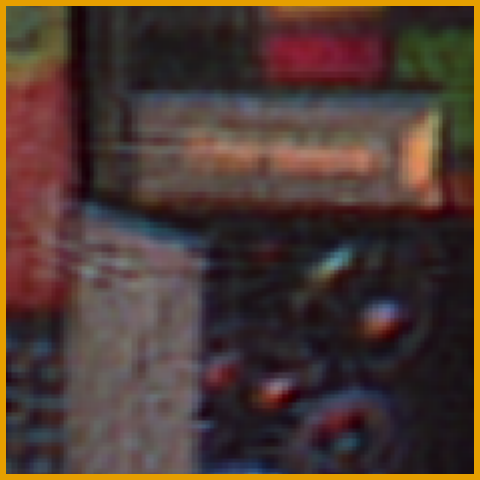}\hfill
            \includegraphics[width=.4875\linewidth]{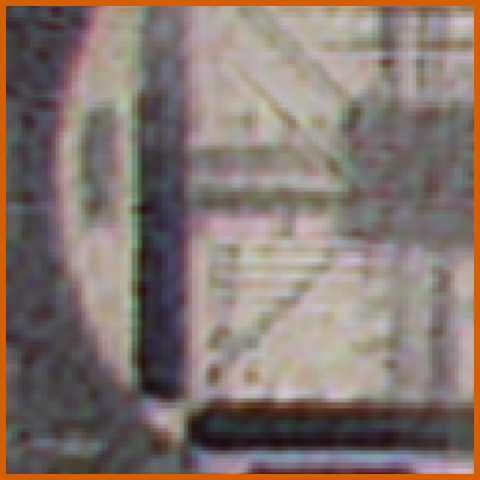}\\[1pt]
            {\scriptsize\mbox{SSC-S~\cite{zhang2025sschs}}\\[-1pt]\mbox{31 bands}\par}
        \end{minipage}%
        \hfill
        \begin{minipage}[t]{.145\linewidth}
            \centering
            \includegraphics[width=\linewidth]{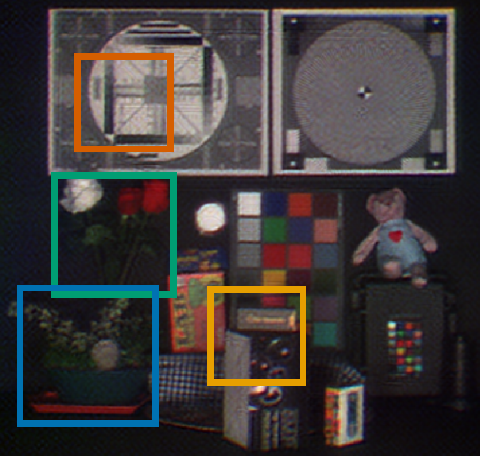}\\[1pt]
            \includegraphics[width=.4875\linewidth]{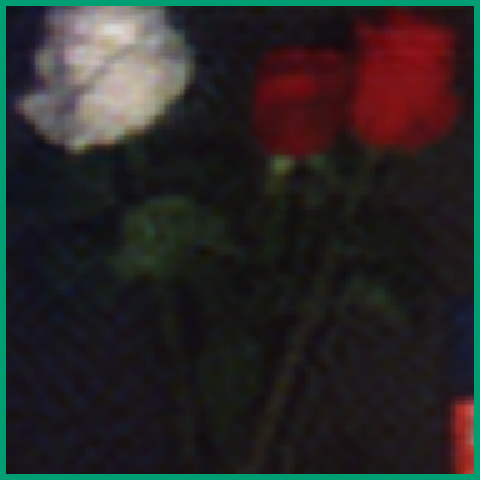}\hfill
            \includegraphics[width=.4875\linewidth]{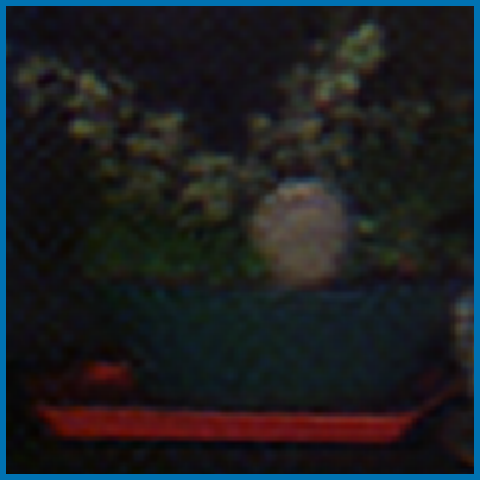}\\[1pt]
            \includegraphics[width=.4875\linewidth]{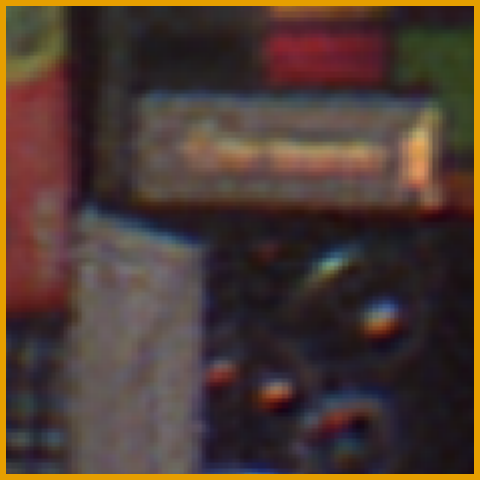}\hfill
            \includegraphics[width=.4875\linewidth]{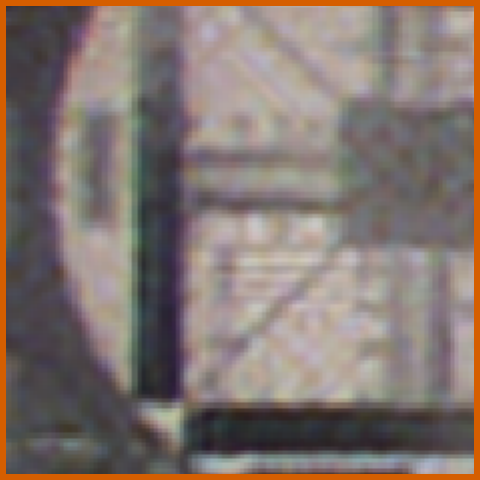}\\[1pt]
            {\scriptsize\mbox{UAFL~\cite{zhang2026uafl}}\\[-1pt]\mbox{31 bands}\par}
        \end{minipage}%
        \hfill
        \begin{minipage}[t]{.145\linewidth}
            \centering
            \includegraphics[width=\linewidth]{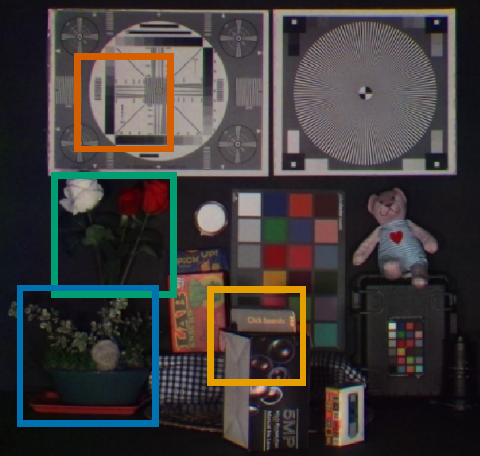}\\[1pt]
            \includegraphics[width=.4875\linewidth]{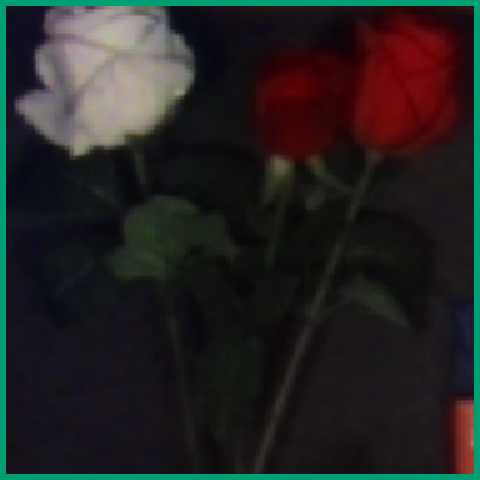}\hfill
            \includegraphics[width=.4875\linewidth]{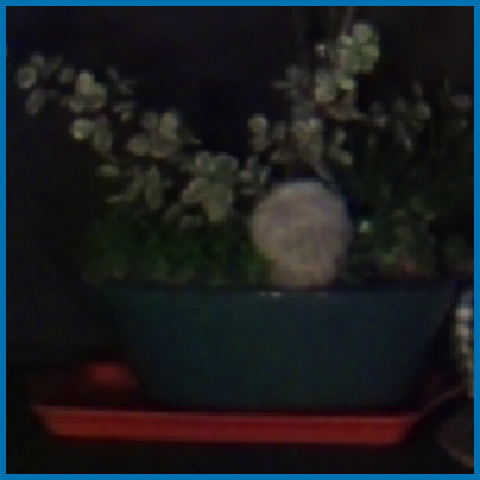}\\[1pt]
            \includegraphics[width=.4875\linewidth]{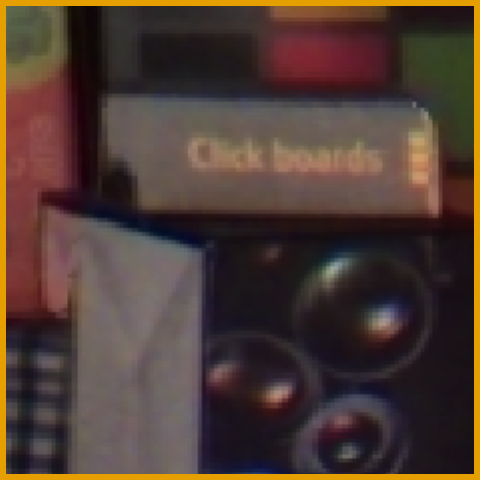}\hfill
            \includegraphics[width=.4875\linewidth]{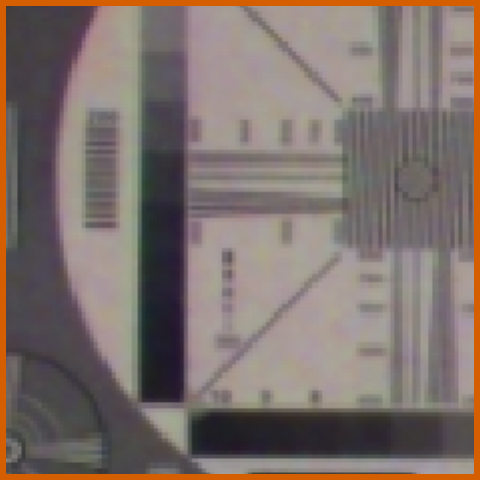}\\[1pt]
            {\scriptsize\mbox{CMF-GSA-wg~\cite{jouni2026bmvc}}\\[-1pt]\mbox{51 bands}\par}
        \end{minipage}%
        \hfill
        \begin{minipage}[t]{.145\linewidth}
            \centering
            \includegraphics[width=\linewidth]{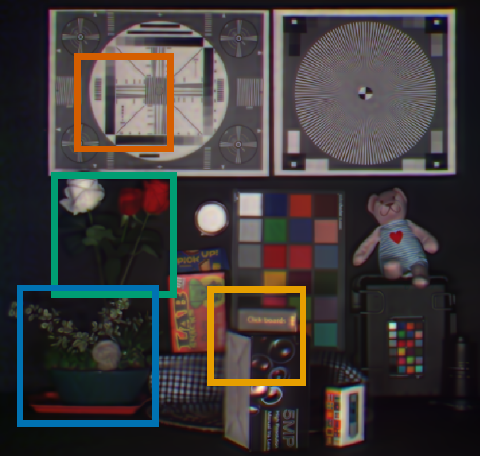}\\[1pt]
            \includegraphics[width=.4875\linewidth]{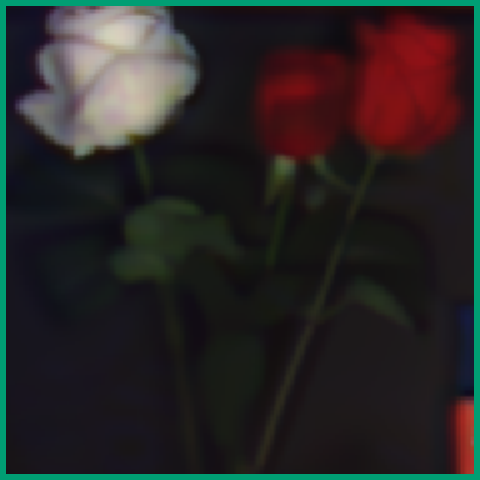}\hfill
            \includegraphics[width=.4875\linewidth]{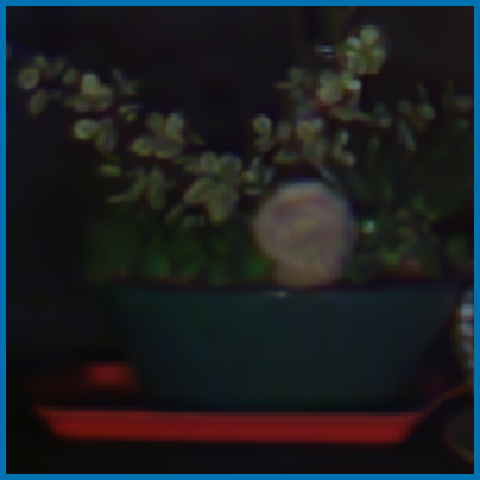}\\[1pt]
            \includegraphics[width=.4875\linewidth]{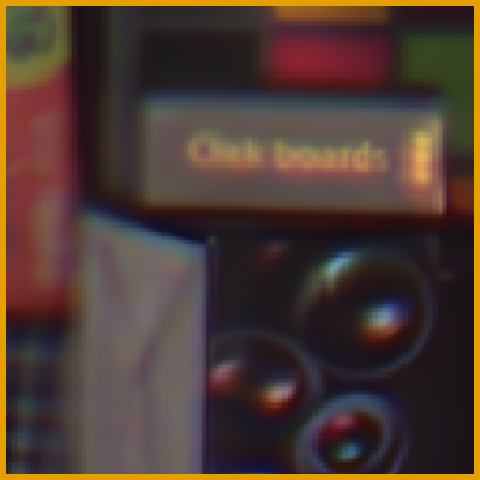}\hfill
            \includegraphics[width=.4875\linewidth]{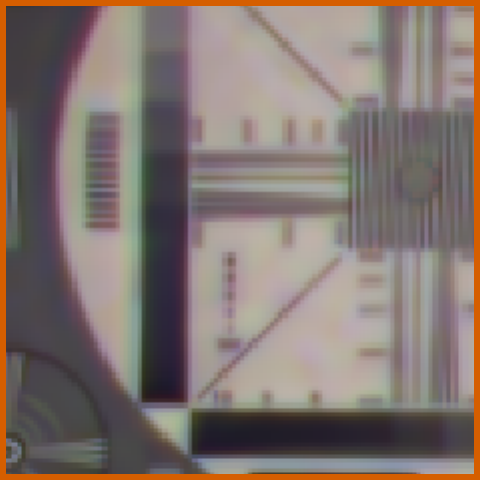}\\[1pt]
            {\scriptsize\mbox{SHARPnP (ours)}\\[-1pt]\mbox{51 bands}\par}
        \end{minipage}%
        \par
        \begin{minipage}[t]{.32\linewidth}
            \centering
            \begin{minipage}[c]{.48\linewidth}
                \includegraphics[width=\linewidth]{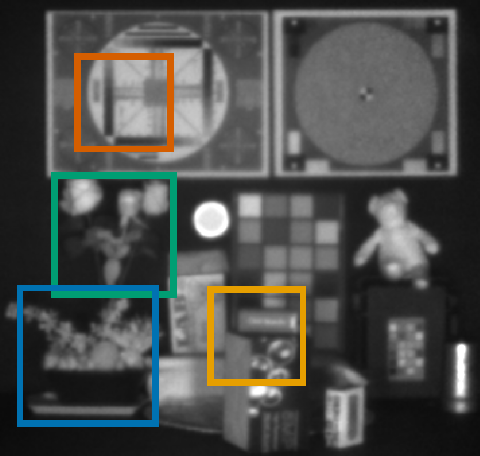}
            \end{minipage}
            \begin{minipage}[c]{.48\linewidth}
                \centering
                \includegraphics[width=.4875\linewidth]{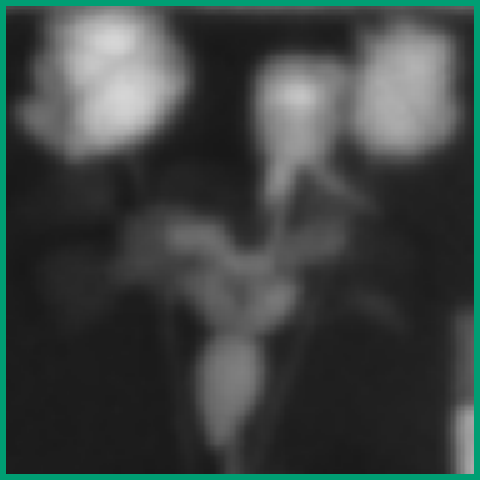}\hfill
                \includegraphics[width=.4875\linewidth]{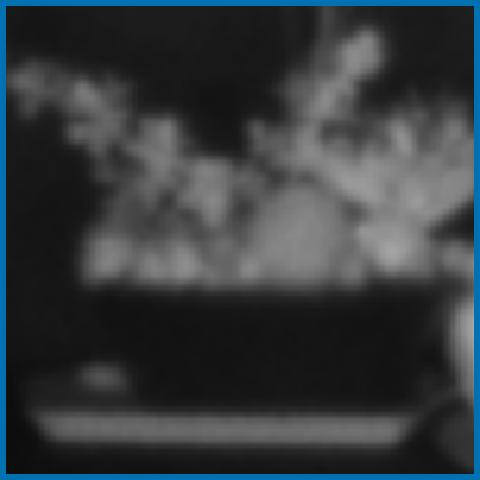}\\[1pt]
                \includegraphics[width=.4875\linewidth]{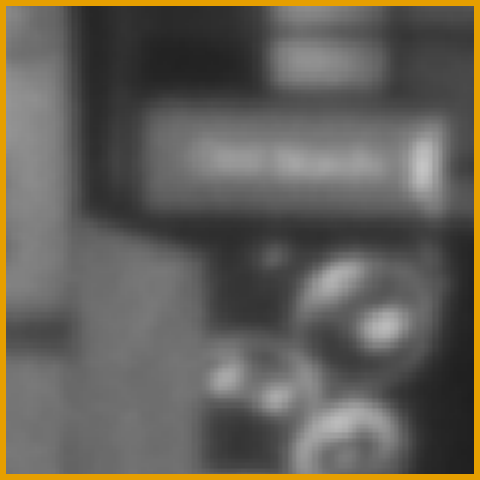}\hfill
                \includegraphics[width=.4875\linewidth]{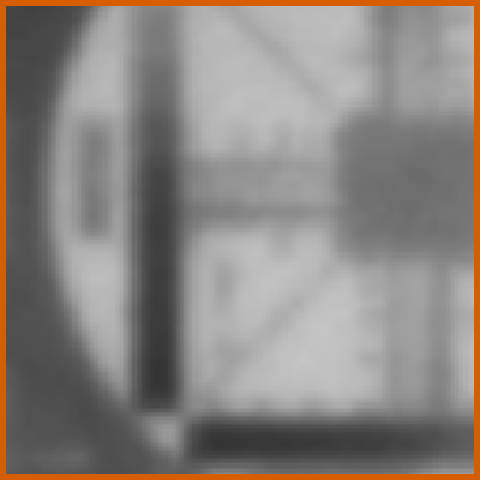}%
            \end{minipage}\\[1pt]
            {\scriptsize\mbox{Bicubic}\\[-1pt]\mbox{51 bands}\par}
        \end{minipage}%
        \hfill
        \begin{minipage}[t]{.32\linewidth}
            \centering
            \begin{minipage}[c]{.48\linewidth}
                \includegraphics[width=\linewidth]{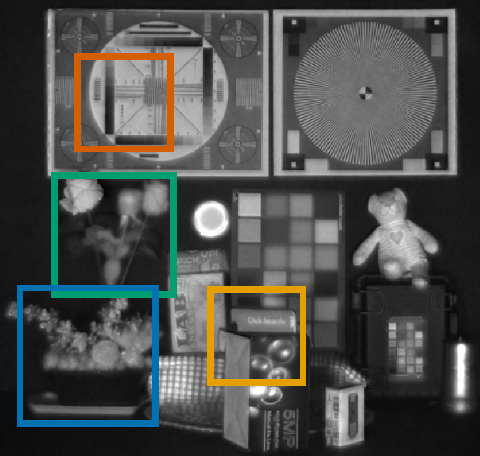}
            \end{minipage}
            \begin{minipage}[c]{.49\linewidth}
                \centering
                \includegraphics[width=.4875\linewidth]{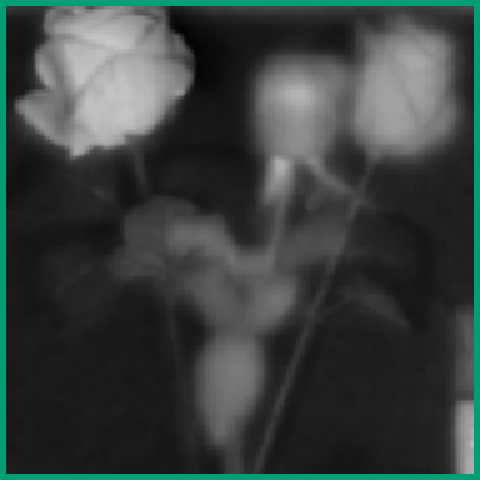}\hfill
                \includegraphics[width=.4875\linewidth]{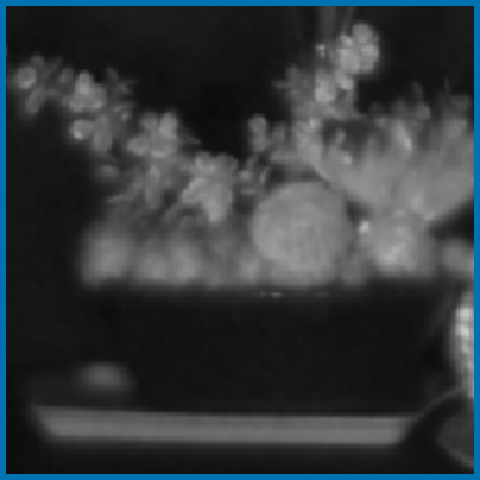}\\[1pt]
                \includegraphics[width=.4875\linewidth]{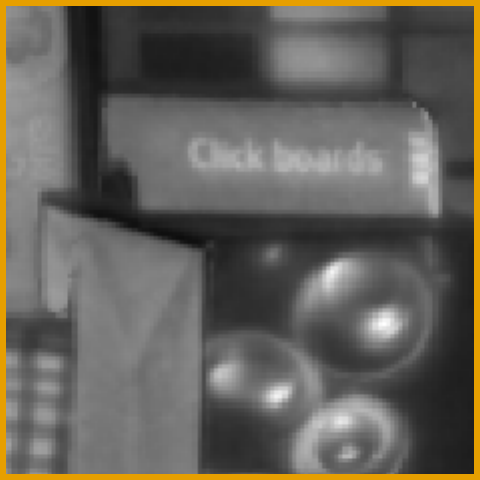}\hfill
                \includegraphics[width=.4875\linewidth]{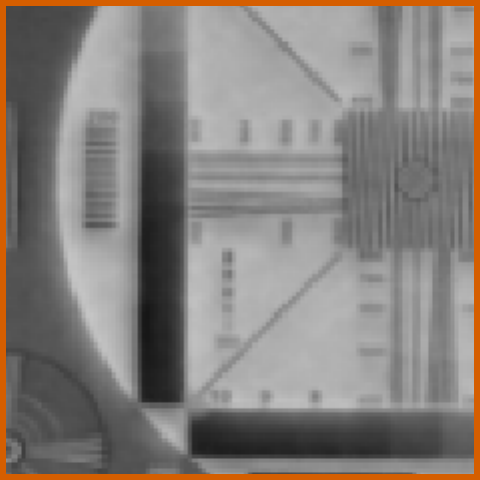}%
            \end{minipage}\\[1pt]
            {\scriptsize\mbox{CMF-GSA-wg~\cite{jouni2026bmvc}}\\[-1pt]\mbox{51 bands}\par}
        \end{minipage}%
        \hfill
        \begin{minipage}[t]{.32\linewidth}
            \centering
            \begin{minipage}[c]{.48\linewidth}
                \includegraphics[width=\linewidth]{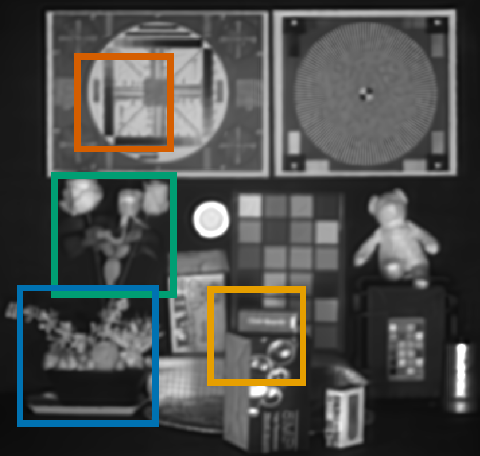}
            \end{minipage}
            \begin{minipage}[c]{.49\linewidth}
                \centering
                \includegraphics[width=.4875\linewidth]{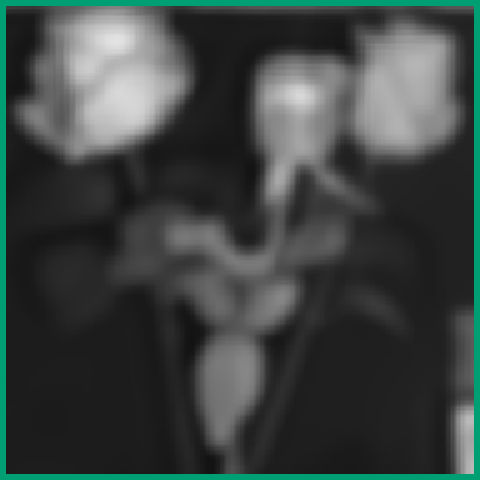}\hfill
                \includegraphics[width=.4875\linewidth]{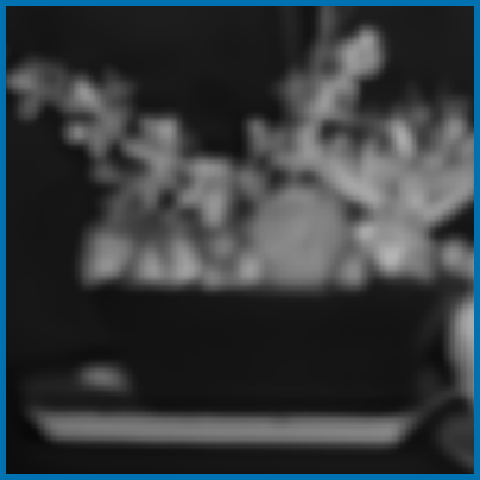}\\[1pt]
                \includegraphics[width=.4875\linewidth]{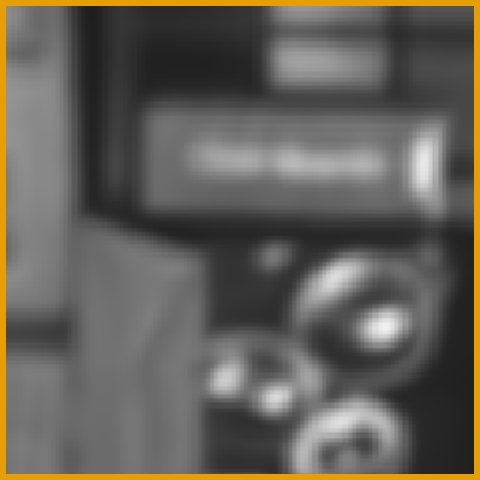}\hfill
                \includegraphics[width=.4875\linewidth]{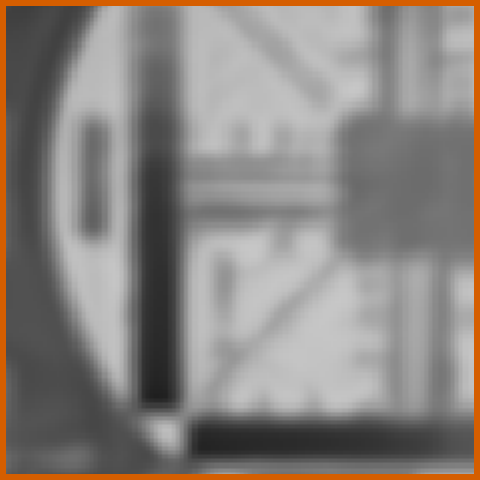}%
            \end{minipage}\\[1pt]
            {\scriptsize\mbox{SHARPnP (ours)}\\[-1pt]\mbox{51 bands}\par}
        \end{minipage}%
        \subcaption{480\,ms HSI exposure. Top: RGB composites and crops. Bottom: NIR at 802\,nm and crops.}
        \label{fig:november-parallax_480ms}
    \end{minipage}
    \vspace{2mm}

    \makebox[0pt][r]{%
        \rotatebox[origin=c]{90}{\footnotesize 30\,ms exposure (10\,fps, snapshot)}%
        \hspace{.026\textwidth}%
    }%
    \begin{minipage}{.90\textwidth}
        \centering
        \begin{minipage}[t]{.145\linewidth}
            \centering
            \includegraphics[width=\linewidth]{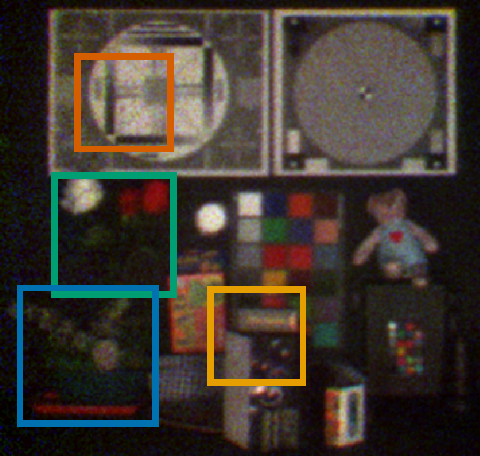}\\[1pt]
            \includegraphics[width=.4875\linewidth]{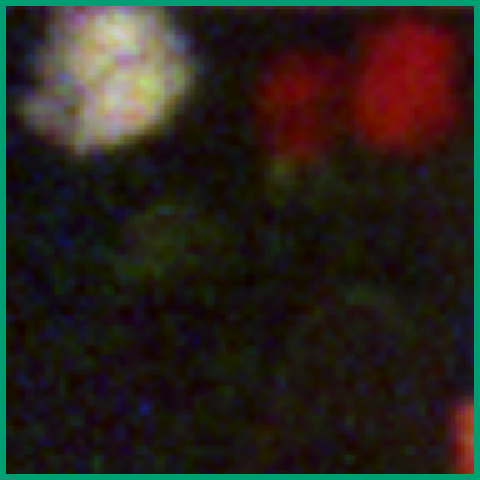}\hfill
            \includegraphics[width=.4875\linewidth]{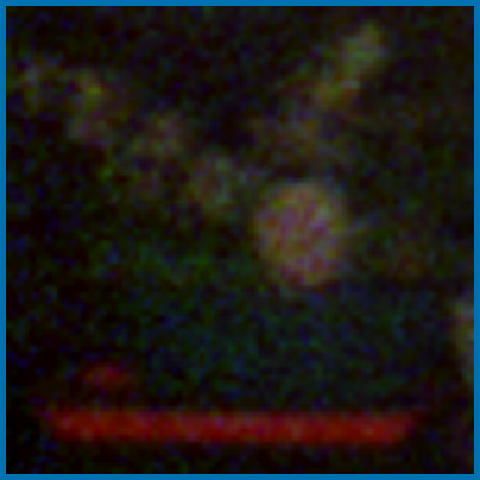}\\[1pt]
            \includegraphics[width=.4875\linewidth]{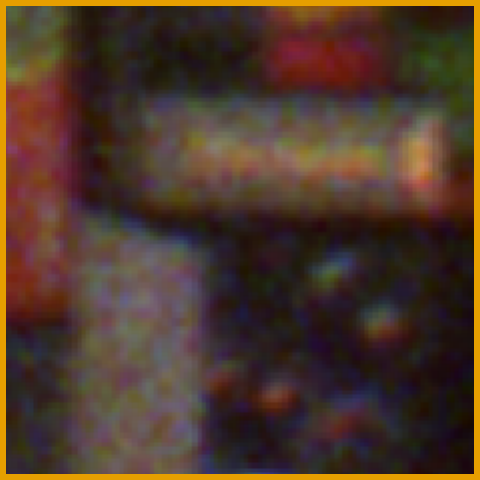}\hfill
            \includegraphics[width=.4875\linewidth]{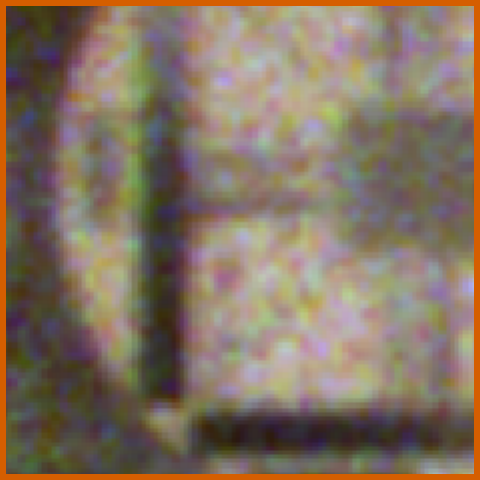}\\[1pt]
            {\scriptsize\mbox{Bicubic}\\[-1pt]\mbox{51 bands}\par}
        \end{minipage}%
        \hfill
        \begin{minipage}[t]{.145\linewidth}
            \centering
            \includegraphics[width=\linewidth]{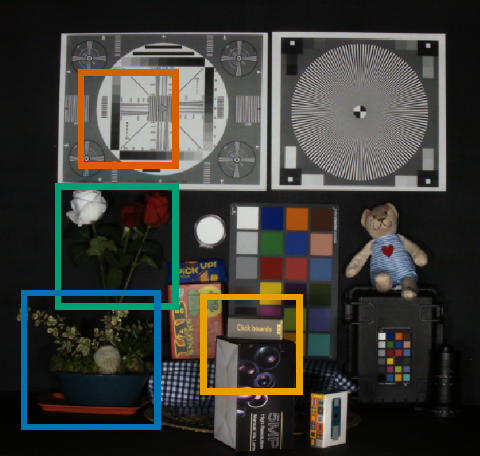}\\[1pt]
            \includegraphics[width=.4875\linewidth]{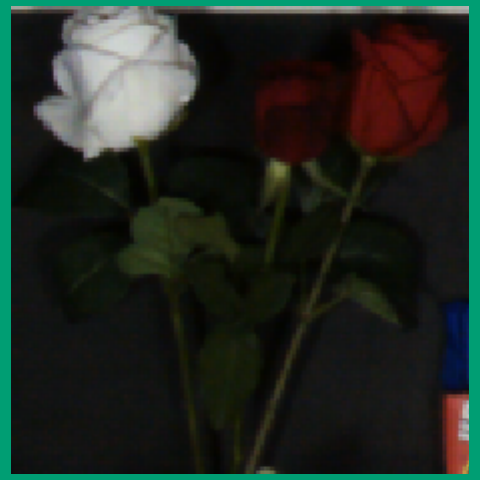}\hfill
            \includegraphics[width=.4875\linewidth]{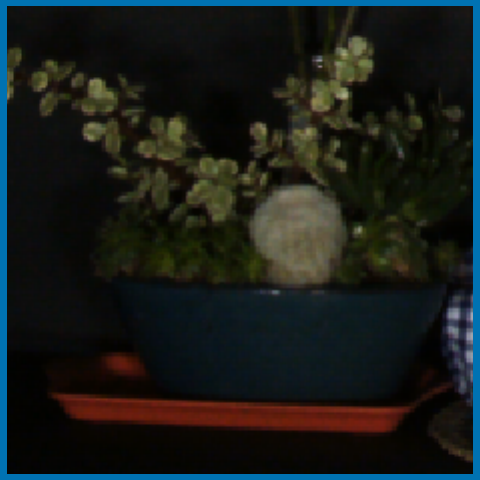}\\[1pt]
            \includegraphics[width=.4875\linewidth]{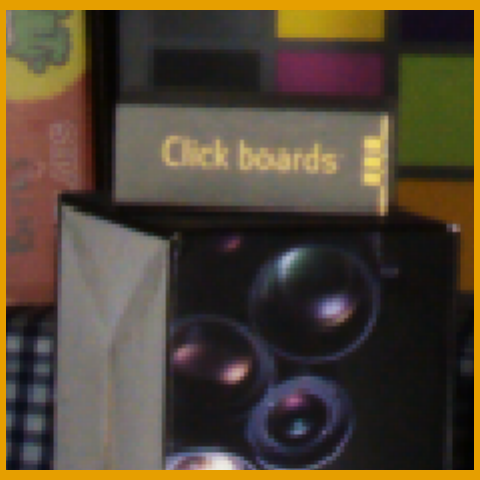}\hfill
            \includegraphics[width=.4875\linewidth]{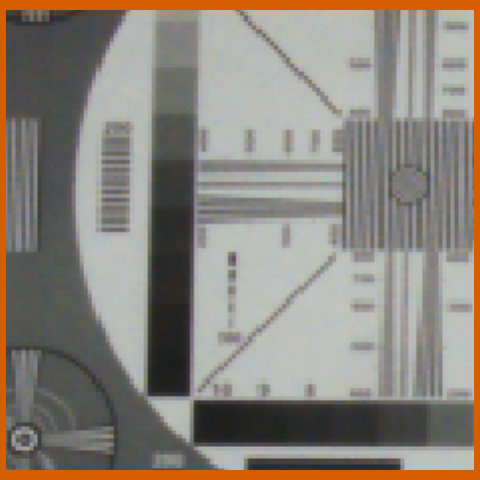}\\[1pt]
            {\scriptsize\mbox{RGB guide}\\[-1pt]\mbox{3 channels}\par}
        \end{minipage}%
        \hfill
        \begin{minipage}[t]{.145\linewidth}
            \centering
            \includegraphics[width=\linewidth]{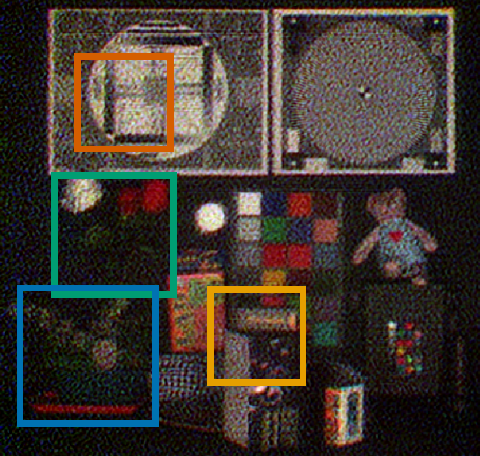}\\[1pt]
            \includegraphics[width=.4875\linewidth]{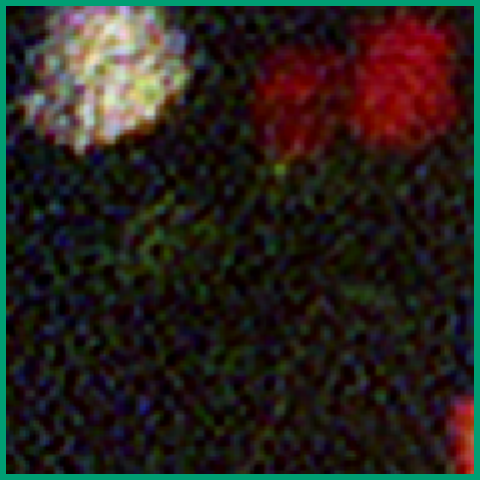}\hfill
            \includegraphics[width=.4875\linewidth]{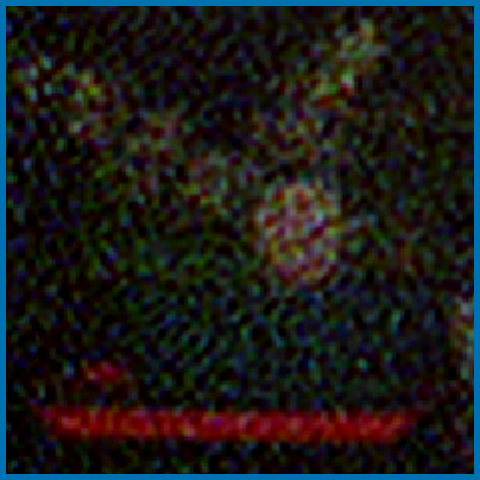}\\[1pt]
            \includegraphics[width=.4875\linewidth]{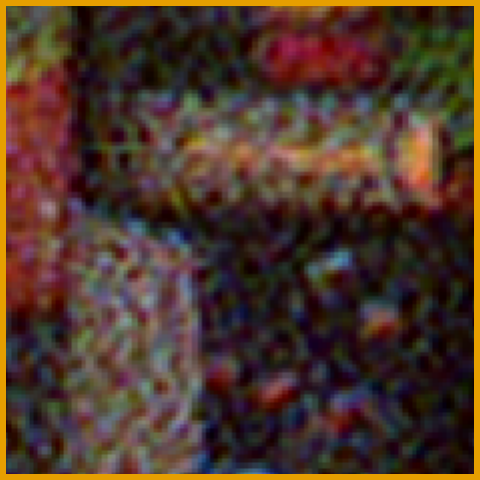}\hfill
            \includegraphics[width=.4875\linewidth]{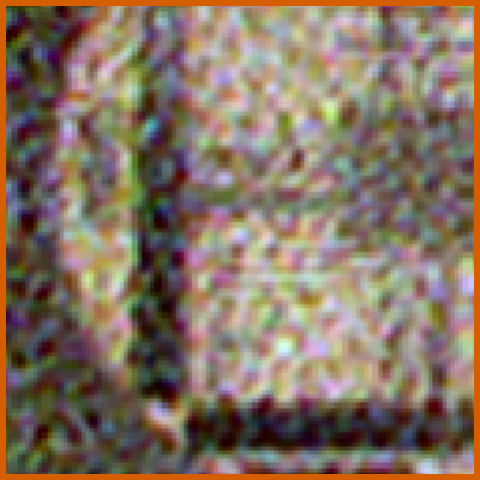}\\[1pt]
            {\scriptsize\mbox{SSC-S~\cite{zhang2025sschs}}\\[-1pt]\mbox{31 bands}\par}
        \end{minipage}%
        \hfill
        \begin{minipage}[t]{.145\linewidth}
            \centering
            \includegraphics[width=\linewidth]{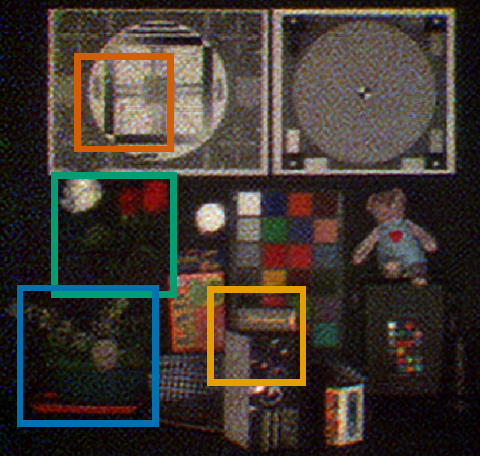}\\[1pt]
            \includegraphics[width=.4875\linewidth]{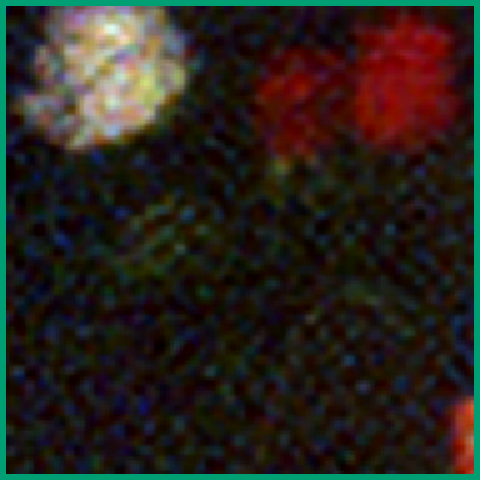}\hfill
            \includegraphics[width=.4875\linewidth]{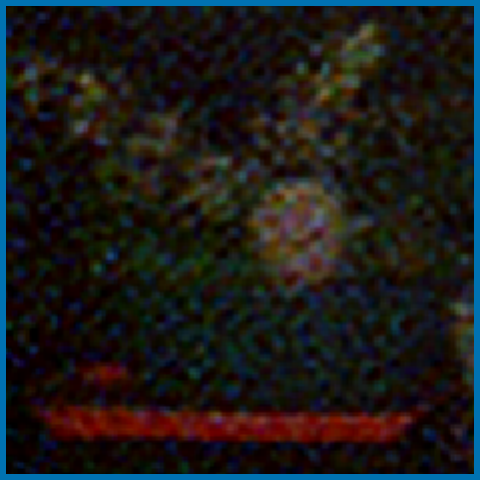}\\[1pt]
            \includegraphics[width=.4875\linewidth]{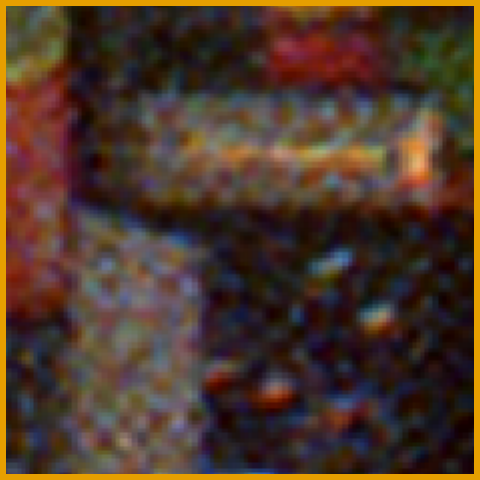}\hfill
            \includegraphics[width=.4875\linewidth]{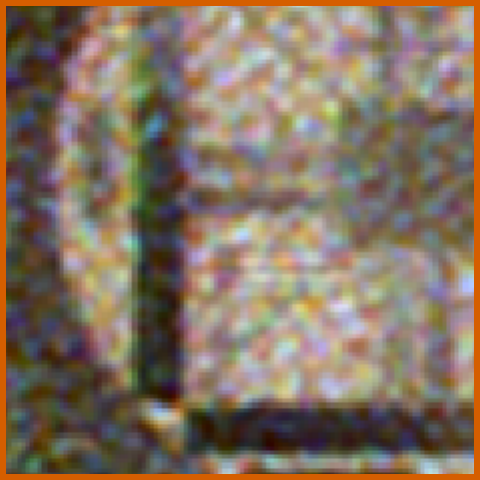}\\[1pt]
            {\scriptsize\mbox{UAFL~\cite{zhang2026uafl}}\\[-1pt]\mbox{31 bands}\par}
        \end{minipage}%
        \hfill
        \begin{minipage}[t]{.145\linewidth}
            \centering
            \includegraphics[width=\linewidth]{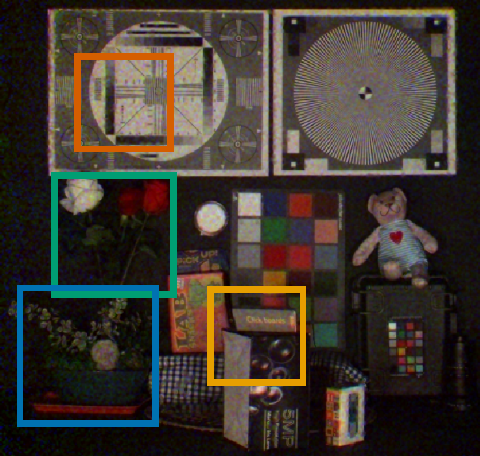}\\[1pt]
            \includegraphics[width=.4875\linewidth]{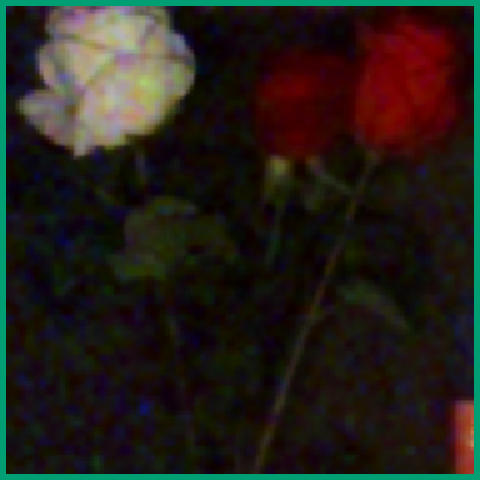}\hfill
            \includegraphics[width=.4875\linewidth]{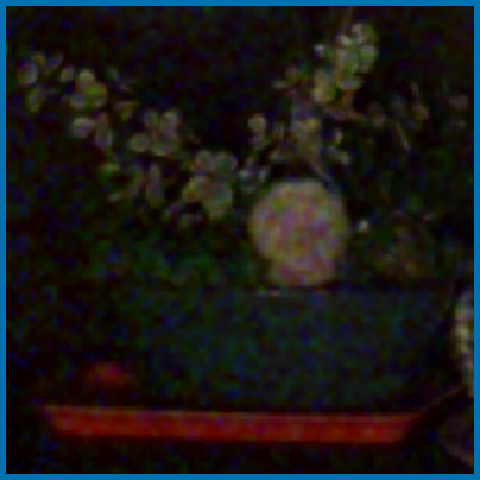}\\[1pt]
            \includegraphics[width=.4875\linewidth]{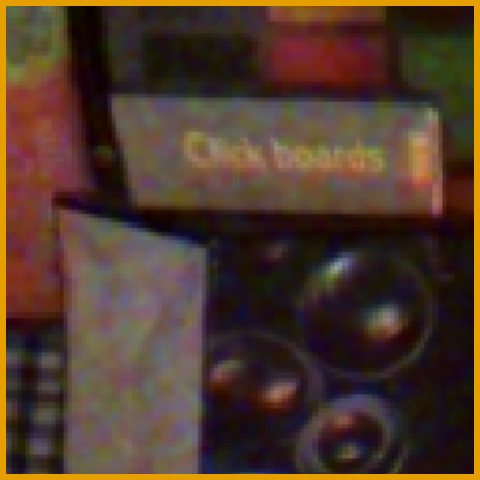}\hfill
            \includegraphics[width=.4875\linewidth]{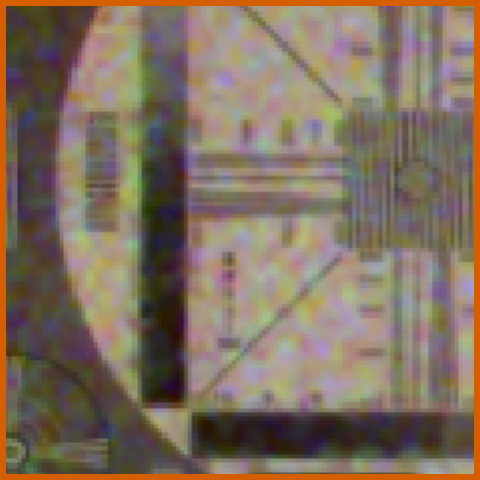}\\[1pt]
            {\scriptsize\mbox{CMF-GSA-wg~\cite{jouni2026bmvc}}\\[-1pt]\mbox{51 bands}\par}
        \end{minipage}%
        \hfill
        \begin{minipage}[t]{.145\linewidth}
            \centering
            \includegraphics[width=\linewidth]{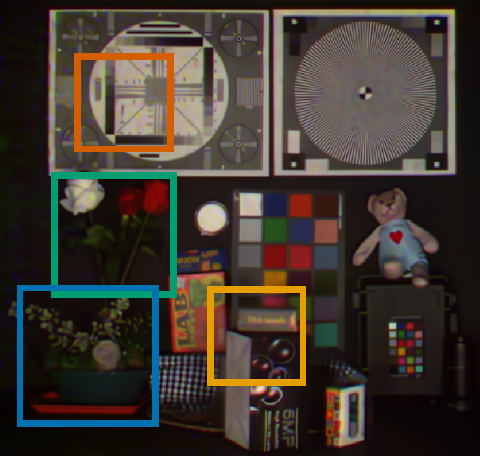}\\[1pt]
            \includegraphics[width=.4875\linewidth]{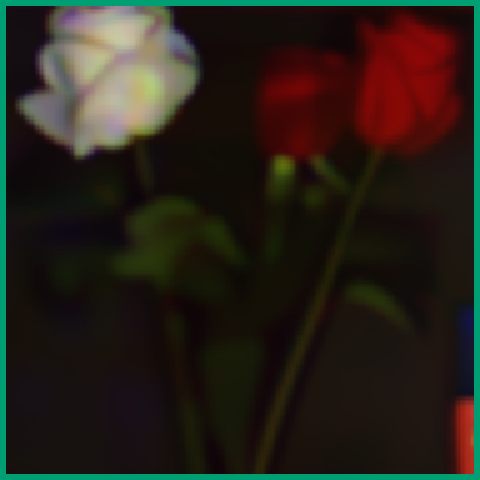}\hfill
            \includegraphics[width=.4875\linewidth]{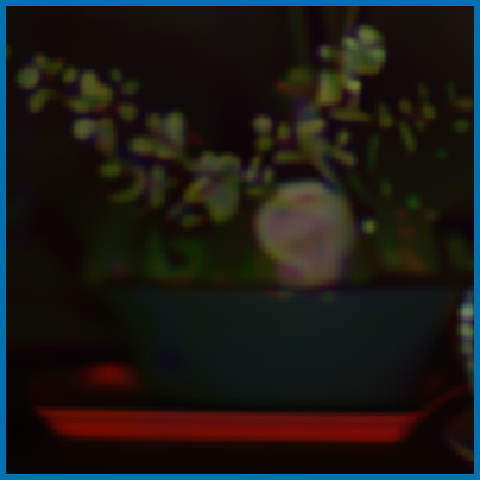}\\[1pt]
            \includegraphics[width=.4875\linewidth]{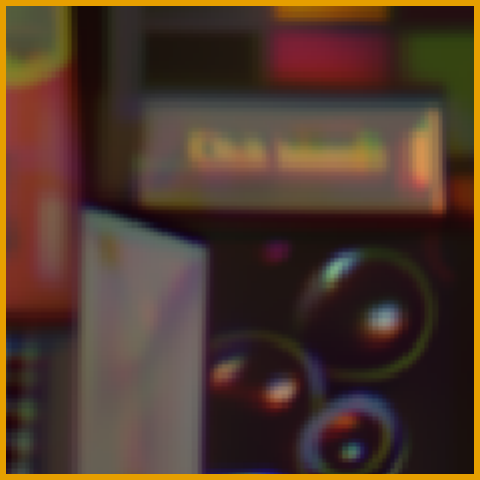}\hfill
            \includegraphics[width=.4875\linewidth]{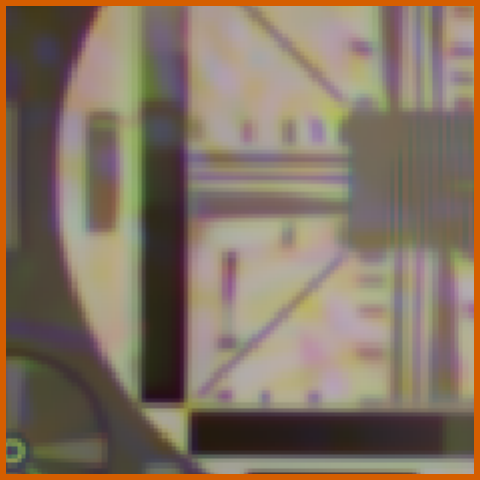}\\[1pt]
            {\scriptsize\mbox{SHARPnP (ours)}\\[-1pt]\mbox{51 bands}\par}
        \end{minipage}%
        \par
        \begin{minipage}[t]{.32\linewidth}
            \centering
            \begin{minipage}[c]{.48\linewidth}
                \includegraphics[width=\linewidth]{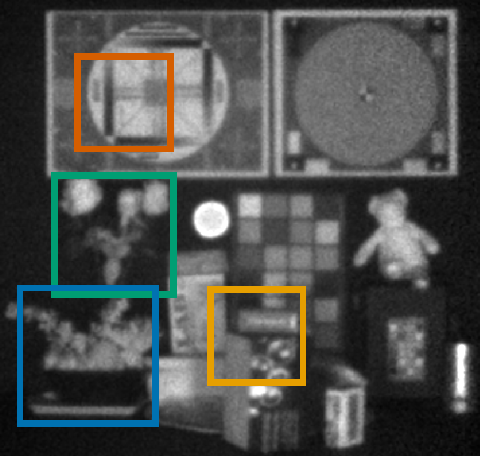}
            \end{minipage}
            \begin{minipage}[c]{.49\linewidth}
                \centering
                \includegraphics[width=.4875\linewidth]{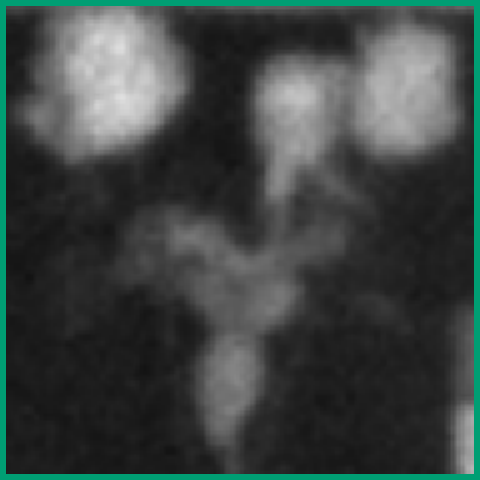}\hfill
                \includegraphics[width=.4875\linewidth]{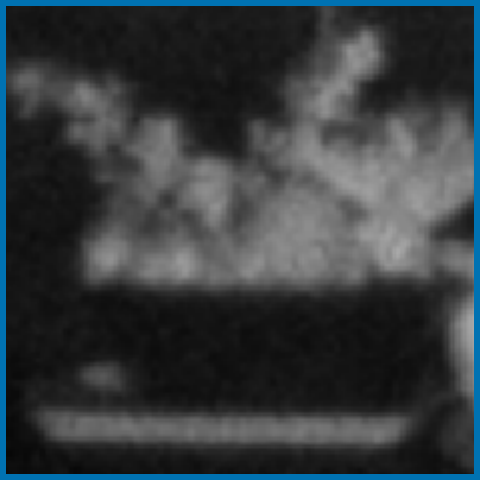}\\[1pt]
                \includegraphics[width=.4875\linewidth]{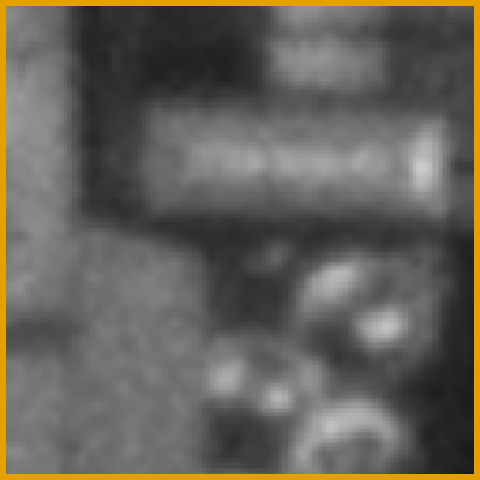}\hfill
                \includegraphics[width=.4875\linewidth]{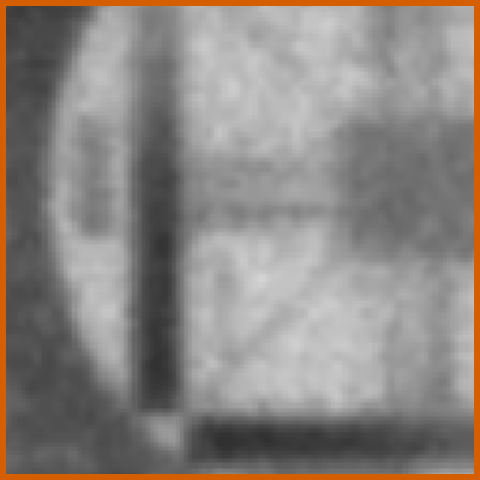}%
            \end{minipage}\\[1pt]
            {\scriptsize\mbox{Bicubic}\\[-1pt]\mbox{51 bands}\par}
        \end{minipage}%
        \hfill
        \begin{minipage}[t]{.32\linewidth}
            \centering
            \begin{minipage}[c]{.48\linewidth}
                \includegraphics[width=\linewidth]{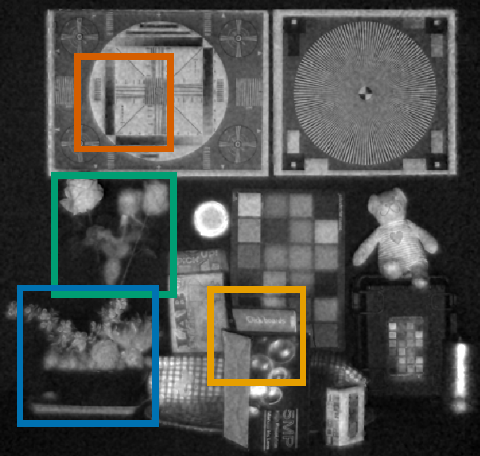}
            \end{minipage}
            \begin{minipage}[c]{.49\linewidth}
                \centering
                \includegraphics[width=.4875\linewidth]{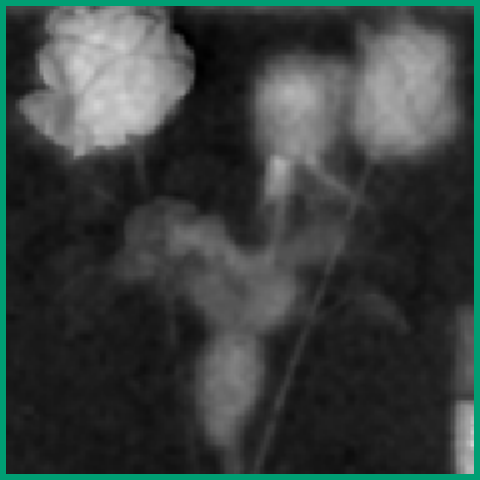}\hfill
                \includegraphics[width=.4875\linewidth]{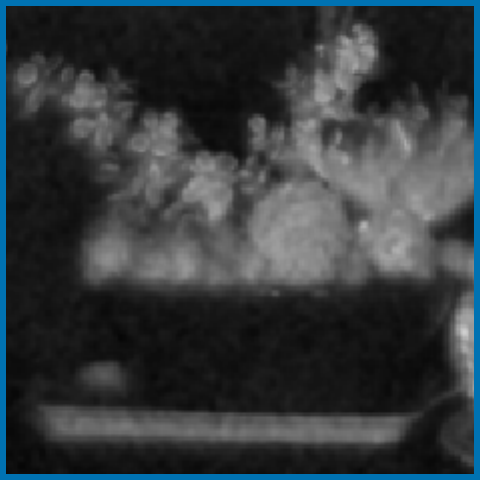}\\[1pt]
                \includegraphics[width=.4875\linewidth]{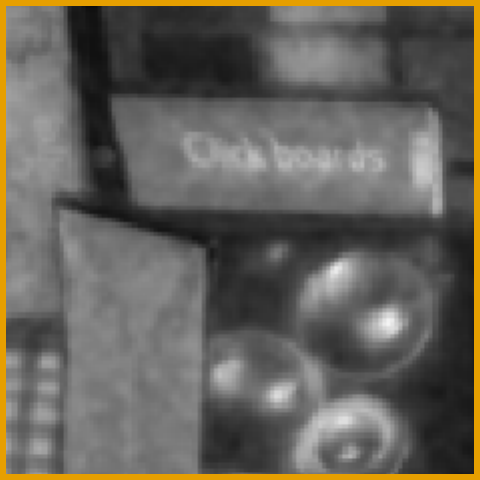}\hfill
                \includegraphics[width=.4875\linewidth]{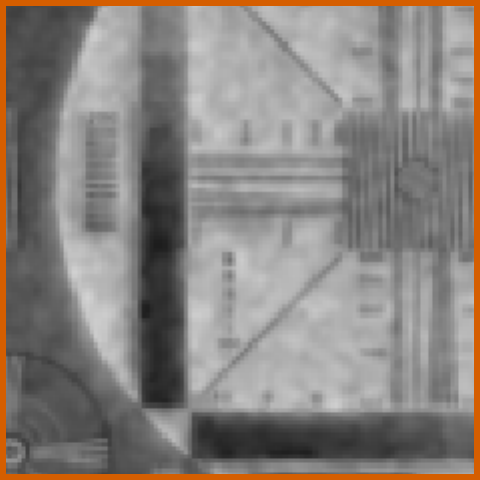}%
            \end{minipage}\\[1pt]
            {\scriptsize\mbox{CMF-GSA-wg~\cite{jouni2026bmvc}}\\[-1pt]\mbox{51 bands}\par}
        \end{minipage}%
        \hfill
        \begin{minipage}[t]{.32\linewidth}
            \centering
            \begin{minipage}[c]{.48\linewidth}
                \includegraphics[width=\linewidth]{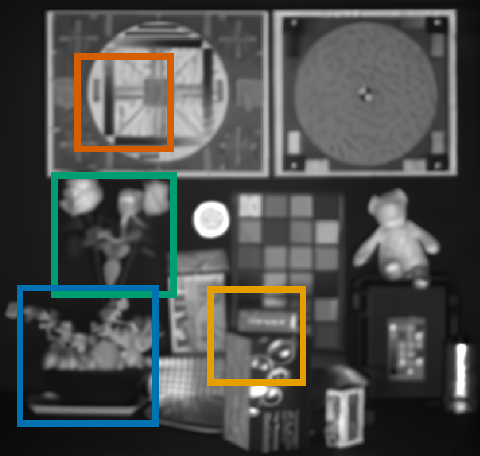}
            \end{minipage}
            \begin{minipage}[c]{.49\linewidth}
                \centering
                \includegraphics[width=.4875\linewidth]{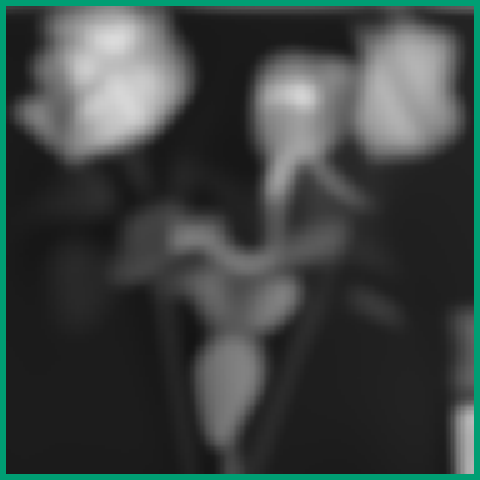}\hfill
                \includegraphics[width=.4875\linewidth]{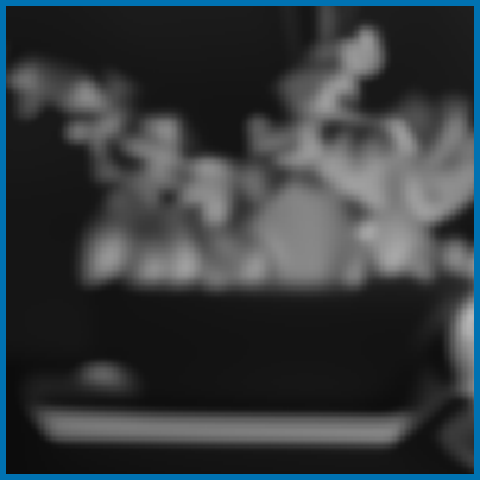}\\[1pt]
                \includegraphics[width=.4875\linewidth]{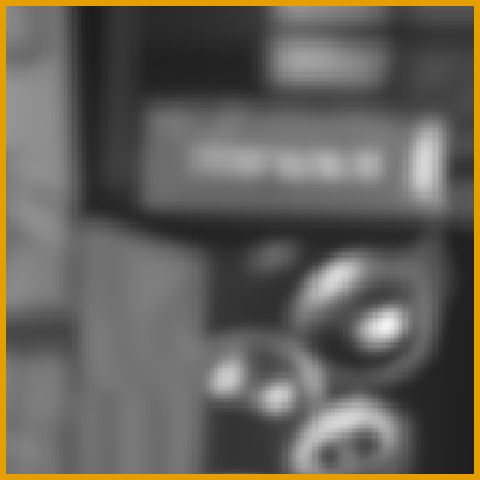}\hfill
                \includegraphics[width=.4875\linewidth]{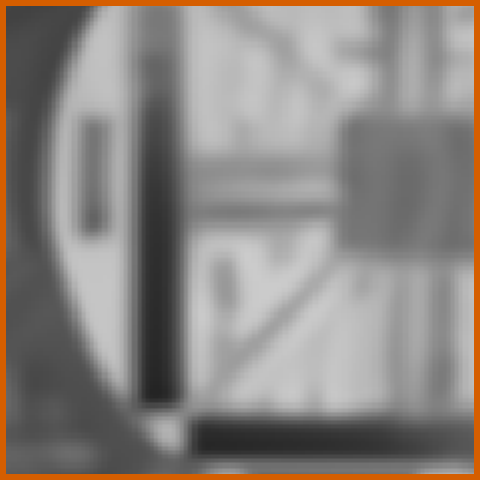}%
            \end{minipage}\\[1pt]
            {\scriptsize\mbox{SHARPnP (ours)}\\[-1pt]\mbox{51 bands}\par}
        \end{minipage}%
        \subcaption{30\,ms HSI exposure. Top: RGB composites and crops. Bottom: NIR at 802\,nm and crops.}
        \label{fig:november-parallax_30ms}
    \end{minipage}
    \vspace{2mm}

    \begin{minipage}[t]{.24\textwidth}
        \centering
        \includegraphics[width=\linewidth]{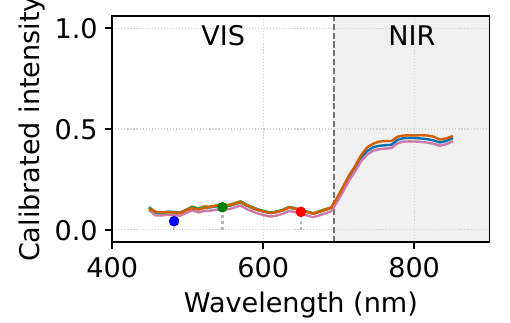}\\[1pt]
        {\scriptsize Real flower leaf}
    \end{minipage}%
    \hfill
    \begin{minipage}[t]{.24\textwidth}
        \centering
        \includegraphics[width=\linewidth]{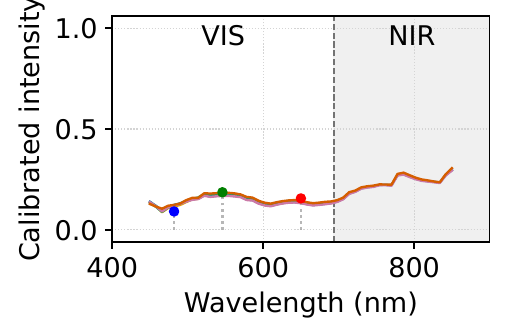}\\[1pt]
        {\scriptsize Artificial leaf}
    \end{minipage}%
    \hfill
    \begin{minipage}[t]{.24\textwidth}
        \centering
        \includegraphics[width=\linewidth]{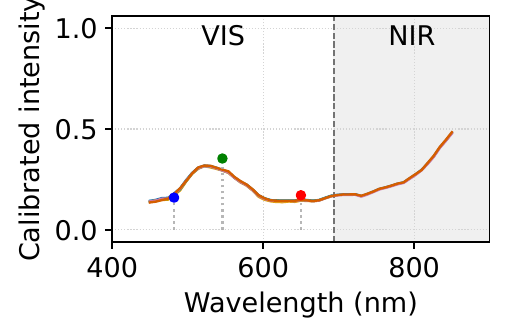}\\[1pt]
        {\scriptsize Green patch}
    \end{minipage}%
    \hfill
    \begin{minipage}[t]{.24\textwidth}
        \centering
        \includegraphics[width=\linewidth]{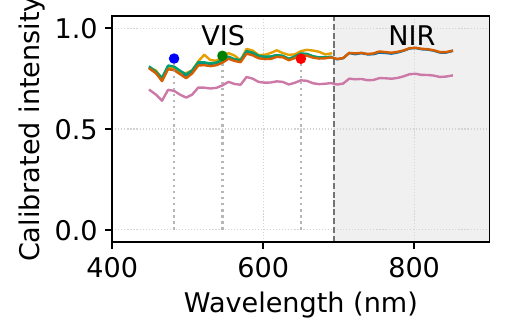}\\[1pt]
        {\scriptsize Spectralon}
    \end{minipage}%
    \par\vspace{1pt}
    \includegraphics[width=\textwidth]{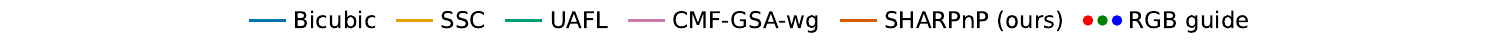}
    \caption{Native Ultris-RPi reconstruction at 480 and 30\,ms exposure. RGB columns: Bicubic, RGB guide, SSC~\cite{zhang2025sschs}, UAFL~\cite{zhang2026uafl}, CMF-GSA-wg~\cite{jouni2026bmvc}, and SHARPnP (ours). NIR shows Bicubic, CMF-GSA-wg and SHARPnP, available only for 51 bands reconstruction. Crops show real/fake roses (petals and leaves), real cactus, parallax region, and a spatial target. No clean HR reference is available.}
    \label{fig:november-comparison}
\end{figure*}

\paragraph{Native configuration.}
We apply $4\times$ super-resolution using 20 HQS iterations, frozen DRUNet and $P_D=I_{51}$. Directional VTV uses $\mu=10^{-5}$; conjugate gradients use tolerance $10^{-5}$. The RGB value/gradient coefficients are $(0.304,0.0909)$ at 30\,ms and $(0.0297,0.0888)$ at 480\,ms.

\Cref{fig:november-comparison} compares the native reconstructions. The 480\,ms acquisition provides stronger signal, although still noisy, whereas the 30\,ms video demonstrates the deterioration caused by low-SNR snapshot imaging. SSC and UAFL propagate measurement noise and lose fine target structure. CMF-GSA-wg transfers RGB details but retains visible noise and locally smooth regions. SHARPnP jointly suppresses noise and recovers target markings and object boundaries, including at the shorter exposure up to some limitations. It reconstructs all 51 VIS-NIR bands with the same frozen denoiser, whereas the transferred neural baselines operate on the 31-band VIS subset. The material spectra show distinct VIS-NIR responses without individual normalization. Since no clean HR HSI reference is available, these native results are evaluated qualitatively.

In the NIR domain, SHARPnP applies super-resolution to a limited amount, but is also careful about injecting the details from the wrong channels. For example, the top left crop shows three roses, two of which are artificial while the middle one is real, where its stem and leaves show higher reflectance in the NIR domain. This is observed in the crop of the NIR, where SHARPnP reconstructs the stem of the real flower, while CMF-GSA-wg projects over all three stems. Another interesting object is the black objective lens on the right side of the scene, which is dark in the RGB preview but is reflective in the NIR preview, showing again that SHARPnP is capable of conserving NIR signatures up to some super-resolution limitation.




\section{Conclusions}
We presented SHARPnP, a self-supervised plug-and-play framework for RGB-guided
hyperspectral super-resolution from noisy, unaligned camera measurements.
SHARPnP needs no high-resolution hyperspectral references. It adapts a
pretrained registration model through reconstruction feedback and fuses
calibrated RGB guidance with a frozen denoising prior. It sets a new state of
the art on REAL dataset, surpassing supervised fusion methods at all scales and
reaching 43.37\,dB at $4\times$ versus 41.91\,dB for the strongest baseline.
Self-supervised fine-tuning on noisy pairs extends it to noisy inputs, where it
reaches 39.33\,dB at $\sigma = 5/255$, nearly 7\,dB above direct supervised
fusion. On a native snapshot rig, it reconstructs all 51 VIS--NIR bands. A relevant finding of this study is that NIR information contributes to the reconstruction of spatial details, even in regions where RGB guidance is limited. This observation motivates future work on improving spectral consistency in the NIR range, ensuring that reconstructed details remain faithful to the underlying spectral information. Additionally, we aim to extend SHARPnP to real-time hyperspectral video super-resolution by accelerating its test-time optimization, enabling more efficient reconstruction of dynamic scenes and moving objects.

\clearpage
\appendix
\section*{Supplementary Material}
\setcounter{figure}{0}
\setcounter{table}{0}
\setcounter{equation}{0}
\renewcommand{\thefigure}{S\arabic{figure}}
\renewcommand{\thetable}{S\arabic{table}}
\renewcommand{\theequation}{S\arabic{equation}}
\renewcommand{\theHfigure}{supp.\arabic{figure}}
\renewcommand{\theHtable}{supp.\arabic{table}}
\renewcommand{\theHequation}{supp.\arabic{equation}}
This supplementary material is organized into four main sections: (1) additional implementation details, including the registration process and reconstruction settings, (2) additional ablation studies,
(3) additional REAL comparisons, and finally,
%
(4) qualitative and quantitative results on native RGB-HSI data, where we revisit the scenes from Section~5.3 of the main paper at 480\,ms and 30\,ms, and evaluate a second scene previously used by Jouni et al.~\cite{jouni2026bmvc}.%
%
The code implementation will be available as open source.  

\section{Additional Implementation Details}
\subsection{Registration fine-tuning settings}
\label{sec:supp-training}
Fine-tuning updates the selected CrossRAFT parameters $\phi$, while the denoiser remains frozen. We use AdamW with learning rate $10^{-5}$, weight decay $10^{-6}$, gradient clipping at one and seed 2022. REAL fine-tuning uses 470 updates and matching $256\times256$ crops. Batch-normalization statistics remain fixed.
Adapter fine-tuning freezes the backbone (RAFT); backbone fine-tuning freezes the adapter.

\paragraph{Noisy pairs.}
(1) \textit{REAL benchmark data:} For noisy inputs, GR2R in the main paper uses $\tau\simeq0.42$. For clean inputs, fine-tuning uses the same observation in both views.
(2) \textit{Our native RGB-HSI data:} Noise2Noise pairs use adjacent frames from a video acquisition of a static scene under a given noise configuration.
In Section 5.3 of the main paper, fine-tuning is performed separately for the 480 ms and 30 ms exposures.
For a dynamic scene, finetuning can first be performed on a static scene acquisition acquired under the same acquisition settings, then transferred to frame-by-frame reconstruction of the dynamic sequence.

\subsection{Reconstruction settings}

\paragraph{Confidence-weighted directional VTV (dVTV).}
We use a confidence-weighted form of directional TV~\cite{bungert2018directional}, coupling the spatial gradients across spectral bands through vectorial TV~\cite{bresson2008fast}.
It reduces smoothing across reliable edges of the registered RGB guide while regularizing spatial variations jointly across spectral bands.

Denote by $X\in\R^{n\times B}$ the reconstructed HSI where $n$ and $B$ represent the output pixels and spectral bands respectively, $L\in\R^{n\times1}$ the grayscale image of the registered RGB guide $R_w\in\R^{n\times3}$,
$C_P$ the confidence map,
and
$\nabla X(p)\in\R^{2\times B}$ the horizontal and vertical gradients of all bands at pixel $p \in \{1, \dots, n\}$.
We define the dVTV regularizer $\mathcal V(X;R_w,C_P)$ as:
\begin{equation}
 \xi_p = \frac{\eta\sqrt{C_P(p)} \;\;\; \nabla L(p)}{\sqrt{\|\nabla L(p)\|_2^2+\delta^2}},
 \; H_p=I_2-\xi_p\xi_p^\top.
\end{equation}
\begin{equation}
	\mathcal V(X;R_w,C_P)
	=\frac{1}{n\sqrt B}\sum_p
	\|H_p\nabla X(p)\|_F,
\end{equation}
where
$\eta$ is the direction strength,
$\delta$ is a gradient stabilizer that prevents division by zero~\cite{bungert2018directional},
$\xi_p\in\R^{2\times1}$ is a confidence-weighted direction vector,
$I_2\in\R^{2\times2}$ is the $2\times2$ identity matrix,
$H_p\in\R^{2\times2}$ is the directional matrix that acts on the horizontal and vertical gradient components,
and
$\|\cdot\|_F$ denotes the Frobenius norm.
We use $\eta=0.99$ and $\delta=0.01$.
For numerical stability, we smooth the TV penalty using $\epsilon=0.001$ to avoid division by zero in uniform regions~\cite{blomgren1998color}.

\begin{table}[b]
    \centering
    \caption{SHARPnP component ablation on clean REAL at three scales.
    Each block reports ten-scene means at iteration 50 without retuning.
    Bold indicates the best unrounded score within each block.}
    \label{tab:sharp_registration_ablation}
    \resizebox{\columnwidth}{!}{%
        \begin{tabular}{clrrrr}
            \hline
            Scale & Variant & PSNR$\uparrow$ & SSIM$\uparrow$ & SAM$\downarrow$ & RMSE$\downarrow$ \\
                  & & (dB)          &                & ($^\circ$)      & ($10^{-2}$) \\
            \hline
            \multirow{4}{*}{\rotatebox[origin=c]{90}{\footnotesize REAL $4\times$}} & SHARPnP                     & \textbf{43.35} & \textbf{0.991} & \textbf{1.596} & \textbf{0.7736} \\
             & w/o CrossRAFT Adaptation    & 43.30          & 0.991          & 1.596          & 0.7780 \\
             & w/o Radiometric Calibration & 42.71          & 0.989          & 1.689          & 1.0806 \\
             & w/o Residual Alignment      & 41.26          & 0.983          & 1.726          & 1.1240 \\
            \hline
            \multirow{4}{*}{\rotatebox[origin=c]{90}{\footnotesize REAL $8\times$}} & SHARPnP                     & 37.37          & 0.974          & \textbf{2.278} & 1.6437 \\
             & w/o CrossRAFT Adaptation    & \textbf{37.51} & \textbf{0.974} & 2.283          & \textbf{1.6178} \\
             & w/o Radiometric Calibration & 35.01          & 0.950          & 2.728          & 2.3267 \\
             & w/o Residual Alignment      & 30.78          & 0.884          & 3.443          & 4.0326 \\
            \hline
            \multirow{4}{*}{\rotatebox[origin=c]{90}{\footnotesize REAL $16\times$}} & SHARPnP                     & \textbf{32.37} & \textbf{0.947} & \textbf{3.126} & \textbf{2.9777} \\
             & w/o CrossRAFT Adaptation    & 31.97          & 0.942          & 3.198          & 3.1175 \\
             & w/o Radiometric Calibration & 28.96          & 0.887          & 5.084          & 4.9438 \\
             & w/o Residual Alignment      & 30.87          & 0.920          & 3.354          & 3.6917 \\
            \hline
        \end{tabular}%
    }
\end{table}

\paragraph{Denoising configuration.}
For denoising, we use the frozen RGB DRUNet configuration in HQS with the DPIR~\cite{zhang2022dpir} parameterization~\cite{picone2026icip}.
For clean REAL reconstruction, we reduce the strength of the denoising correction in later iterations. For noisy inputs, we retain the full correction.
For our native RGB-HSI data, the denoiser noise level is fixed from temporal noise estimates at each exposure.

\begin{figure*}[!b]
    \centering
    \begin{minipage}{\textwidth}
        \centering
        \includegraphics[width=\linewidth]{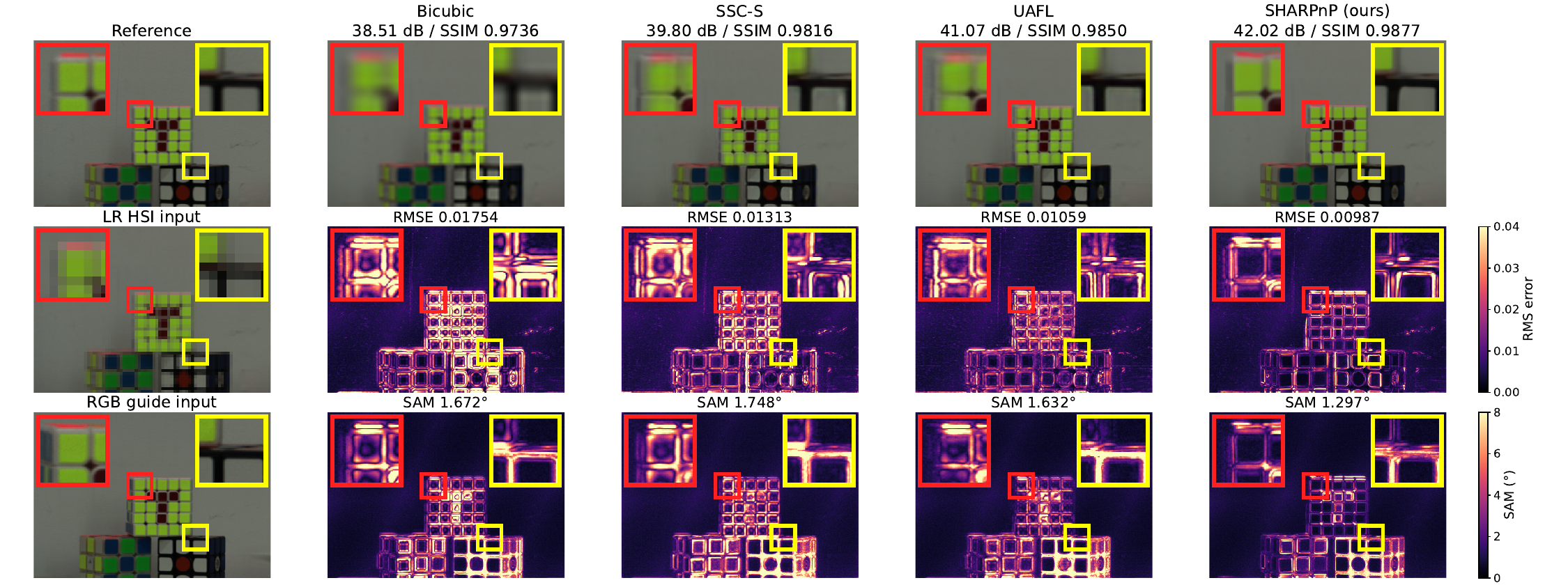}
        \subcaption{REAL scene 59; clean; $8\times$}
    \end{minipage}\\[2pt]
    \begin{minipage}{\textwidth}
        \centering
        \includegraphics[width=\linewidth]{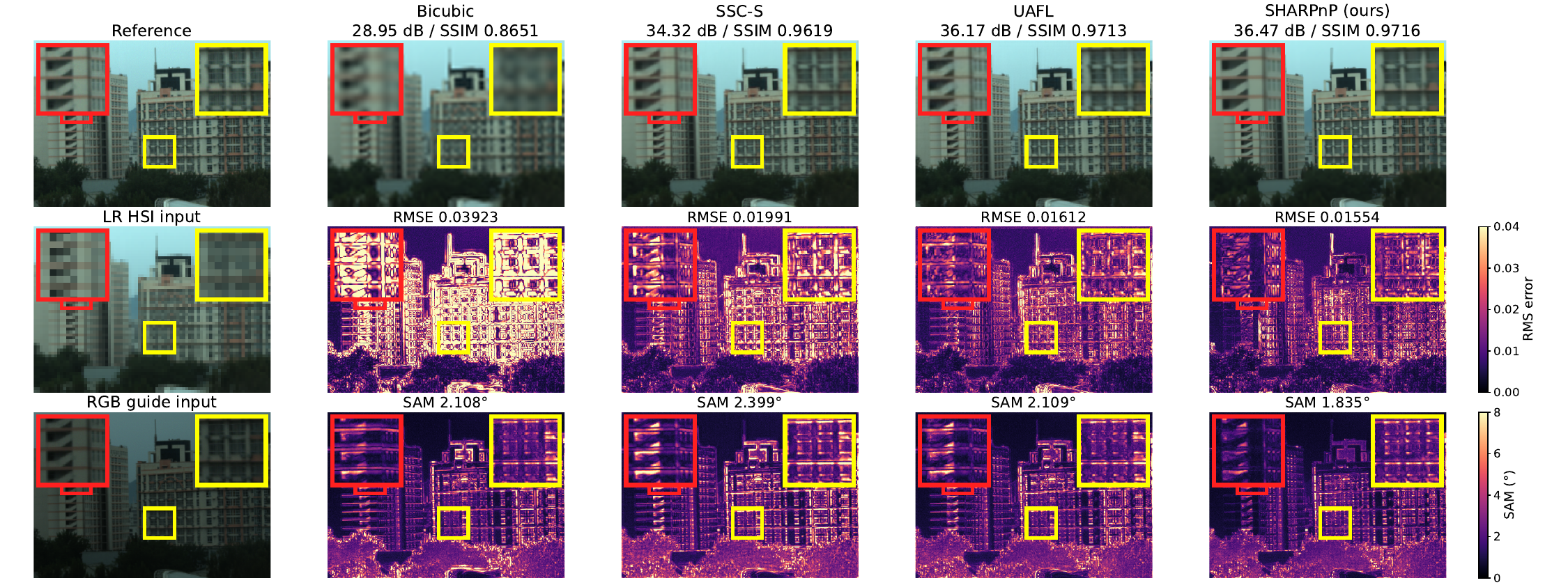}
        \subcaption{REAL scene 35; clean; $8\times$}
    \end{minipage} \vspace{-0.5em}
    \caption{Clean REAL comparisons at $8\times$ with Bicubic, SSC-S~\cite{zhang2025sschs}, UAFL~\cite{zhang2026uafl}, and SHARPnP (ours). Rows show RGB previews, RMS error and SAM maps against the clean HR reference. Insets share the same regions and scales across methods. Annotations are full-frame metrics computed from the displayed predictions.} \vspace{-1em}
    \label{fig:supp-real}
\end{figure*}

\section{Additional Ablations}

\paragraph{RGB Registration Ablation.}
Table~\ref{tab:sharp_registration_ablation} evaluates three components
of SHARPnP on ten clean REAL scenes at $4\times$, $8\times$, and
$16\times$, using iteration 50 without retuning. Removing residual
alignment reduces PSNR by 2.09, 6.59, and 1.50~dB, respectively.
Replacing the fitted radiometric map with a fixed three-band proxy,
while preserving geometry and effective RGB weights, causes drops
of 0.64, 2.36, and 3.41~dB. Both components therefore provide substantial
benefits across all tested scales. CrossRAFT fine-tuning has a smaller,
scale-dependent effect: it improves PSNR by 0.05~dB at $4\times$ and
0.40~dB at $16\times$, but reduces it by 0.14~dB at $8\times$, while
slightly improving SAM. Thus, task-specific fine-tuning is not uniformly
necessary when released weights already provide effective initial
correspondences. These findings support prioritizing radiometric
calibration and residual alignment, while treating frontend fine-tuning
as a setting-dependent refinement. Its improvement across all four
metrics at $16\times$ demonstrates its value in this challenging regime.

\section{Additional Results in REAL dataset}
\begin{figure*}
    \centering
    \begin{minipage}{\textwidth}
        \centering
        \includegraphics[width=\linewidth]{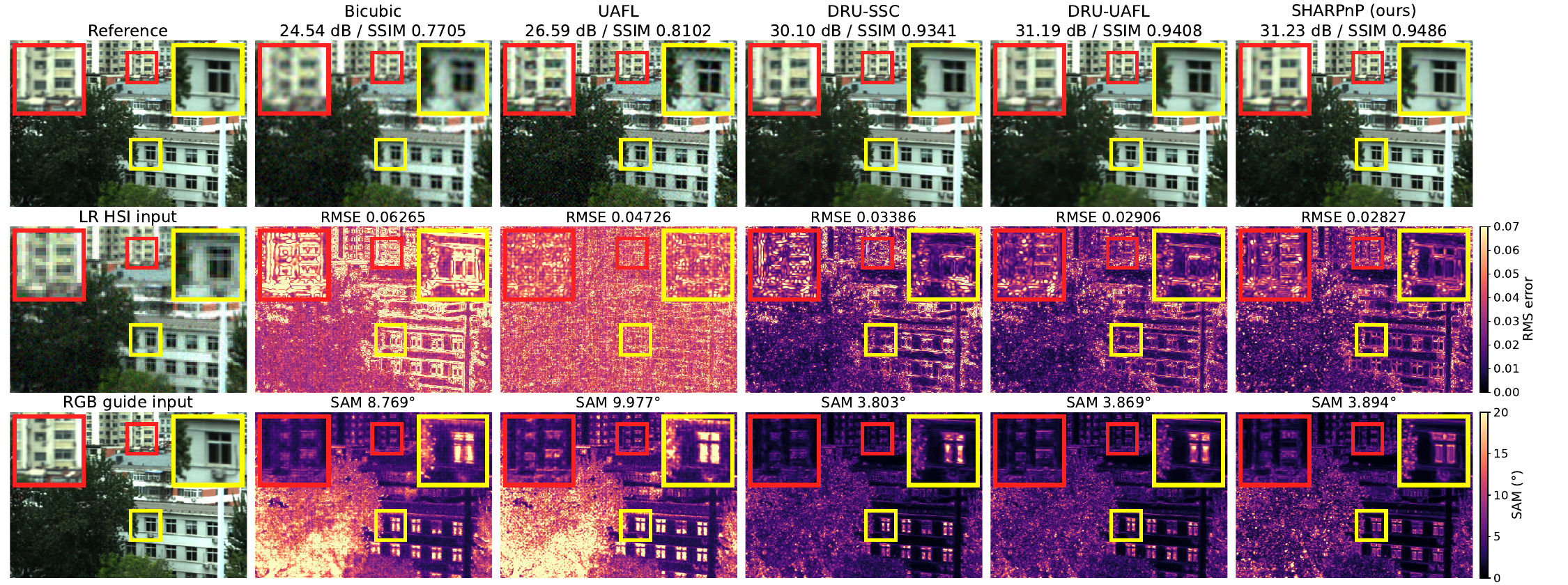}
        \subcaption{REAL scene 0; $10/255$ LR-HSI noise; $4\times$}
    \end{minipage}\\[2pt]
    \begin{minipage}{\textwidth}
        \centering
        \includegraphics[width=\linewidth]{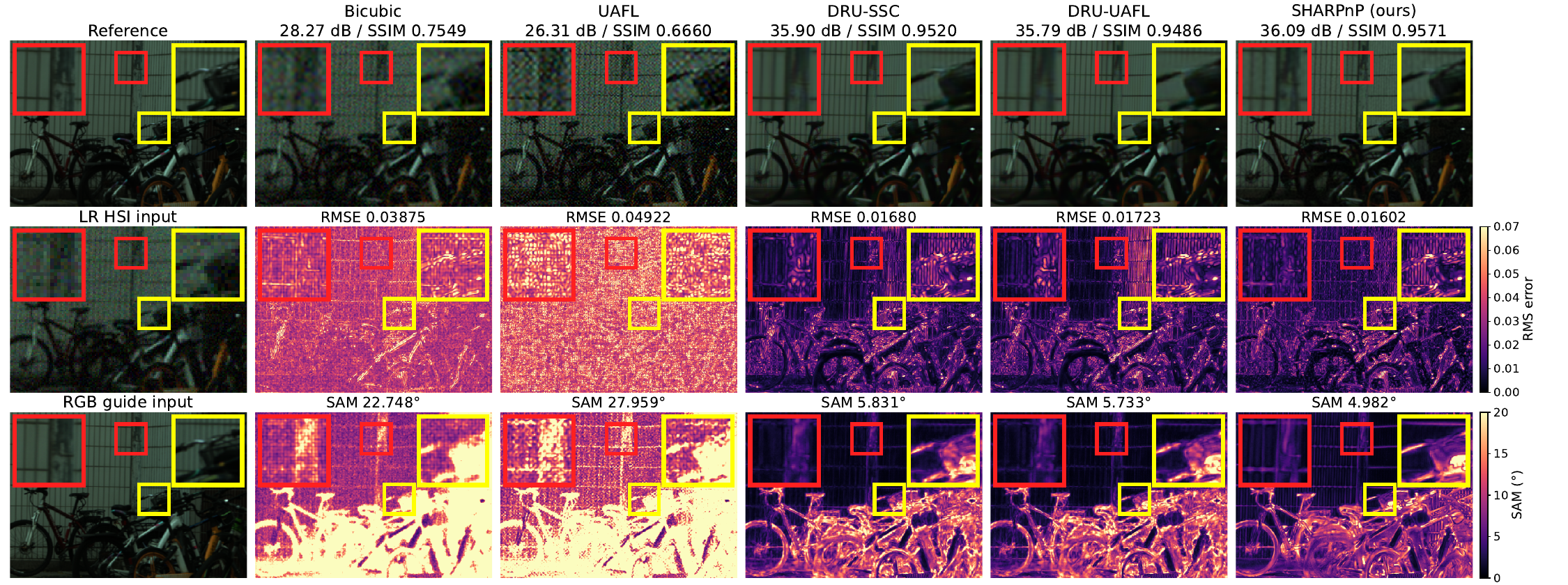}
        \subcaption{REAL scene 31; $10/255$ LR-HSI noise; $4\times$}
    \end{minipage}
    \caption{Noisy REAL comparisons at $4\times$ and $10/255$ LR-HSI noise with Bicubic, UAFL~\cite{zhang2026uafl}, DRU-SSC, DRU-UAFL, and SHARPnP (ours). UAFL receives the noisy LR-HSI directly. DRU denotes sequential denoise-then-fuse, i.e., DRUNet denoising followed by the frozen supervised fusion model. SHARPnP improves all four full-frame metrics in scene 31, while scene 0 shows a small SAM trade-off despite improved PSNR, SSIM and RMSE. Rows show RGB previews, RMS error and SAM maps against the clean HR reference.}
    \label{fig:supp-real-noisy}
\end{figure*}

\paragraph{Clean x8.}
Figure~\ref{fig:supp-real} complements the main paper with two clean $8\times$ examples, a more stressful case than $4\times$. Scene 59 shows close-range Rubik's cubes from two viewpoints, with some faces visible in only one camera view. SHARPnP preserves smooth regions better while reducing errors along the edges. Scene 35 shows fine outdoor details such as balconies and windows, which are blurred in the LR-HSI input. SHARPnP gives the best scores on all four metrics.

\paragraph{Noisy 10/255 x4.}
Figure~\ref{fig:supp-real-noisy} shows two $4\times$ examples with $10/255$ noise added to the low-resolution HSI.
Scene 0 revisits the main paper's clean example under stronger $10/255$ noise, while scene 31 provides an additional example.
We compare bicubic, direct UAFL~\cite{zhang2026uafl} to show the effect of applying supervised fusion directly to noisy inputs, our denoise-then-fuse baselines using the same DRUNet denoiser used in SHARPnP, and SHARPnP itself.  Scene 0 shows building windows and dark vegetation. Direct UAFL retains visible noise, while denoising before fusion substantially improves the reconstruction. SHARPnP gives the best PSNR, SSIM, and RMSE, with a slightly higher SAM than the DRU-baselines. Scene 31 shows smooth wall tiles alongside thin bicycle structures. SHARPnP preserves the tile boundaries while reducing noise in the darker bicycle regions. It gives the best scores on all four metrics.


\section{Additional Results in Native RGB-HSI data}

In this section, we expand on the real native RGB-HSI experimental section by a quantitative analysis and an additional real native scene.
In Section \ref{subsec:two_exposures},
we revisit the scenes from Section 5.3 of the main paper at the two exposure times, 480 ms and 30 ms, and add a quantitative analysis of their results with no clean high-resolution reference.
In Section \ref{subsec:bmvc_scene}, we add results on a different real native scene, which was used in \cite{jouni2026bmvc}, while considering the same qualitative and quantitative analysis style as our data from Section 5.3 of the main paper.

Neither acquisition provides a clean high-resolution HSI reference.
On the high-resolution grid, we rely on the controlled spatial and spectral targets that are present in the different scenes in order to inspect how well each method recovers spatial and spectral details in the final super-resolved outputs.
As for reported full-image metrics, the analysis is performed on the low-resolution HSI grid.
In particular, we blur and downsample each super-resolved output to the observed low-resolution HSI grid and compare it with temporally averaged HSI observations of the same scene. This is done in order to reduce the effect of noise of a single observed HSI frame and focus more on spatial and spectral structures.

We report RMSE and SAM separately over the 31 VIS bands and the 20 NIR bands. We also report our so-called neighborhood SAM (\textit{N-SAM}) and Spectralon variation (\textit{S-Var}).
For \textit{N-SAM}, we select a small neighborhood (patch) inside each ColorChecker square used in the experiment and one on the white Spectralon surface. We then average the spectra in each neighborhood and compute the angle between the measured and reconstructed visible spectra, giving one angle per neighborhood. We report the median of these angles.
For \textit{S-Var}, we select a uniform neighborhood on the white Spectralon surface, which is expected to be flat and smooth. We compute the standard deviation of the pixel intensities in this neighborhood to measure how much the reconstructed white surface varies from pixel to pixel.

\subsection{Different exposure times}
\label{subsec:two_exposures}

\begin{figure}[tbp]
    \centering
    \begin{subfigure}[t]{.49\linewidth}
        \centering
        \includegraphics[width=\linewidth]{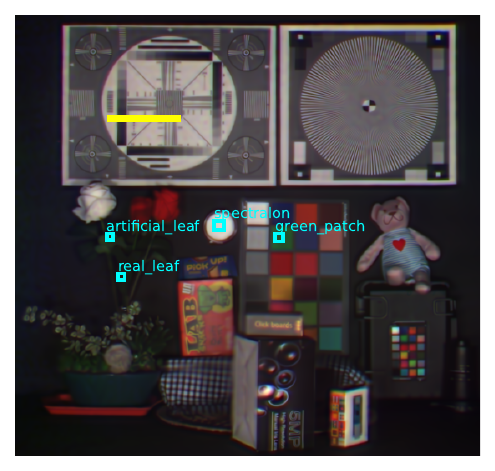}
        \caption{480 ms}
    \end{subfigure}\hfill%
    \begin{subfigure}[t]{.49\linewidth}
        \centering
        \includegraphics[width=\linewidth]{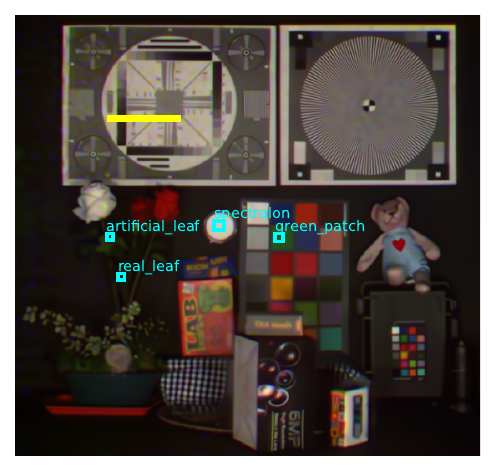}
        \caption{30 ms}
    \end{subfigure}\par
    \caption{Yellow marks the strip used for the target profiles. Cyan marks the four regions used for the spectra.}
    \label{fig:supp-native-regions}
\end{figure}

\begin{figure}[tbp]
    \centering
    \begin{subfigure}[t]{.49\linewidth}
        \centering
        \includegraphics[width=\linewidth]{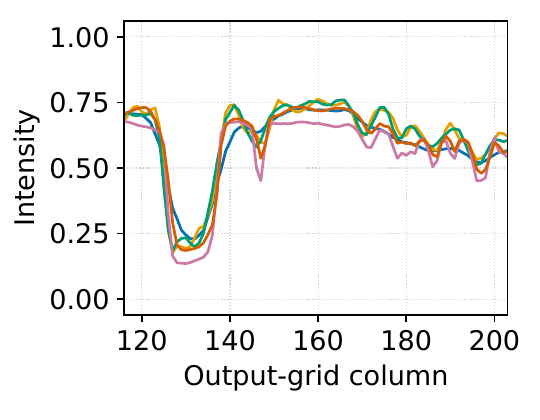}
        \caption{480 ms, 650 nm}
    \end{subfigure}\hfill%
    \begin{subfigure}[t]{.49\linewidth}
        \centering
        \includegraphics[width=\linewidth]{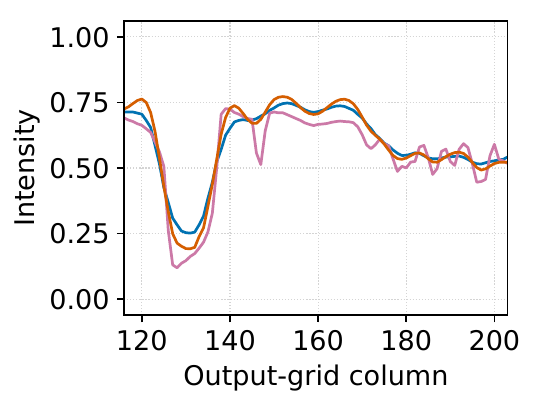}
        \caption{480 ms, 802 nm}
    \end{subfigure}\par
    \begin{subfigure}[t]{.49\linewidth}
        \centering
        \includegraphics[width=\linewidth]{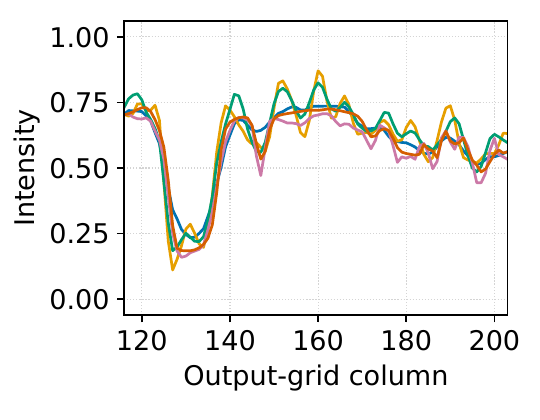}
        \caption{30 ms, 650 nm}
    \end{subfigure}\hfill%
    \begin{subfigure}[t]{.49\linewidth}
        \centering
        \includegraphics[width=\linewidth]{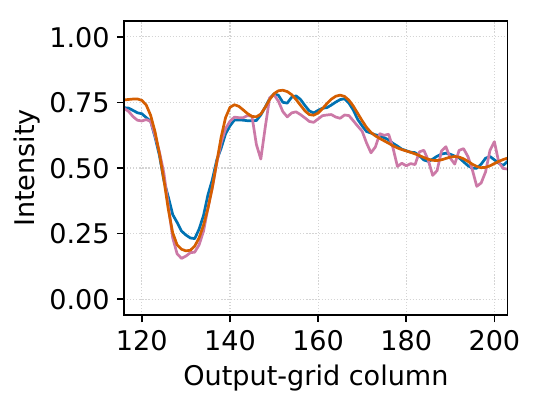}
        \caption{30 ms, 802 nm}
    \end{subfigure}\par
    \includegraphics[width=\linewidth]{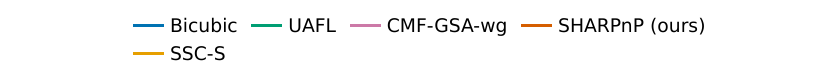} \vspace{-2.5em}
    \caption{Intensity across the same target at both exposures. Each curve averages four image rows.}
    \label{fig:supp-native-profiles}
\end{figure}

\begin{figure}[tbp]
    \centering
    \begin{subfigure}[t]{.49\linewidth}
        \centering
        \includegraphics[width=\linewidth]{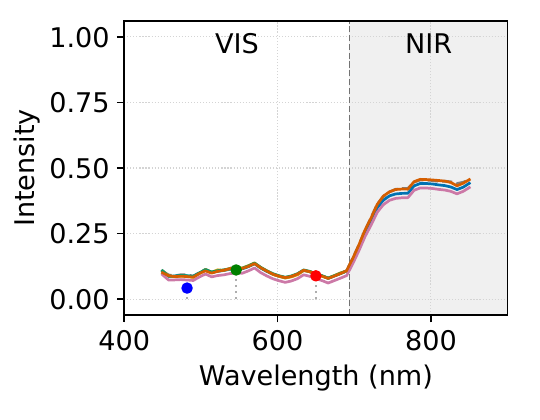}
        \caption{Real leaf}
    \end{subfigure}\hfill%
    \begin{subfigure}[t]{.49\linewidth}
        \centering
        \includegraphics[width=\linewidth]{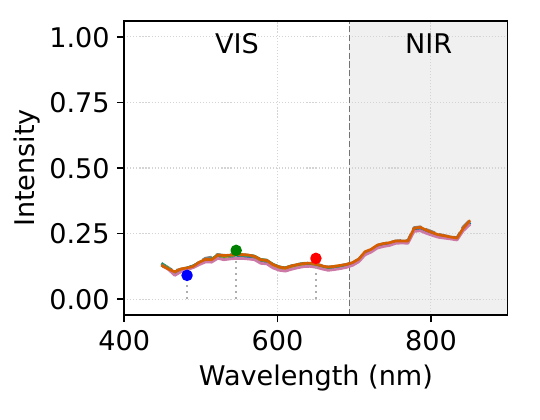}
        \caption{Artificial leaf}
    \end{subfigure}\par
    \begin{subfigure}[t]{.49\linewidth}
        \centering
        \includegraphics[width=\linewidth]{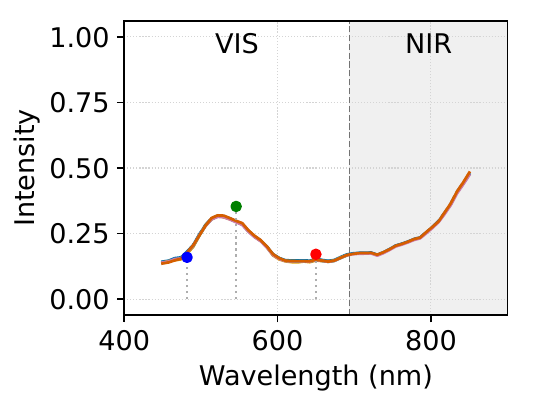}
        \caption{Green patch}
    \end{subfigure}\hfill%
    \begin{subfigure}[t]{.49\linewidth}
        \centering
        \includegraphics[width=\linewidth]{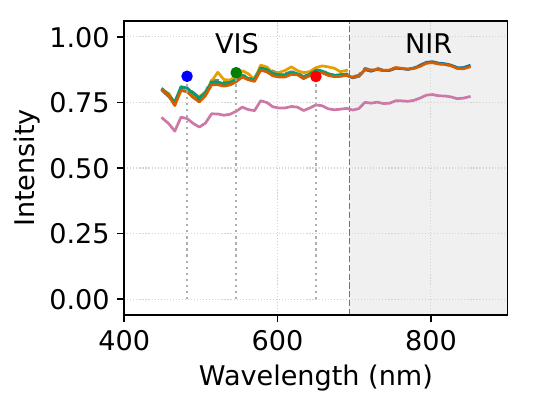}
        \caption{White reference}
    \end{subfigure}\par
    \includegraphics[width=\linewidth]{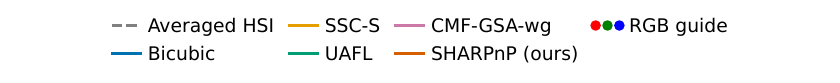} \vspace{-2em}
    \caption{Patch spectra at 480 ms. The dashed line shows the average of ten HSI frames.} \vspace{-1em}
    \label{fig:supp-native-spectra}
\end{figure}

\begin{table}[tbp]
    \centering
    \caption{Static scene, 480 ms exposure.}
    \label{tab:supp-native-consistency}
    \setlength{\tabcolsep}{1.75pt}
    \footnotesize
    \begin{tabular*}{\linewidth}{@{\extracolsep{\fill}}lrrrrrr@{}}
        \toprule
         & \multicolumn{2}{c}{VIS} & \multicolumn{2}{c}{NIR} & \multicolumn{2}{c}{VIS patches} \\
         & \multicolumn{2}{c}{$[450,690]$~nm} & \multicolumn{2}{c}{$[698,850]$~nm} & \multicolumn{2}{c}{} \\
        \cmidrule(lr){2-3}\cmidrule(lr){4-5}\cmidrule(l){6-7}
        Method & RMSE$\downarrow$ & SAM$\downarrow$ & RMSE$\downarrow$ & SAM$\downarrow$ & N-SAM$\downarrow$ & S-Var$\downarrow$ \\
               & ($10^{-2}$) & ($^\circ$) & ($10^{-2}$) & ($^\circ$) & ($^\circ$) & ($10^{-2}$) \\
        \midrule
        Bicubic & 1.891 & 1.879 & 2.087 & 1.182 & 0.711 & 1.431 \\
        SSC-S~\cite{zhang2025sschs} & 2.240 & 2.298 & -- & -- & 1.330 & 2.424 \\
        UAFL~\cite{zhang2026uafl} & 2.131 & 1.972 & -- & -- & 0.882 & 2.137 \\
        CMF-GSA~\cite{jouni2026bmvc} & 2.316 & 2.106 & 2.544 & 1.287 & 1.069 & \textbf{1.209} \\
        SHARPnP & \textbf{0.815} & \textbf{1.580} & \textbf{0.787} & \textbf{0.861} & \textbf{0.702} & 1.388 \\
        \bottomrule
    \end{tabular*}
\end{table}

We revisit the static scene from Section 5.3 of the main paper at 480 ms and 30 ms. At each exposure, we average ten additional HSI frames. Here N-SAM uses the 20 apparent ColorChecker patches and the Spectralon.
Figure~\ref{fig:supp-native-regions} marks the target strip and the four regions used for the spectra.

\paragraph{Spatial detail.}
In general, SHARPnP has the cleanest trade-off between sharpness and overshooting around the spatial profile.
For instance, at 650 nm, Bicubic spreads the transition from bright to dark over about six pixels at both exposures. SHARPnP narrows it to about four pixels (Figure~\ref{fig:supp-native-profiles}). SSC-S gives an even sharper edge, but at 30 ms its intensity overshoots the edge by about 24\% of the bright-to-dark intensity difference. The corresponding overshoot for SHARPnP is below 1\%.

\paragraph{Spectral consistency.}
At 480 ms, SHARPnP agrees most closely with the averaged HSI measurements in both VIS and NIR, giving the lowest RMSE and SAM (Table~\ref{tab:supp-native-consistency}). CMF-GSA gives the most uniform Spectralon patch, but it underestimates its brightness. At 802 nm, the measured intensity is 0.90, compared with 0.78 for CMF-GSA and 0.90 for SHARPnP (Figure~\ref{fig:supp-native-spectra}).

\begin{figure}[b]
    \centering
    \begin{subfigure}[t]{.49\linewidth}
        \centering
        \includegraphics[width=\linewidth]{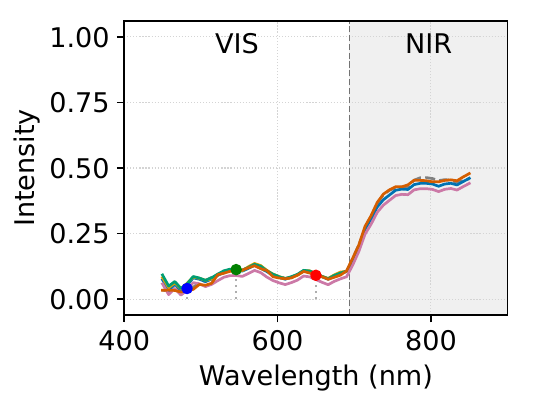}
        \caption{Real leaf}
    \end{subfigure}\hfill%
    \begin{subfigure}[t]{.49\linewidth}
        \centering
        \includegraphics[width=\linewidth]{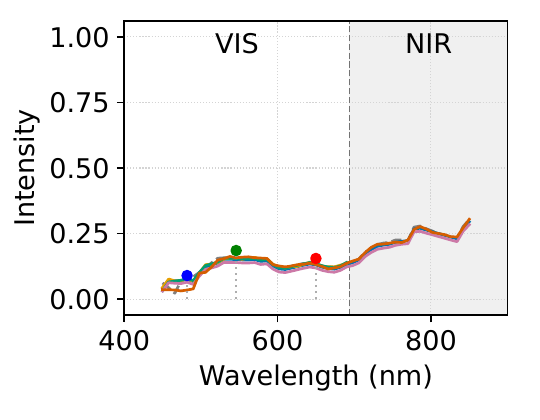}
        \caption{Artificial leaf}
    \end{subfigure}\par
    \begin{subfigure}[t]{.49\linewidth}
        \centering
        \includegraphics[width=\linewidth]{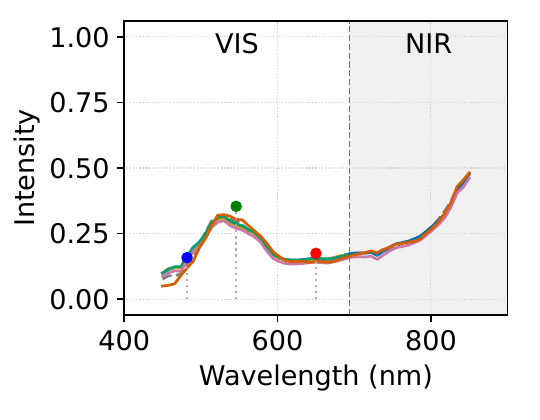}
        \caption{Green patch}
    \end{subfigure}\hfill%
    \begin{subfigure}[t]{.49\linewidth}
        \centering
        \includegraphics[width=\linewidth]{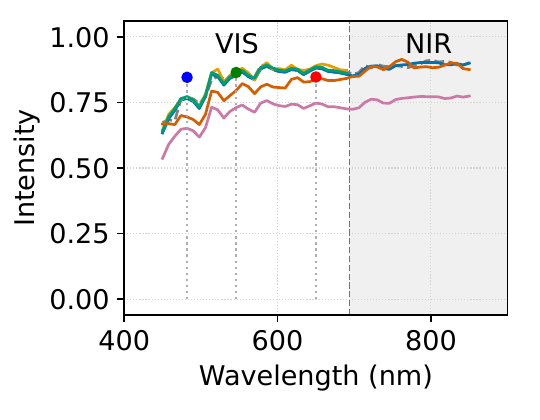}
        \caption{White reference}
    \end{subfigure}\par
    \includegraphics[width=\linewidth]{figures/supplementary/native_diagnostics/spectrum_legend.pdf}
    \caption{Patch spectra at 30 ms. The dashed line shows the average of ten HSI frames.}
    \label{fig:supp-native-spectra-30ms}
\end{figure}

At 30 ms, SHARPnP gives the lowest values on all six measures (Table~\ref{tab:supp-native-30ms}). The largest difference is on the Spectralon: the variation within this white patch is about one third of that of CMF-GSA, the next best method. SHARPnP also preserves the difference between the two leaves (Figure~\ref{fig:supp-native-spectra-30ms}). At 802 nm, it reconstructs intensities of 0.45 for the real leaf and 0.25 for the artificial leaf, close to the measured values of 0.46 and 0.26.

\begin{table}[tbp]
    \centering
    \caption{Static scene, 30 ms exposure.}
    \label{tab:supp-native-30ms}
    \setlength{\tabcolsep}{1.75pt}
    \footnotesize
    \begin{tabular*}{\linewidth}{@{\extracolsep{\fill}}lrrrrrr@{}}
        \toprule
         & \multicolumn{2}{c}{VIS} & \multicolumn{2}{c}{NIR} & \multicolumn{2}{c}{VIS patches} \\
         & \multicolumn{2}{c}{$[450,690]$~nm} & \multicolumn{2}{c}{$[698,850]$~nm} & \multicolumn{2}{c}{} \\
        \cmidrule(lr){2-3}\cmidrule(lr){4-5}\cmidrule(l){6-7}
        Method & RMSE$\downarrow$ & SAM$\downarrow$ & RMSE$\downarrow$ & SAM$\downarrow$ & N-SAM$\downarrow$ & S-Var$\downarrow$ \\
               & ($10^{-2}$) & ($^\circ$) & ($10^{-2}$) & ($^\circ$) & ($^\circ$) & ($10^{-2}$) \\
        \midrule
        Bicubic & 2.759 & 7.983 & 2.254 & 2.397 & 4.396 & 5.077 \\
        SSC-S~\cite{zhang2025sschs} & 3.056 & 8.506 & -- & -- & 4.544 & 8.099 \\
        UAFL~\cite{zhang2026uafl} & 3.087 & 8.599 & -- & -- & 4.903 & 7.743 \\
        CMF-GSA~\cite{jouni2026bmvc} & 3.073 & 8.385 & 2.728 & 2.471 & 5.150 & 4.916 \\
        SHARPnP & \textbf{2.201} & \textbf{6.957} & \textbf{1.389} & \textbf{2.152} & \textbf{3.622} & \textbf{1.646} \\
        \bottomrule
    \end{tabular*}
\end{table}

\subsection{Second Native Scene.}
\label{subsec:bmvc_scene}

\paragraph{Experiment settings.}
We consider one frame from the static Ultris SR5 and Raspberry Pi HQ sequence described in~\cite{jouni2026bmvc}.
SHARPnP reconstructs all 51 bands from 450 to 850 nm using 20 HQS iterations and frozen DRUNet with $P_D=I_{51}$ (the $51\times51$ identity matrix). The normalized weights for RGB values and gradients are $0.009$ and $0.200$, respectively.

SSC-S and UAFL process only the first 31 bands, covering the visible range from 450 to 690 nm. Alternatively, retraining them on all 51 bands would require clean high-resolution HSI references, which are not available here.

Since we do not have clean high-resolution HSI references, we perform a qualitative and quantitative analysis based on controlled spectral and spatial materials in the scene.

\begin{figure*}[t]
    \centering

    \raisebox{3mm}{\rotatebox[origin=c]{90}{\scriptsize \textbf{RGB views}}}\hfill%
    \begin{minipage}[c]{.974\linewidth}\centering
        \begin{minipage}[t]{.145\textwidth}
            \centering
            \includegraphics[width=\linewidth]{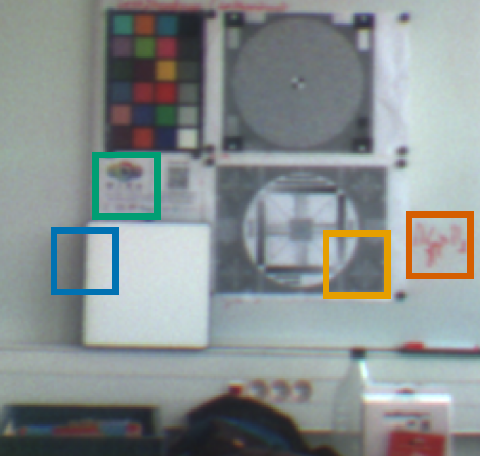}%
            \\[1pt]
            \includegraphics[width=0.4875\linewidth]{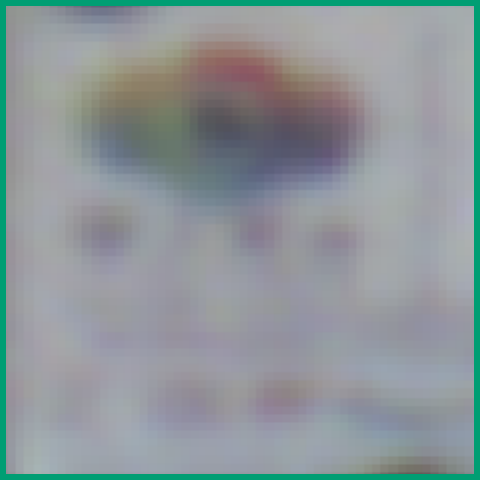}%
            \hfill
            \includegraphics[width=0.4875\linewidth]{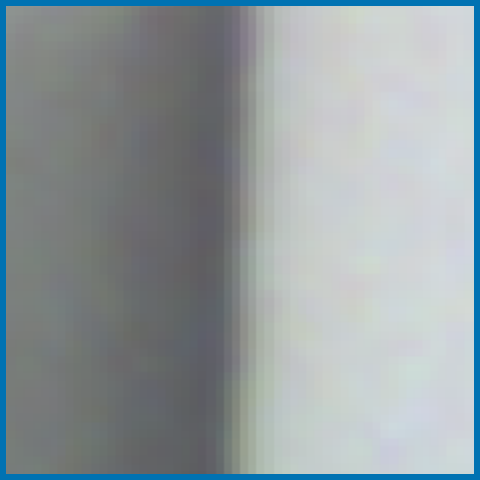}%
            \\[1pt]
            \includegraphics[width=0.4875\linewidth]{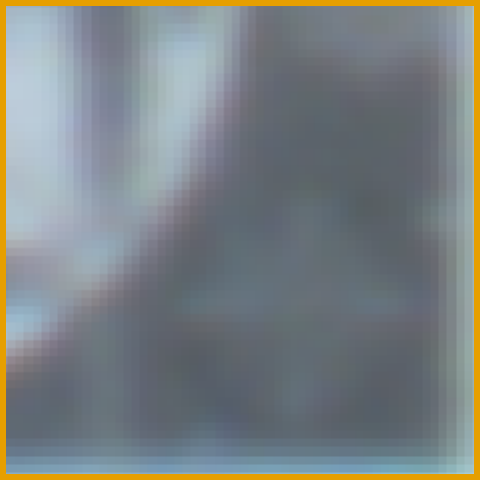}%
            \hfill
            \includegraphics[width=0.4875\linewidth]{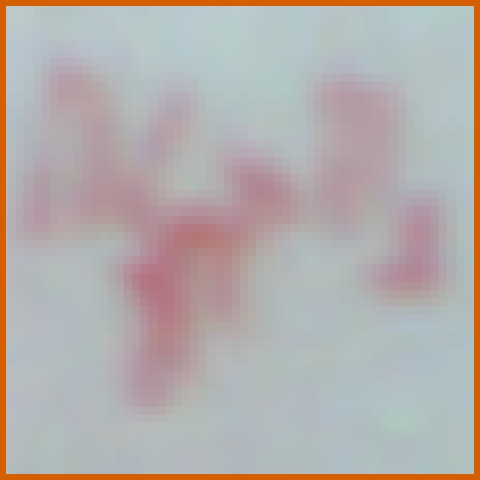}%
            \\[1pt]
            {\scriptsize\mbox{(a) Bicubic}}
        \end{minipage}%
        \hfill
        \begin{minipage}[t]{.145\textwidth}
            \centering
            \includegraphics[width=\linewidth]{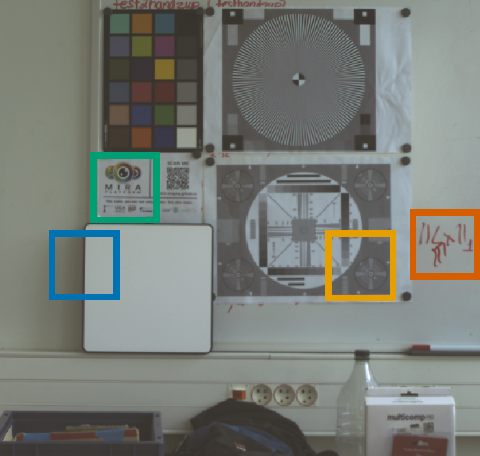}%
            \\[1pt]
            \includegraphics[width=0.4875\linewidth]{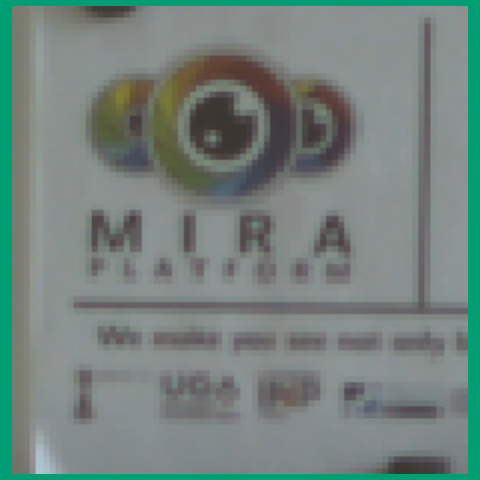}%
            \hfill
            \includegraphics[width=0.4875\linewidth]{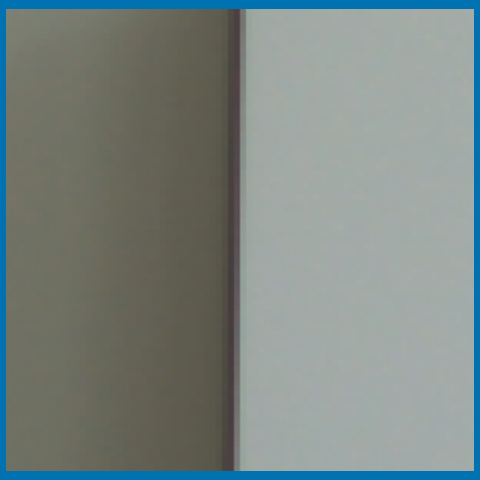}%
            \\[1pt]
            \includegraphics[width=0.4875\linewidth]{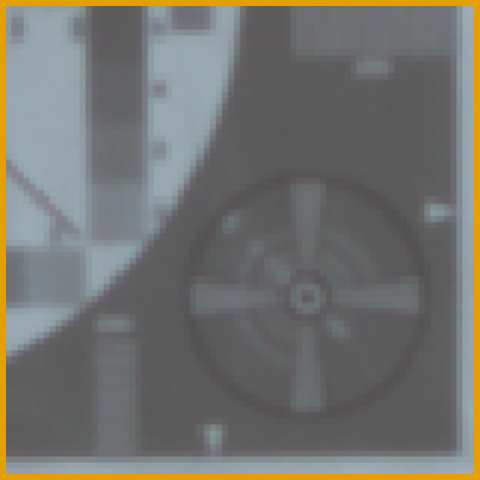}%
            \hfill
            \includegraphics[width=0.4875\linewidth]{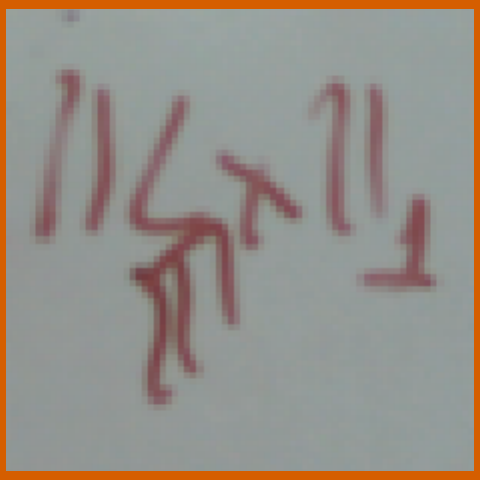}%
            \\[1pt]
            {\scriptsize\mbox{(b) RGB guide}}
        \end{minipage}%
        \hfill
        \begin{minipage}[t]{.145\textwidth}
            \centering
            \includegraphics[width=\linewidth]{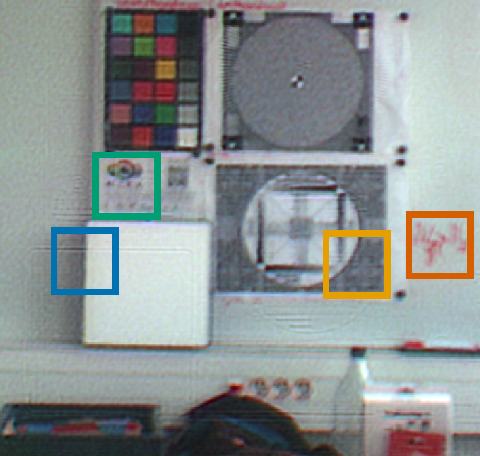}%
            \\[1pt]
            \includegraphics[width=0.4875\linewidth]{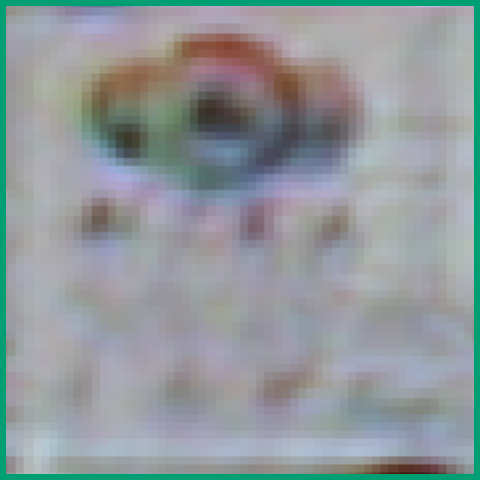}%
            \hfill
            \includegraphics[width=0.4875\linewidth]{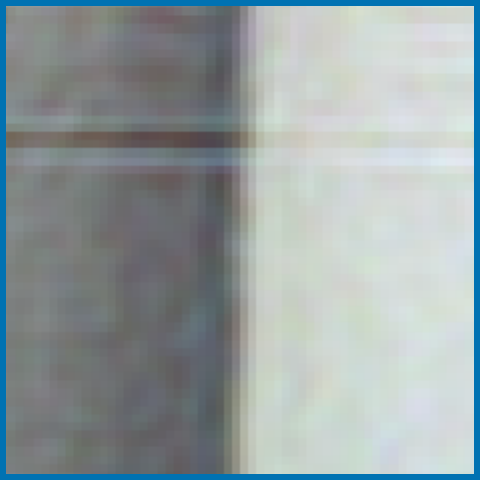}%
            \\[1pt]
            \includegraphics[width=0.4875\linewidth]{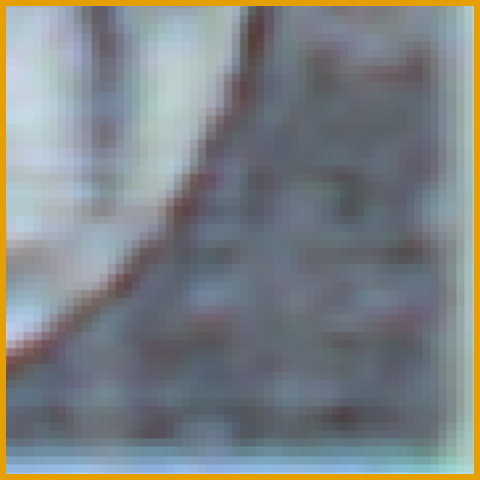}%
            \hfill
            \includegraphics[width=0.4875\linewidth]{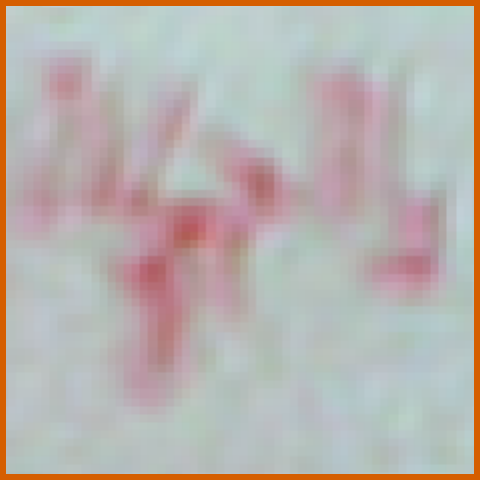}%
            \\[1pt]
            {\scriptsize\mbox{(c) SSC-S~\cite{zhang2025sschs}}}
        \end{minipage}%
        \hfill
        \begin{minipage}[t]{.145\textwidth}
            \centering
            \includegraphics[width=\linewidth]{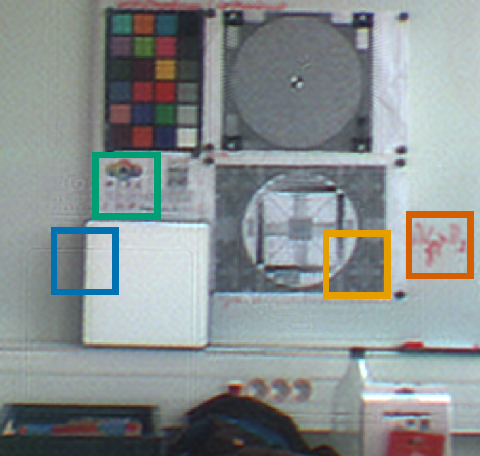}%
            \\[1pt]
            \includegraphics[width=0.4875\linewidth]{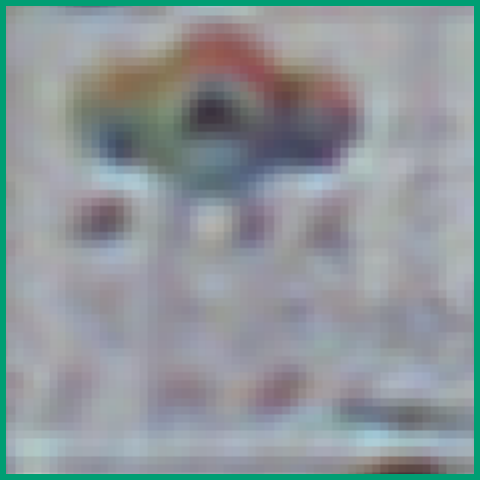}%
            \hfill
            \includegraphics[width=0.4875\linewidth]{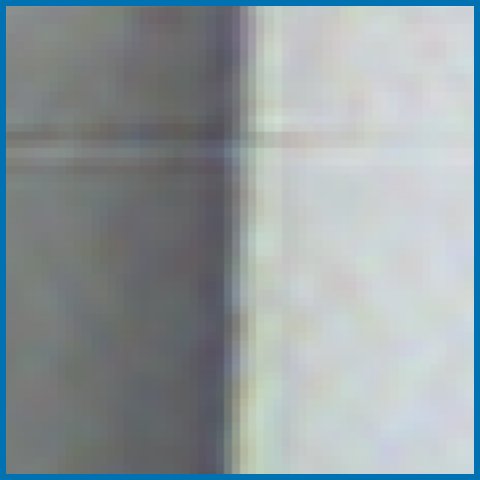}%
            \\[1pt]
            \includegraphics[width=0.4875\linewidth]{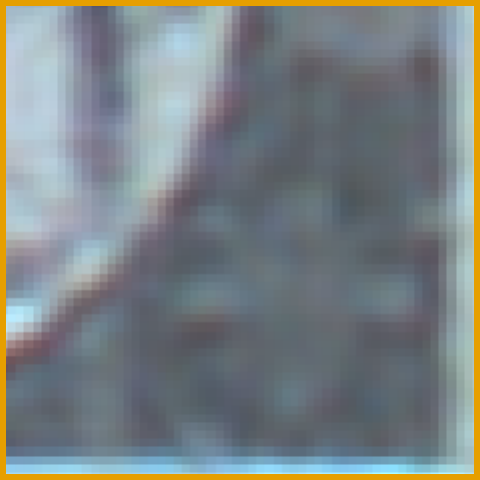}%
            \hfill
            \includegraphics[width=0.4875\linewidth]{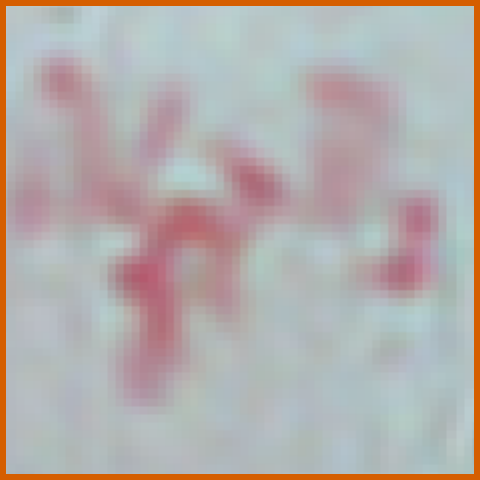}%
            \\[1pt]
            {\scriptsize\mbox{(d) UAFL~\cite{zhang2026uafl}}}
        \end{minipage}%
        \hfill
        \begin{minipage}[t]{.145\textwidth}
            \centering
            \includegraphics[width=\linewidth]{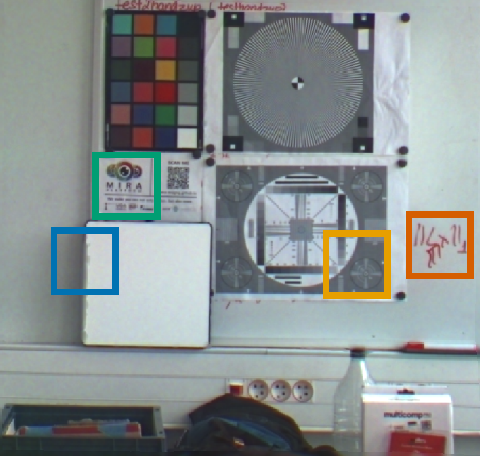}%
            \\[1pt]
            \includegraphics[width=0.4875\linewidth]{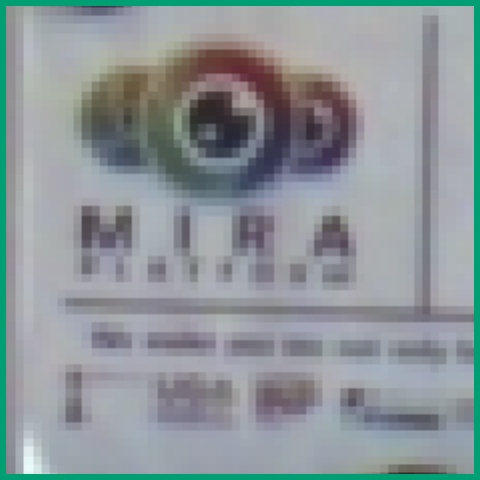}%
            \hfill
            \includegraphics[width=0.4875\linewidth]{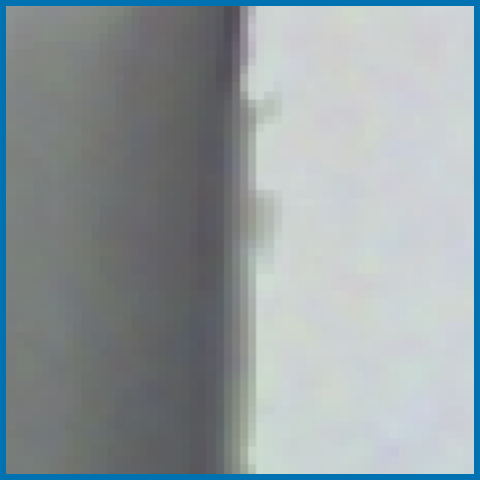}%
            \\[1pt]
            \includegraphics[width=0.4875\linewidth]{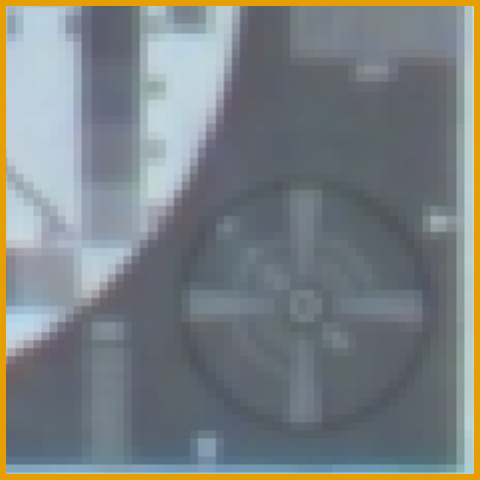}%
            \hfill
            \includegraphics[width=0.4875\linewidth]{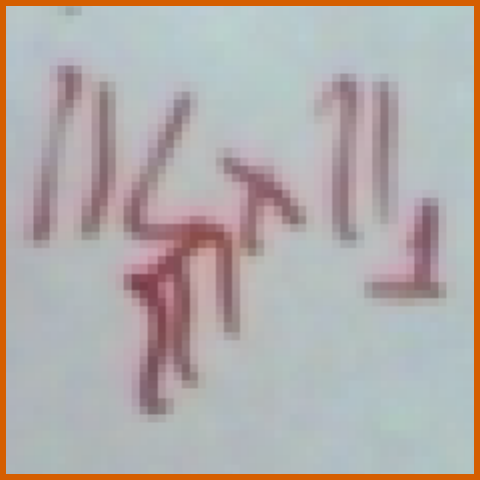}%
            \\[1pt]
            {\scriptsize\mbox{(e) CMF-GSA-wg~\cite{jouni2026bmvc}}}
        \end{minipage}%
        \hfill
        \begin{minipage}[t]{.145\textwidth}
            \centering
            \includegraphics[width=\linewidth]{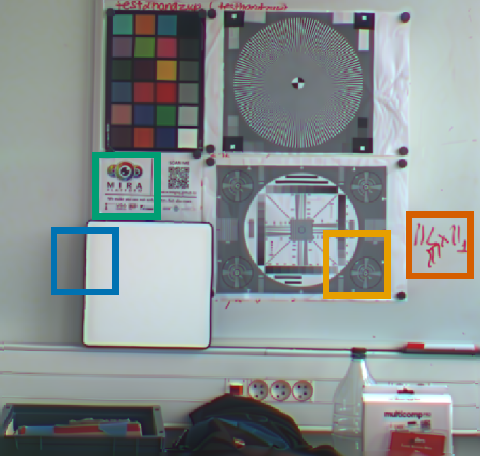}%
            \\[1pt]
            \includegraphics[width=0.4875\linewidth]{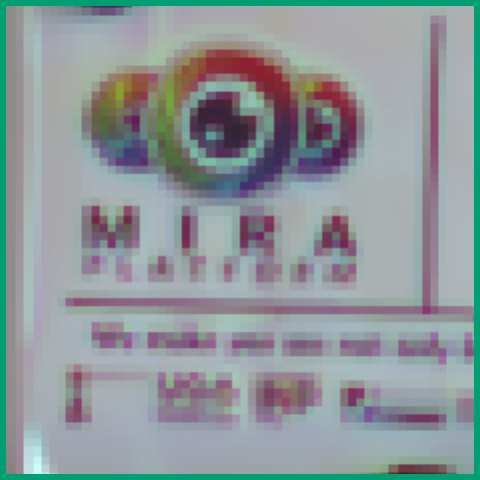}%
            \hfill
            \includegraphics[width=0.4875\linewidth]{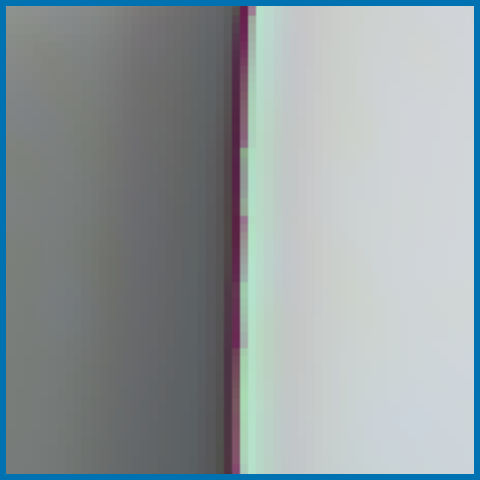}%
            \\[1pt]
            \includegraphics[width=0.4875\linewidth]{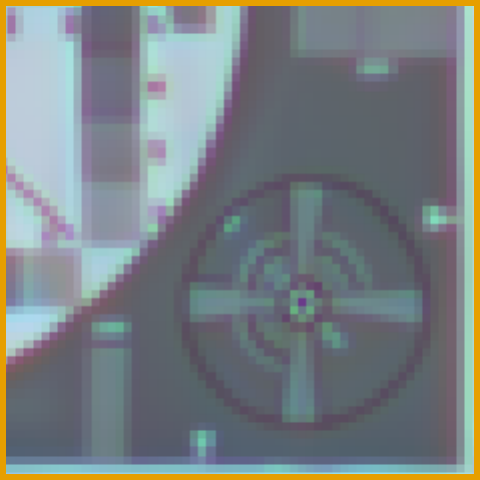}%
            \hfill
            \includegraphics[width=0.4875\linewidth]{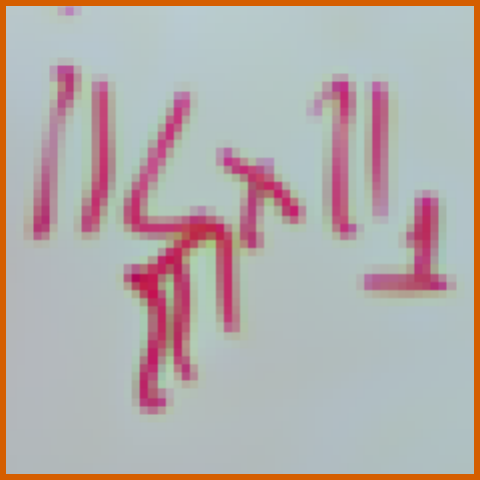}%
            \\[1pt]
            {\scriptsize\mbox{(f) SHARPnP (ours)}}
        \end{minipage}%
    \end{minipage}\par

\vspace{2mm}

    \raisebox{8mm}{\rotatebox[origin=c]{90}{\scriptsize \textbf{VIS RMSE}}}\hfill%
    \begin{minipage}[c]{.974\linewidth}\centering
        \begin{minipage}[t]{.145\textwidth}
            \centering
            \includegraphics[width=\linewidth]{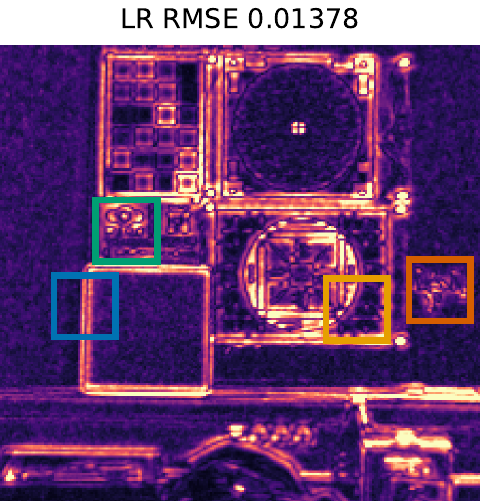}%
            \\[1pt]
            \includegraphics[width=0.4875\linewidth]{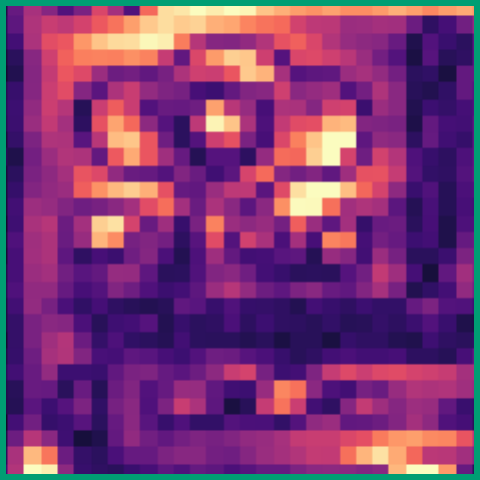}%
            \hfill
            \includegraphics[width=0.4875\linewidth]{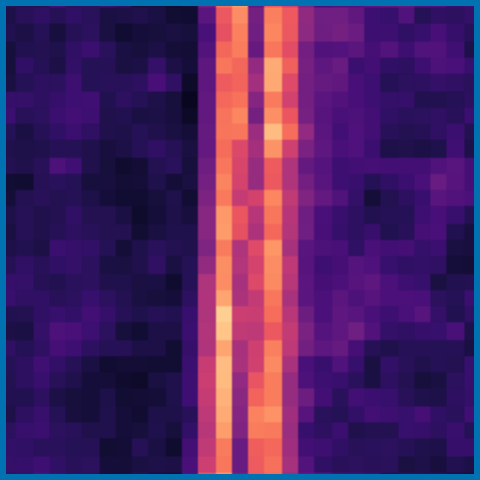}%
            \\[1pt]
            \includegraphics[width=0.4875\linewidth]{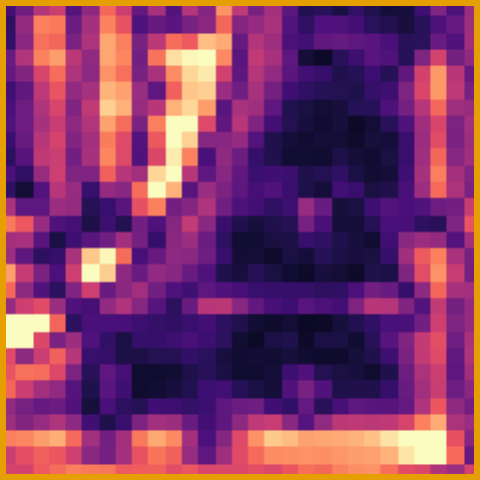}%
            \hfill
            \includegraphics[width=0.4875\linewidth]{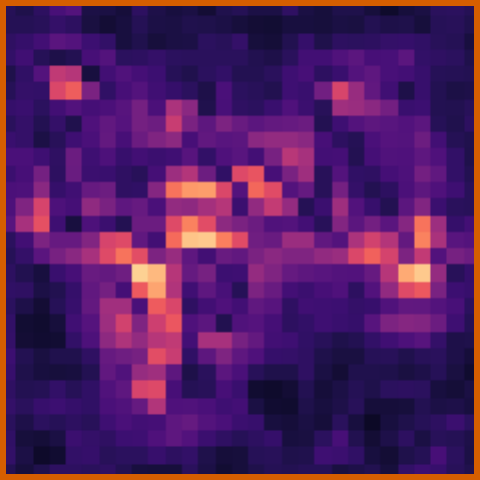}%
            \\[1pt]
            {\scriptsize\mbox{(a) Bicubic}}
        \end{minipage}%
        \hfill
        \begin{minipage}[t]{.145\textwidth}
            \centering
            \includegraphics[width=\linewidth]{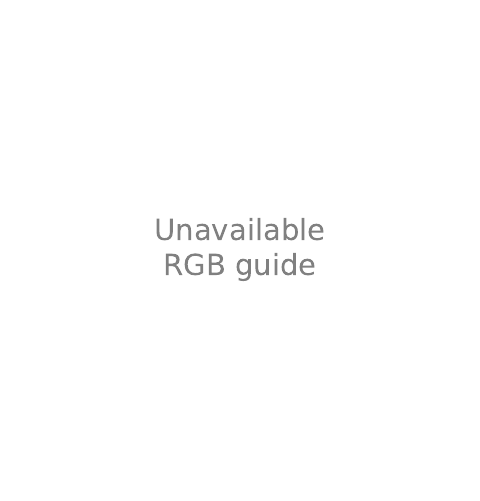}%
            \\[1pt]
            \includegraphics[width=0.4875\linewidth]{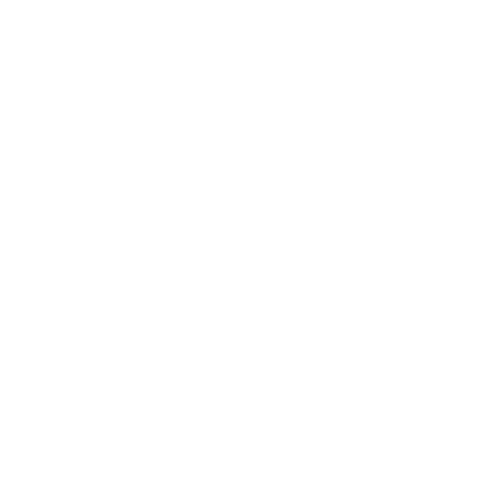}%
            \hfill
            \includegraphics[width=0.4875\linewidth]{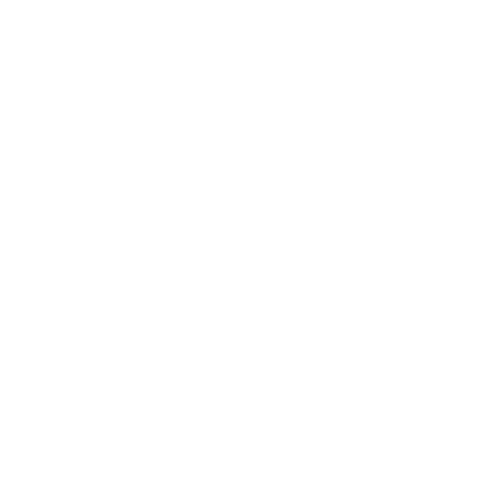}%
            \\[1pt]
            \includegraphics[width=0.4875\linewidth]{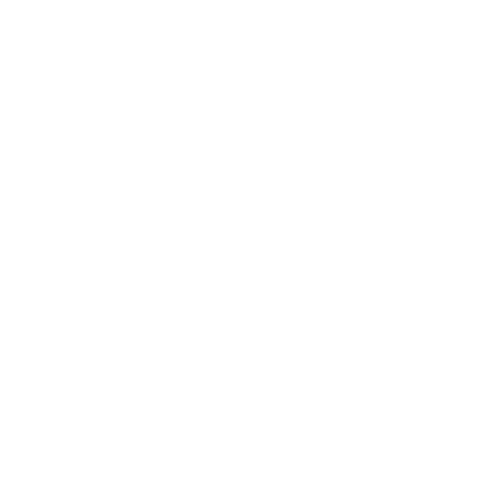}%
            \hfill
            \includegraphics[width=0.4875\linewidth]{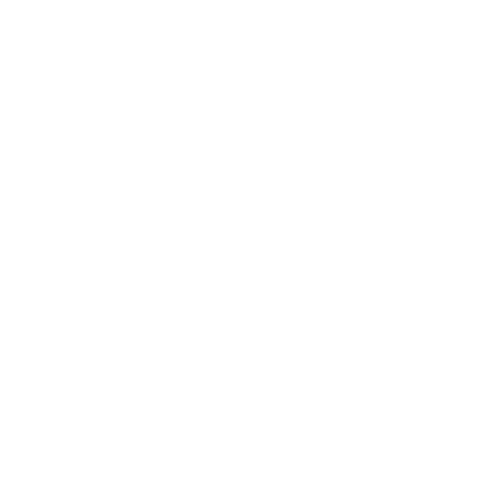}%
            \\[1pt]
            {\scriptsize\mbox{(b) RGB guide}}
        \end{minipage}%
        \hfill
        \begin{minipage}[t]{.145\textwidth}
            \centering
            \includegraphics[width=\linewidth]{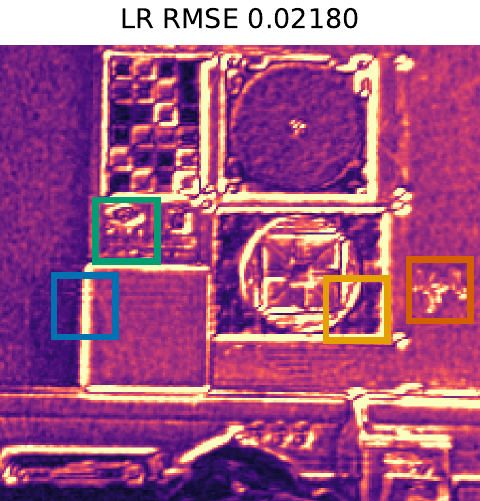}%
            \\[1pt]
            \includegraphics[width=0.4875\linewidth]{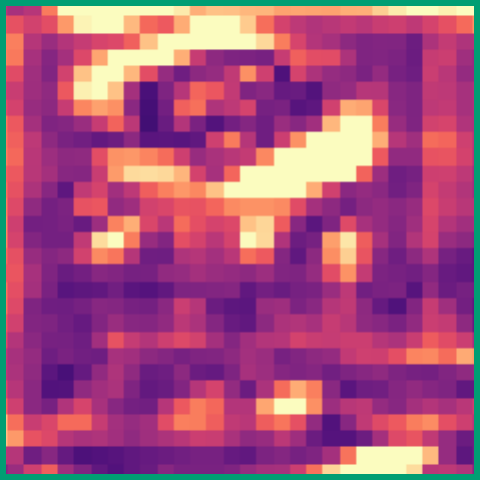}%
            \hfill
            \includegraphics[width=0.4875\linewidth]{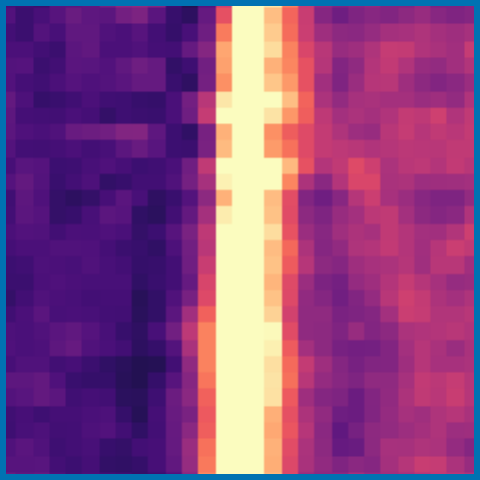}%
            \\[1pt]
            \includegraphics[width=0.4875\linewidth]{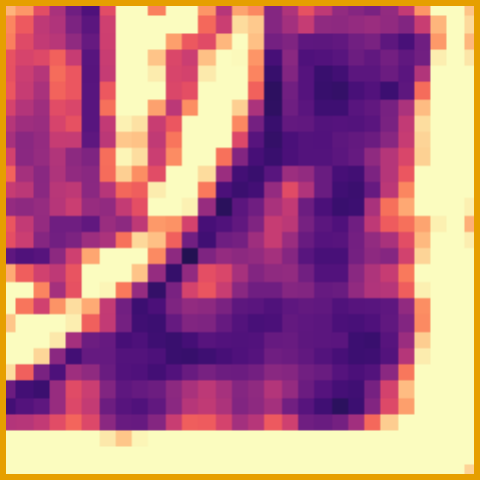}%
            \hfill
            \includegraphics[width=0.4875\linewidth]{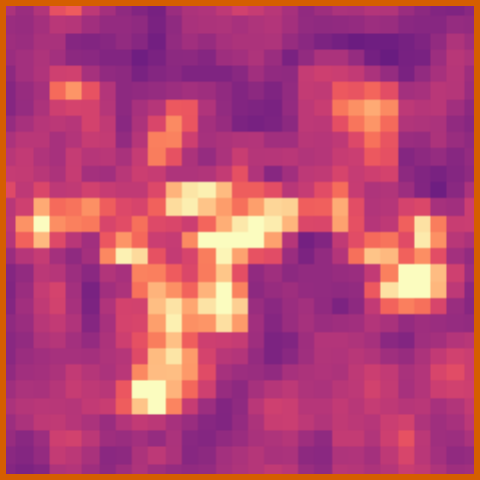}%
            \\[1pt]
            {\scriptsize\mbox{(c) SSC-S~\cite{zhang2025sschs}}}
        \end{minipage}%
        \hfill
        \begin{minipage}[t]{.145\textwidth}
            \centering
            \includegraphics[width=\linewidth]{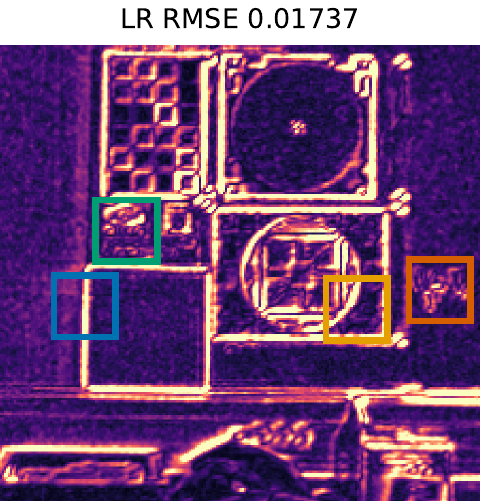}%
            \\[1pt]
            \includegraphics[width=0.4875\linewidth]{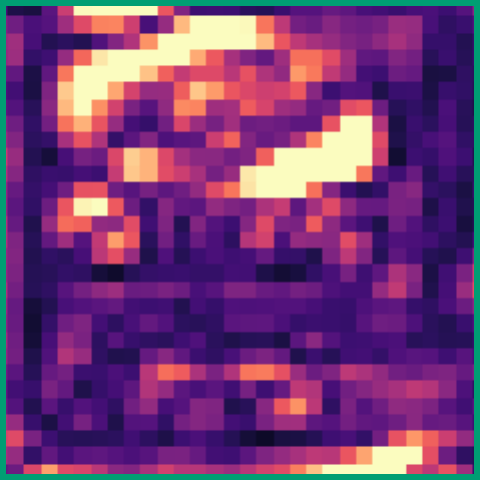}%
            \hfill
            \includegraphics[width=0.4875\linewidth]{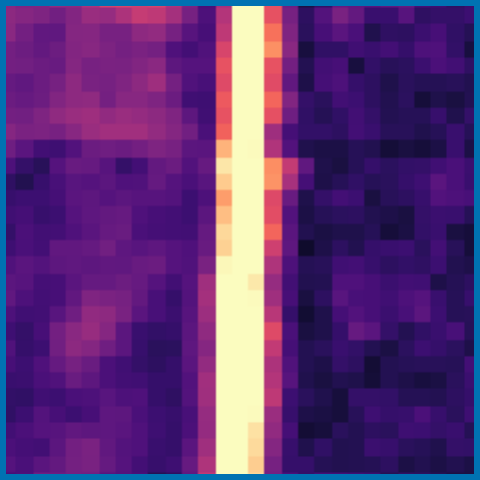}%
            \\[1pt]
            \includegraphics[width=0.4875\linewidth]{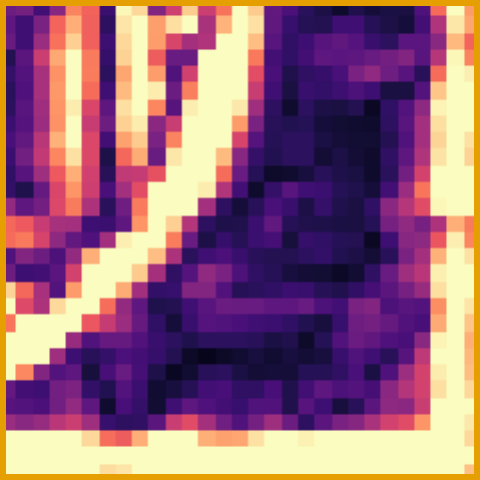}%
            \hfill
            \includegraphics[width=0.4875\linewidth]{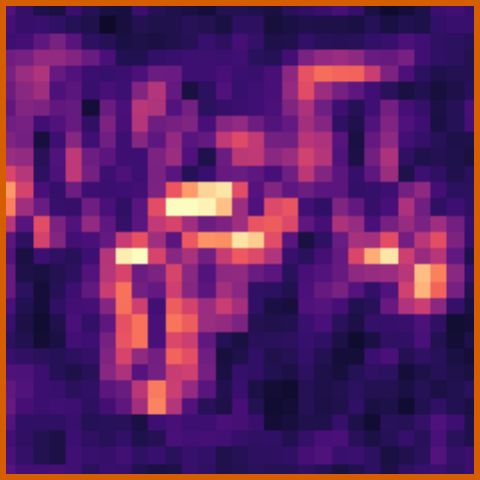}%
            \\[1pt]
            {\scriptsize\mbox{(d) UAFL~\cite{zhang2026uafl}}}
        \end{minipage}%
        \hfill
        \begin{minipage}[t]{.145\textwidth}
            \centering
            \includegraphics[width=\linewidth]{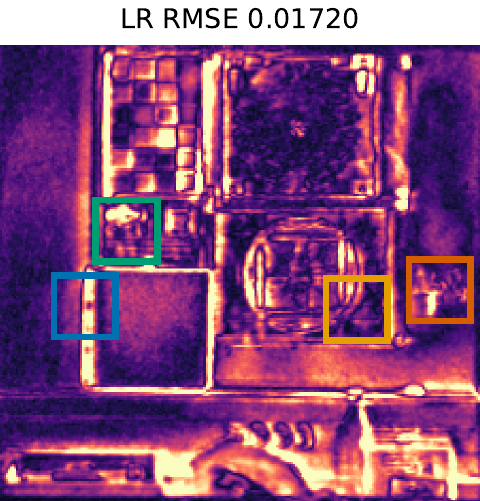}%
            \\[1pt]
            \includegraphics[width=0.4875\linewidth]{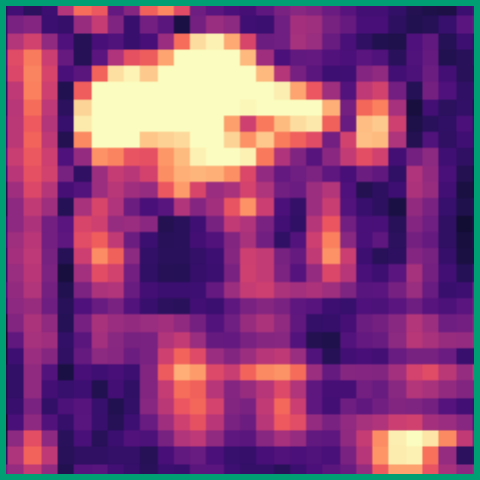}%
            \hfill
            \includegraphics[width=0.4875\linewidth]{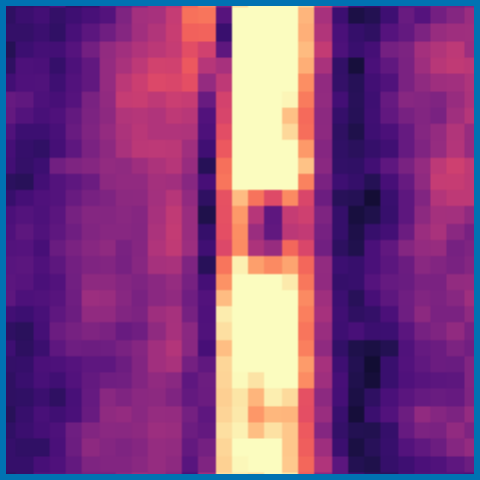}%
            \\[1pt]
            \includegraphics[width=0.4875\linewidth]{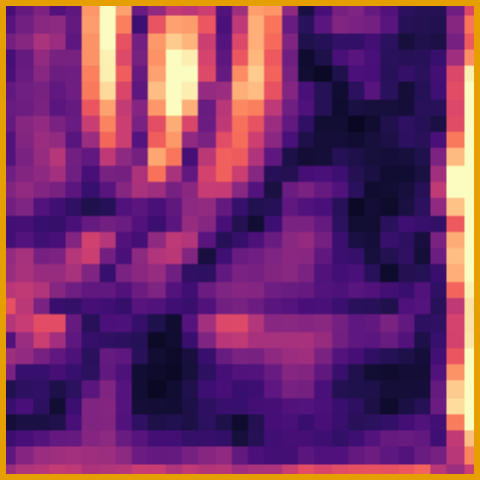}%
            \hfill
            \includegraphics[width=0.4875\linewidth]{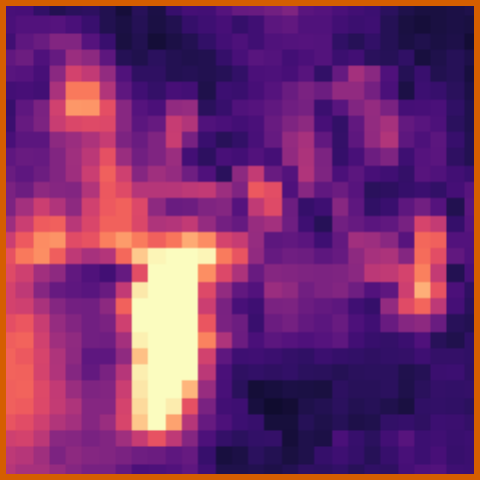}%
            \\[1pt]
            {\scriptsize\mbox{(e) CMF-GSA-wg~\cite{jouni2026bmvc}}}
        \end{minipage}%
        \hfill
        \begin{minipage}[t]{.145\textwidth}
            \centering
            \includegraphics[width=\linewidth]{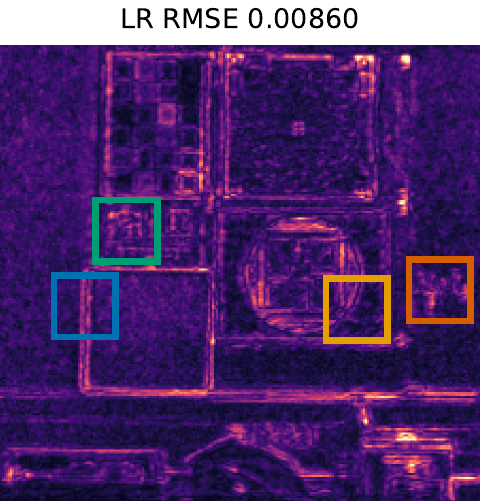}%
            \\[1pt]
            \includegraphics[width=0.4875\linewidth]{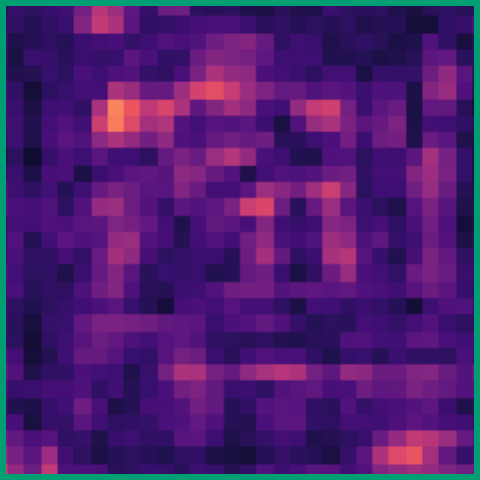}%
            \hfill
            \includegraphics[width=0.4875\linewidth]{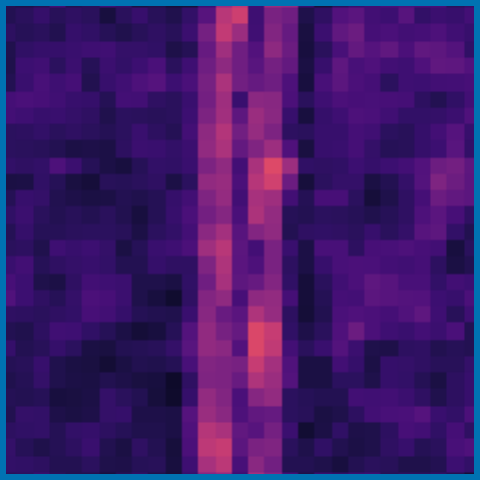}%
            \\[1pt]
            \includegraphics[width=0.4875\linewidth]{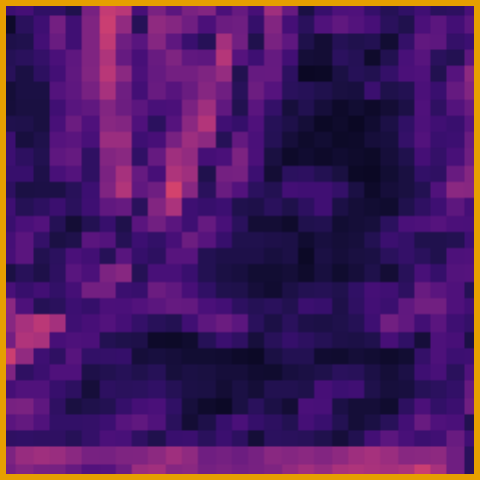}%
            \hfill
            \includegraphics[width=0.4875\linewidth]{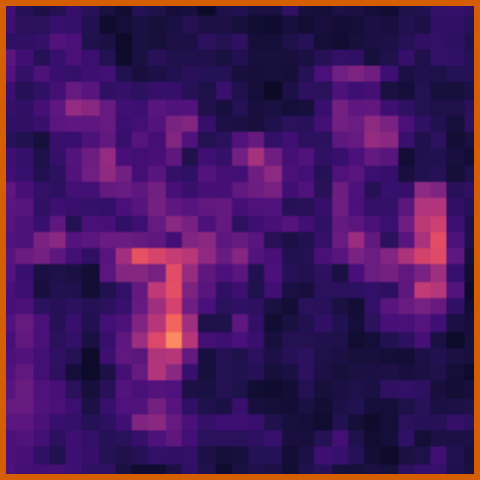}%
            \\[1pt]
            {\scriptsize\mbox{(f) SHARPnP (ours)}}
        \end{minipage}%
        \\[0pt]
        \includegraphics[width=.42\textwidth]{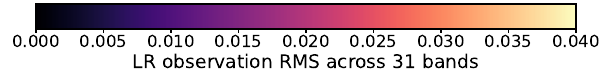}
    \end{minipage}\par
    \caption{Additional native Ultris-RPi example from the setup used in CMF-GSA-wg~\cite{jouni2026bmvc}. The rows show RGB views and VIS RMSE maps computed against the averaged HSI observations. Crops show small textures, the Spectralon edge, the target corner, and wall writing. Clean references are not available.}
    \label{fig:supp-bmvc}
    \label{fig:supp-bmvc-rmse}
\end{figure*}

\begin{figure}[tbp]
    \centering
    \begin{minipage}[t]{.65\linewidth}
        \centering
        \begin{minipage}[c]{.48\linewidth}
            \includegraphics[width=\linewidth]{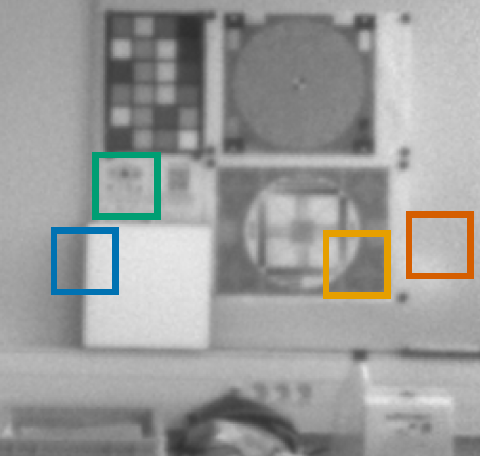}
        \end{minipage}%
        \hfill
        \begin{minipage}[c]{.49\linewidth}
            \centering
            \includegraphics[width=0.4875\linewidth]{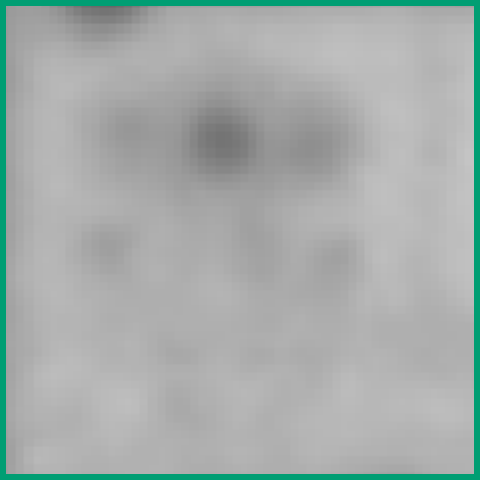}%
            \hfill
            \includegraphics[width=0.4875\linewidth]{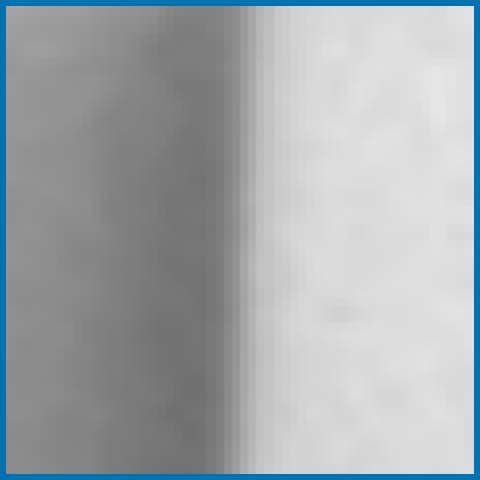}%
            \\[1pt]
            \includegraphics[width=0.4875\linewidth]{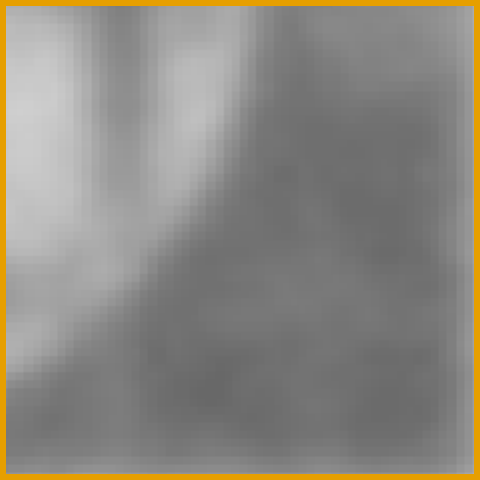}%
            \hfill
            \includegraphics[width=0.4875\linewidth]{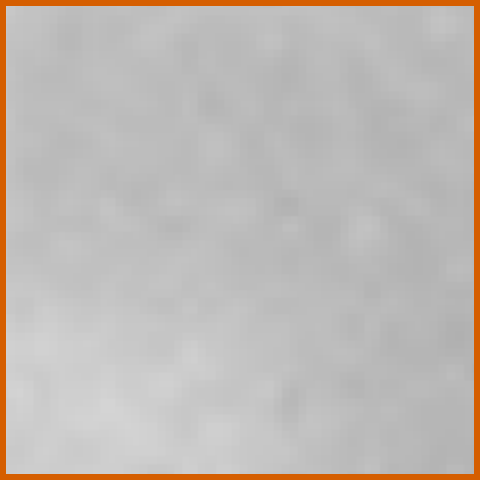}%

        \end{minipage}\\[1pt]
        {\scriptsize\mbox{(a) Bicubic}}
    \end{minipage}%
    \par\smallskip
    \begin{minipage}[t]{.65\linewidth}
        \centering
        \begin{minipage}[c]{.48\linewidth}
            \includegraphics[width=\linewidth]{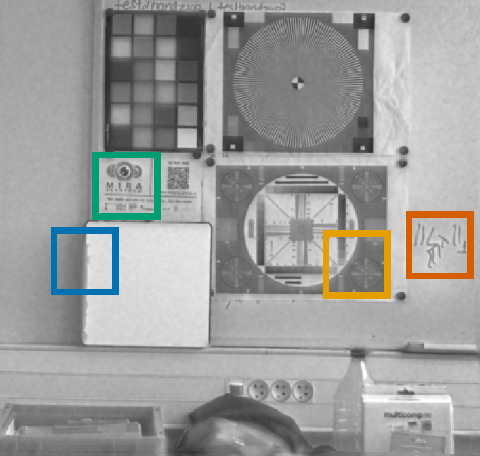}
        \end{minipage}%
        \hfill
        \begin{minipage}[c]{.49\linewidth}
            \centering
            \includegraphics[width=0.4875\linewidth]{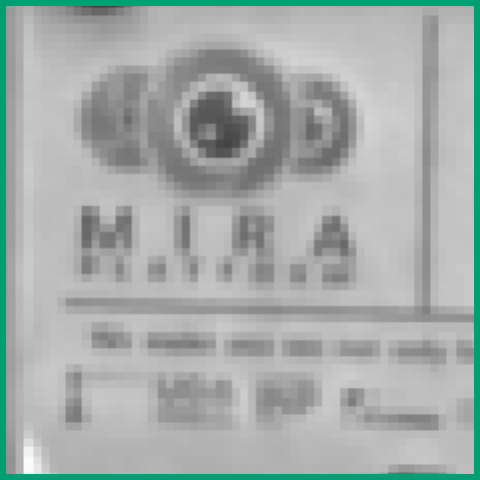}%
            \hfill
            \includegraphics[width=0.4875\linewidth]{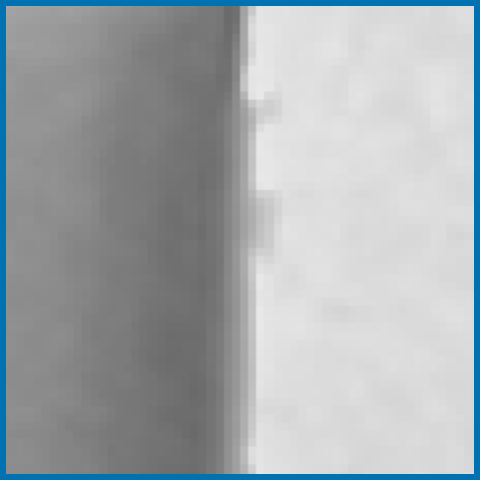}%
            \\[1pt]
            \includegraphics[width=0.4875\linewidth]{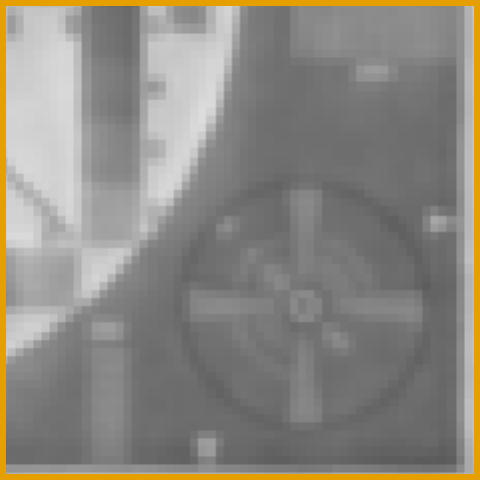}%
            \hfill
            \includegraphics[width=0.4875\linewidth]{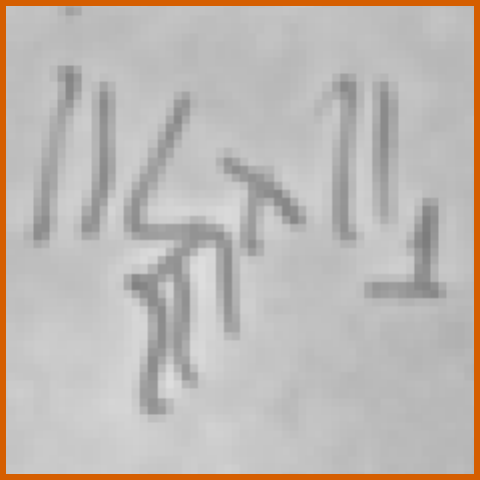}%

        \end{minipage}\\[1pt]
        {\scriptsize\mbox{(b) CMF-GSA-wg~\cite{jouni2026bmvc}}}
    \end{minipage}%
    \par\smallskip
    \begin{minipage}[t]{.65\linewidth}
        \centering
        \begin{minipage}[c]{.48\linewidth}
            \includegraphics[width=\linewidth]{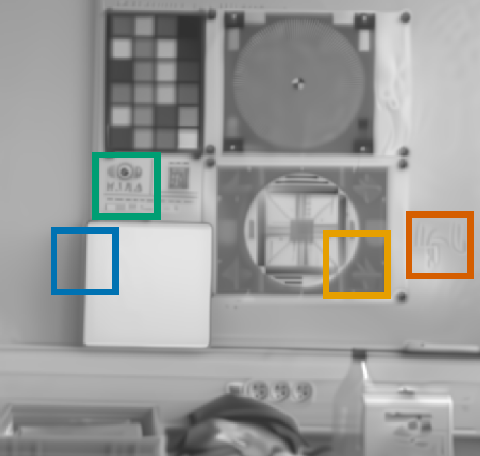}
        \end{minipage}%
        \hfill
        \begin{minipage}[c]{.49\linewidth}
            \centering
            \includegraphics[width=0.4875\linewidth]{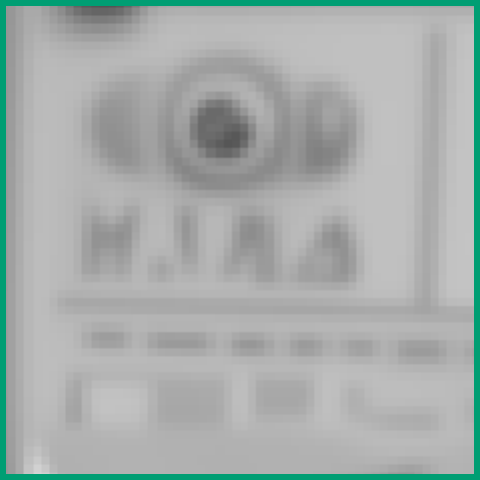}%
            \hfill
            \includegraphics[width=0.4875\linewidth]{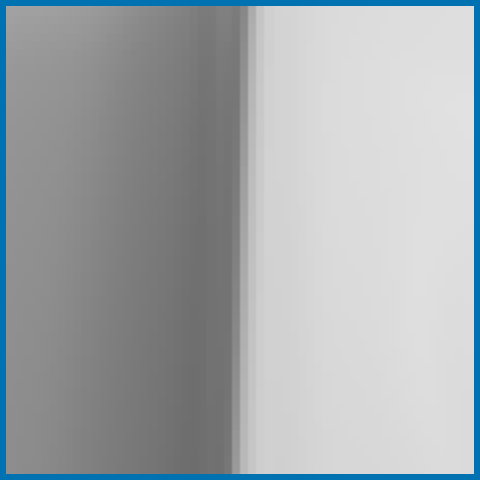}%
            \\[1pt]
            \includegraphics[width=0.4875\linewidth]{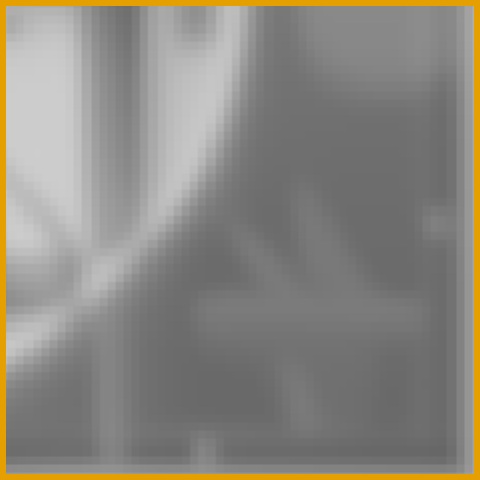}%
            \hfill
            \includegraphics[width=0.4875\linewidth]{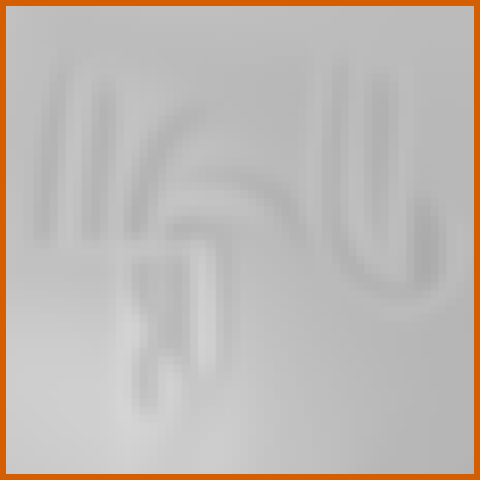}%

        \end{minipage}\\[1pt]
        {\scriptsize\mbox{(c) SHARPnP (ours)}}
    \end{minipage}%
    \caption{Reconstructions at 802 nm. Each row shows the full image and the same four crops as Figure~\ref{fig:supp-bmvc}.}
    \label{fig:supp-bmvc-nir}
\end{figure}

\begin{figure}[tbp]
    \centering
    \begin{subfigure}[t]{.49\linewidth}
        \centering
        \includegraphics[width=\linewidth]{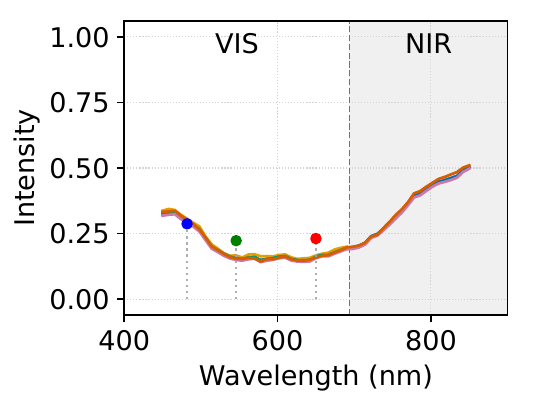}
        \caption{Blue patch}
    \end{subfigure}\hfill%
    \begin{subfigure}[t]{.49\linewidth}
        \centering
        \includegraphics[width=\linewidth]{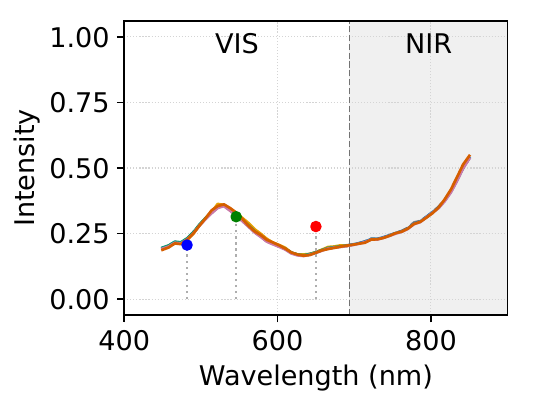}
        \caption{Green patch}
    \end{subfigure}\par
    \begin{subfigure}[t]{.49\linewidth}
        \centering
        \includegraphics[width=\linewidth]{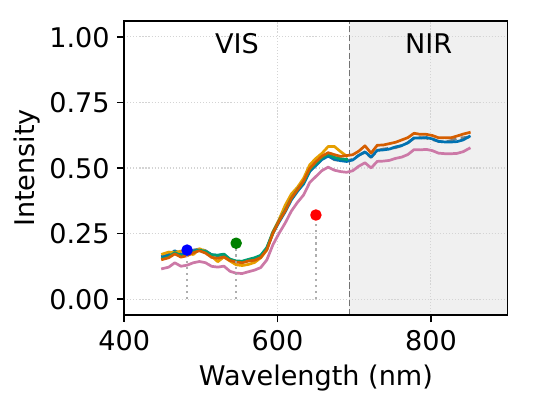}
        \caption{Red patch}
    \end{subfigure}\hfill%
    \begin{subfigure}[t]{.49\linewidth}
        \centering
        \includegraphics[width=\linewidth]{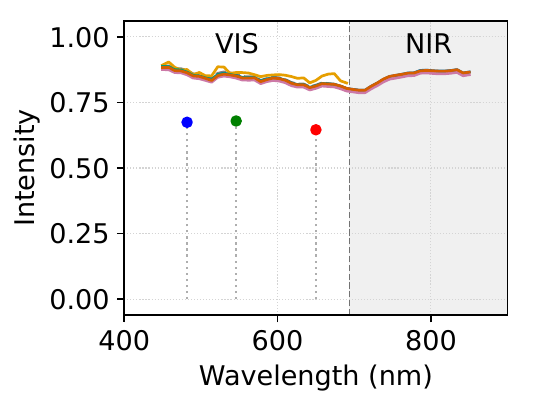}
        \caption{White reference}
    \end{subfigure}\par
    \includegraphics[width=\linewidth]{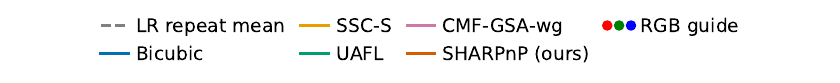}
    \caption{Mean spectra of three ColorChecker patches and the Spectralon. The dashed curve shows the average of four HSI frames. The divider at 694 nm separates VIS and NIR.}
    \label{fig:supp-bmvc-spectra}
\end{figure}

Figures~\ref{fig:supp-bmvc}, \ref{fig:supp-bmvc-nir}, and \ref{fig:supp-bmvc-spectra} show the visible reconstructions, the 802 nm band, and the patch spectra, respectively. Table~\ref{tab:native} compares the reconstructions with temporally averaged HSI observations of the same scene to reduce sensor noise and focus on structural errors and object edges. Some residual variation may still appear because the averaged observations may still contain noise.

This example complements the native acquisition in Section 5.3 of the main paper.

\begin{table}[t]
    \centering
    \caption{Additional native frame from the Ultris-RPi sequence described in~\cite{jouni2026bmvc}.}
    \label{tab:native}
    \setlength{\tabcolsep}{1.75pt}
    \footnotesize
    \begin{tabular*}{\linewidth}{@{\extracolsep{\fill}}lrrrrrr@{}}
        \toprule
         & \multicolumn{2}{c}{VIS} & \multicolumn{2}{c}{NIR} & \multicolumn{2}{c}{VIS patches} \\
         & \multicolumn{2}{c}{$[450,690]$~nm} & \multicolumn{2}{c}{$[698,850]$~nm} & \multicolumn{2}{c}{} \\
        \cmidrule(lr){2-3}\cmidrule(lr){4-5}\cmidrule(l){6-7}
        Method & RMSE$\downarrow$ & SAM$\downarrow$ & RMSE$\downarrow$ & SAM$\downarrow$ & N-SAM$\downarrow$ & S-Var$\downarrow$ \\
               & ($10^{-2}$) & ($^\circ$) & ($10^{-2}$) & ($^\circ$) & ($^\circ$) & ($10^{-2}$) \\
        \midrule
        Bicubic & 1.378 & 0.845 & 1.118 & 0.625 & \textbf{0.497} & 1.478 \\
        HSIFN~\cite{lai2025hsifn} & 2.050 & 1.405 & -- & -- & 1.164 & 1.734 \\
        SSC-S~\cite{zhang2025sschs} & 2.180 & 1.384 & -- & -- & 1.606 & 1.708 \\
        UAFL~\cite{zhang2026uafl} & 1.737 & 0.899 & -- & -- & 0.665 & 1.728 \\
        CMF-GSA~\cite{jouni2026bmvc} & 1.720 & 0.858 & 1.907 & 0.631 & 0.927 & 1.367 \\
        SHARPnP & \textbf{0.860} & \textbf{0.819} & \textbf{0.691} & \textbf{0.567} & 0.810 & \textbf{1.275} \\
        \bottomrule
    \end{tabular*}
\end{table}

\paragraph{Qualitative analysis.}
Figure~\ref{fig:supp-bmvc} shows a qualitative analysis of the results.
In the visible domain, we can see where RGB detail helps and where it can introduce artifacts.
SSC-S and UAFL, transferred from supervised training on the REAL data, show poor generalization to real native RGB-HSI sensor setup. For instance, the overall images propagate some noise, show blurry and poorly reconstructed areas, propagate what looks like artifacts from the unregistered RGB guide (partly visible in the upper-right crop showing the Spectralon edge), show faint horizontal lines crossing the Spectralon edge and extra contours along the lower edge of the whiteboard, and show larger errors in the RMSE maps.
CMF-GSA-wg recovers a cleaner image with more details, but it provides limited noise reduction and a confidence-gated strategy that falls back on the original low-resolution HSI in unreliable regions (partly visible in the upper-right crop showing the Spectralon edge).
SHARPnP gives a cleaner Spectralon edge and recovers more distinct details and strokes, including the gray steps and inner rings on the small target, as well as the lines of the circular target.

Figure~\ref{fig:supp-bmvc-nir} shows the reconstructions at 802 nm. We compare Bicubic, CMF-GSA-wg and SHARPnP, the three methods that reconstruct all 51 bands.
The writing visible in the RGB guide also appears in the CMF-GSA-wg reconstruction, although it is faint or absent in the bicubic NIR crop.
SHARPnP, on the other hand, refines the NIR details and recovers some details with respect to the bicubic, but unlike CMF-GSA-wg, the projection of details from the visible domain is much more suppressed.

The spectra in Figure~\ref{fig:supp-bmvc-spectra} come from the blue, green and red ColorChecker patches, as well as the Spectralon, by considering the averaged spectrum of a neighborhood of pixels to reduce the effect of the noise for a single pixel spectrum. These probes are selected because they are controlled elements known for their spectral properties and cover VIS-NIR ranges. SHARPnP acts on all 51 bands and remains close to the reference.

\paragraph{Quantitative analysis.}
For Table~\ref{tab:native}, we average four HSI frames of the static scene. Figure~\ref{fig:supp-bmvc} shows the \textit{VIS RMSE} maps. We report the median angle over the 24  patches of the ColorChecker as well as the Spectralon patch.

SHARPnP gives the lowest RMSE and SAM in both spectral ranges and the lowest S-Var. Bicubic and UAFL achieve lower N-SAM than SHARPnP, noting that bicubic and UAFL better preserve the average patch spectra, while SHARPnP recovers clearer lettering and target rings in Figure~\ref{fig:supp-bmvc}.

\small
\bibliographystyle{ieeenat_fullname}
\bibliography{main,supplementary}
\end{document}